\documentclass{article} 
\usepackage{iclr2027_conference,times}

\newcommand{\R}{\mathbb R}
\newcommand{\1}{\mathbf 1}
\newcommand{\softmax}{\operatorname{softmax}}
\newcommand{\KL}{\operatorname{KL}}

\usepackage{url}
\usepackage{booktabs,array,graphicx}
\usepackage{enumitem}
\usepackage{amsmath,amssymb,amsthm,mathtools,mathrsfs}

\usepackage{microtype}
\usepackage{hyperref}
\usepackage{xurl}
\usepackage[nameinlink,noabbrev]{cleveref}

\usepackage{wrapfig}

\newcommand{\rank}{\operatorname{rank}}
\newcommand{\CE}{\operatorname{CE}}
\newcommand{\tr}{\operatorname{tr}}
\newcommand{\sym}{\operatorname{sym}}
\newcommand{\op}{\mathrm{op}}
\newcommand{\fro}{\mathrm F}

\newcommand{\cR}{\mathcal R}

\newcommand{\cF}{\mathcal F}
\newcommand{\cH}{\mathcal H}

\newtheorem{theorem}{Theorem}[section]

\newtheorem{lemma}[theorem]{Lemma}
\newtheorem{corollary}[theorem]{Corollary}
\theoremstyle{definition}
\newtheorem{definition}[theorem]{Definition}
\newtheorem{remark}[theorem]{Remark}

\title{Training Dynamics of Large State-Transition Models}

\author{Yee Man Choi \\
University of Waterloo \\
\And
Fengchi Wang\\
University of Waterloo \\
\And
Ruiquan Huang \\
University of Kentucky
}

\iclrfinalcopy 
\begin{document}

\maketitle
\lhead{Preprint} 
\begin{abstract}
Next-token prediction relies on learned token embeddings whose dimension $d$ is much smaller than the vocabulary size $V$. Existing analyses of training
dynamics either fix the embeddings with $d\ge V$ or train them only in restricted settings. It remains unclear whether jointly training the embeddings by gradient-based methods from random initialization leads to accurate prediction when $d<V$. In this paper, we study this question in the
simplest setting where it is already nontrivial, namely next-token prediction with a one-token context, which we refer to as a state-transition model. The model consists of an input embedding and an output embedding, either untied
or tied, and is trained from Gaussian initialization by perturbed gradient descent with weight decay on the population cross-entropy loss. We consider ground-truth transition laws whose centered log-probability table has low rank and bounded entries, and measure the prediction quality by the relative prediction gain (RPG). It is zero for the uniform predictor and one for the optimal predictor. For the untied model, we show that when $d=\Omega(\sqrt V)$, with probability at least $1-e^{-cd}$ for some constant $c$, the RPG is above $7/8$ at every iterate in the final $1-1/d$ fraction of training iterations. For the tied model trained on a subset of states, we prove that the RPG on the observed states is $1-O(V/d^2)$, and under a coverage condition on the ground truth and a bound on the learned embeddings, we further prove a prediction guarantee on the unseen states. Our analysis is based on the bounded trace of the Hessian of the softmax cross-entropy loss, which implies that the approximate second-order stationary points reached by training have high prediction gain. Numerical experiments with practical parameter choices illustrate the persistent training gain and the transfer of the tied model.

\end{abstract}

\section{Introduction}
\label{sec:iclr-introduction}
Next-token prediction (NTP) is the training objective of modern language models. A basic ingredient of such models is the learned token embedding~\citep{bengio2003neural}: each of the $V$ tokens in the vocabulary has an input embedding and an output embedding in $\R^d$, where $d$ is much smaller than $V$ and the two embeddings are often tied
\citep{press2017using,inan2017tyingwordvectorsword}. Recent theory has made substantial progress on the training dynamics of transformers for NTP and related sequence tasks, with a focus on how attention layers are learned. To make the analysis tractable, these works take the token embeddings as given and linearly independent, which requires $d \ge V $~\citep{li2024mechanics,huang2024nonasymptotic,huang2025regular,varre2025ngrams,yang2024cooccurrence}, or train them only in restricted settings, such as a binary vocabulary~\citep{makkuva2024attention} or a single gradient step~\citep{wu2026attention}, or characterize the global minimizers under $d\ge V$ without analyzing the training~\citep{zhao2024implicitgeometry}. As a result, it remains open whether training the embeddings by gradient-based methods from random initialization leads to accurate prediction when $d<V$.

In this paper, we study this question in a nontrivial simplified setting, namely NTP with a one-token context. In this case, the
model has no attention, and consists of an input embedding $\phi(s)\in\R^d$ of the current token $s$ and an output embedding
$\psi(s')\in\R^d$ of the next token $s'$. The predicted distribution is
$\widehat P(s'\mid s)\propto\exp(\phi(s)^\top\psi(s'))$. In other words, the model learns a state-transition law $P(s'\mid s)$ over $V$ states with $2Vd$ parameters, or $Vd$ parameters when the two embeddings are tied. We thus refer to it as a \textbf{state-transition model}. Although the model is simple, its training is far
from trivial: the loss is nonconvex because the two embeddings are jointly trained with $d<V$, and random initialization provides no alignment with the
ground truth. Moreover, since embeddings are part of every language model, these difficulties are also present in models with attention. Resolving
them is a necessary step towards analyzing such models. This motivates the following question:
\begin{center}
\emph{Can low-dimensional embeddings, jointly trained from random initialization,\\ accurately predict the next state (token)?}
\end{center}

\begin{wrapfigure}{r}{0.5\textwidth}
\centering
\vspace{-10pt}
\includegraphics[width=\linewidth]{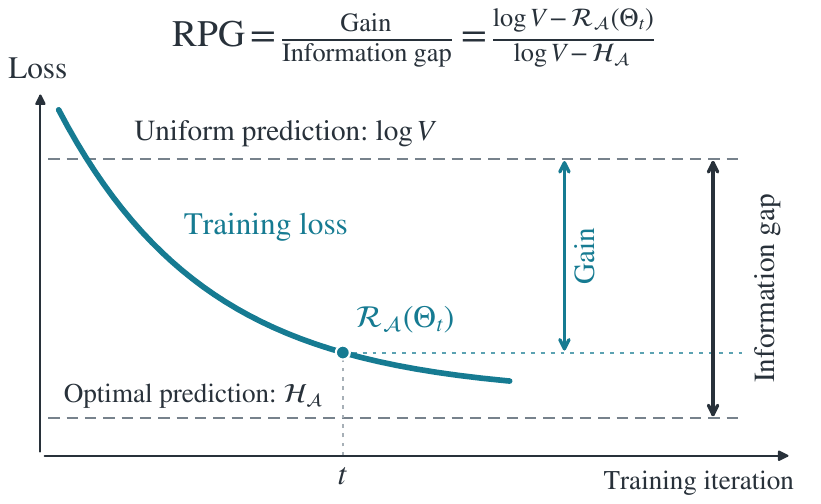}
\caption{Illustration of relative prediction gain (RPG).}
\label{fig:rpg}
\end{wrapfigure}
To answer this question, we must specify what successful training
means. Since the embeddings can be transformed without changing the predictions, they are not identifiable, and we measure the success of training by the prediction quality of the trained model. A stationary point may not give good predictions, because even a binary Markov-transformer model can converge to local minima depending on the data and initialization
\citep{makkuva2024attention}. Conversely, a model can be useful without attaining the optimal loss. We therefore measure prediction quality
by relative prediction gain (RPG; \Cref{fig:rpg}): the fraction of the uniform predictor's excess loss removed by training, with the true transition law as the reference optimum. Uniform prediction has gain zero, the true law has gain one. RPG is defined for any set of current states, so we can also study the prediction on unseen states. In the tied model, an unseen state's embedding is still updated through its role as an output label, providing a mechanism for transfer.

Our contributions are as follows.
\begin{itemize}[leftmargin=*]
\item \textbf{Untied training: high gain with sublinear embedding dimension.}
For every ground truth in our full-support, fixed-rank class, we jointly
train both embeddings from Gaussian initialization using
cross-entropy, weight decay, and fixed-step perturbed gradient
descent. When the embedding dimension satisfies $d\in[C\sqrt V, V)$, the prediction gain exceeds $7/8$
with probability at least $1-e^{-cd}$ at \emph{every} iterate in
$[T/d,T]$, for a polynomial number of iterations $T$ fixed in advance.
Thus, the guarantee holds for the whole final window of the training
trajectory.

\item \textbf{Tied training and transfer.}
When only $N=\Theta(V)<V$ current states are observed, jointly
trained tied embeddings achieve observed-state gain
$1-O(V/d^2)$ throughout the same final window at a comparable
embedding dimension. 
Tying also updates unseen-state embeddings through their role as
output labels. With a larger embedding dimension and a different
weight decay, a coverage condition, and a fixed unseen-state
information gap, we obtain gain
$1-O\!\left(V^4/[Nd^3(V-N)]\right)$ on the \emph{entire} unseen set
at every iterate in the final window whose learned embeddings have
uniformly bounded norms. 

\item \textbf{Proof technique.}
The softmax cross-entropy has limited total curvature across logit directions. By rotating the low-rank ground truth among embedding directions,
we find one where that curvature is small. In the region reached by
training, a low-gain point with a small gradient must therefore have
a direction of negative curvature. We then control the training steps
between repeated visits to approximate stationary points, extending
high gain from those visits to most training steps.
\end{itemize}
We also conduct synthetic experiments to show that high training gain persists over long runs and that tied embeddings can also predict well on unseen states. These patterns remain visible under sampled-transition training, although our theorem does not cover that setting.

\section{Related Work}
\paragraph{Low-rank softmax optimization.}
The softmax bottleneck concerns the representational capacity of
low-rank logits \citep{yang2018softmax,ganea2019softmax}. We assume a
realizable, fixed-rank ground truth and ask whether training reaches
useful predictions. For squared-loss matrix factorization, vanilla gradient descent converges globally from a sufficiently small random initialization when the factor width equals the target rank \citep{ye2021asymmetric}. Global-landscape results for deep linear networks
\citep{laurent2018deep}, low-rank recovery \citep{ge2017nospurious},
and factored nuclear-norm regularization \citep{li2017factorednuclear}
use width or loss-curvature conditions specific to their settings.
\citet{ouyang2025burer} give more general connections between factor
second-order stationarity and the nuclear-norm-regularized problem.
\citet{guo2008exponential} study low-rank exponential-family models,
including multinomial data, through a global-optimization
reformulation rather than the dynamics of the original factor updates. Related recommender-system work establishes local convergence of projected factor updates from a warm start \citep{sankagiri2025recommendations}, or statistical guarantees for convex multinomial-logit estimators \citep{negahban2018learning}.
For our loss, we use the trace of its Hessian to guarantee
\emph{prediction gain} at approximate stationary points. A direct uniform-curvature comparison and the trace-based refinement are given
in Appendix~\ref{app:comparison}. Neural-collapse analysis also studies
jointly learned features and classifiers under cross-entropy and decay,
but with one-hot targets and a different feature-width regime
\citep{zhu2021neuralcollapse}.

\paragraph{Perturbations and weight tying.}
Perturbation methods can reach approximate second-order stationary
points in nonconvex objectives \citep{ge2015escaping,jin2019nonconvex}.
Closer to factor training, \citet{liu2021noisyfactorization} analyze
noisy gradient descent for matrix factorization under squared loss,
with a finite-window guarantee in the rank-one case. Their noise
perturbs the factors before gradient evaluation, and their analysis
does not address the cross-entropy loss or transfer to unseen
states.
We use the always-perturbed algorithm and descent concentration bound
of \citet{jin2019nonconvex}; our model-specific steps turn repeated
visits into gain for every iterate in a final time window. Weight
tying shares input and output embeddings in language models
\citep{press2017using,inan2017tyingwordvectorsword}. Our tied training
bound needs no coverage assumption. The unseen-state bound additionally
requires centered coverage and, for the complete unseen set, a bound
on the learned embedding rows.

\paragraph{Next-token prediction.}
\if{0} 
\citet{li2024mechanics} characterize what a self-attention layer learns
from next-token prediction, while \citet{huang2024nonasymptotic} give
non-asymptotic convergence and a dataset-shift result for a two-stage
transformer training procedure. \citet{thrampoulidis2024implicit}
study the implicit bias of a linear next-token predictor with fixed
context embeddings and sparse conditional laws. Training dynamics for
regular-language recognition and in-context $n$-grams are studied by
\citet{huang2025regular} and \citet{varre2025ngrams}, respectively.
Our model omits attention and instead learns the factors of a
full-support state-transition table jointly. We seek finite-time
prediction gain along an interval of the training trajectory, rather
than a characterization of attention or implicit bias. In a distinct
binary Markov-transformer setting, \citet{makkuva2024attention} show
that initialization and data properties can lead to global or local
minima; their reduced parameterization is different from the
growing-state factor model studied here.
\fi

Our work is motivated by the theory of next-token prediction, which studies attention-based models
\citep{li2024mechanics,huang2024nonasymptotic}, linear predictors with fixed context embeddings \citep{thrampoulidis2024implicit}, and structured language or binary Markov tasks \citep{huang2025regular,varre2025ngrams,makkuva2024attention}.
Many of these works include attention and longer contexts, but fix the token embeddings or train them only in restricted settings \citep{wu2026attention,zhao2024implicitgeometry}. In contrast, we jointly train the input and output embeddings with $d<V$ from Gaussian initialization, and prove that high prediction gain is maintained throughout a
final window of training. A more detailed comparison is listed in \Cref{tab:related-work-comparison}.

\begin{table}[h]
\caption{Ingredients explicitly analyzed in representative results:
factorized optimization \citep{li2017factorednuclear,ouyang2025burer},
noisy matrix factorization \citep{liu2021noisyfactorization},
neural collapse \citep{zhu2021neuralcollapse},
weight tying \citep{press2017using,inan2017tyingwordvectorsword},
binary Markov transformers \citep{makkuva2024attention}, and
perturbed optimization \citep{jin2019nonconvex}.
``Partial'' means that a related ingredient is analyzed in a different
setting, i.e., weight tying and binary Markov transformers analyze
related factorized losses, and noisy matrix factorization perturbs the factors before the gradient evaluation, while we add noise to the
gradient.
The last row summarizes our untied and tied theorems together.}
\centering
\small
\setlength{\tabcolsep}{3pt}
\begin{tabular}{@{}p{0.255\linewidth}*{5}{>{\centering\arraybackslash}p{0.13\linewidth}}@{}}
\toprule
Work & Factorized loss analysis & Tied input/output & Full-support transition law & Persistent-noise analysis & Gain on unseen states \\
\midrule
Factorized optimization
& Yes & No & No & No & No \\
Noisy matrix factorization
& Yes & No & No & Partial & No \\
Neural collapse
& Yes & No & No & No & No \\
Weight tying
& Partial & Yes & No & No & No \\
Binary Markov transformers
& Partial & Yes & Yes & No & No \\
Perturbed optimization
& No & No & No & Yes & No \\
 \textbf{This work}  
& \textbf{Yes} & \textbf{Yes} & \textbf{Yes} & \textbf{Yes} & \textbf{Yes} \\
\bottomrule
\end{tabular}
\label{tab:related-work-comparison}
\end{table}

\section{Problem Setup}
\label{sec:iclr-setup}

Let $[V]=\{1,\ldots,V\}$ be the state space. The ground-truth transition
table $Q\in\R^{V\times V}$ has all positive entries and row sums all equal 1; row
$q_i^\top$ is the distribution of the next state given current state
$i$. All softmax operations below act row-wise. The untied model predicts
$P^{E,U}=\softmax(EU^\top)$, with $E,U\in\R^{V\times d}$, where the $i$-th rows of $E$ and $U$ are the input embedding $\phi(i)$ and the output embedding $\psi(i)$ of state $i$, respectively.
The tied model predicts $P^X=\softmax(XX^\top)$, with
$X\in\R^{V\times d}$.
We write $\Theta=(E,U)$ or $\Theta=X$ accordingly. Thus the numbers of
trainable parameters are $2Vd$ and $Vd$. Throughout, the embedding dimension satisfies $d<V$.
We use calligraphic letters for sets, except for standard spaces and $[V]$, and uppercase letters for matrices.

For a nonempty set $\mathcal A\subseteq[V]$ of current states, define population and Bayes (optimal) risk:
\begin{align*}
 \cR_{\mathcal A}(\Theta)=-\frac1{|\mathcal A|}\sum_{i\in\mathcal A}\sum_{j=1}^Vq_{ij}\log P^\Theta_{ij},
 \qquad \cH_{\mathcal A}=-\frac1{|\mathcal A|}\sum_{i\in\mathcal A}\sum_{j=1}^Vq_{ij}\log q_{ij}.
\end{align*}
Note that the risk \(\mathcal R_{\mathcal A}\) assigns uniform weights to the current states in \(\mathcal A\). Although we call it the population risk, \(Q\) is only required to be a fixed transition table that satisfies our assumptions. For instance, \(Q\) can be the empirical transition table of a dataset, in which case \(\mathcal R_{\mathcal A}\) is the full-batch training loss with uniform weights on the current states. 

\begin{definition}[Relative prediction gain (RPG)]
When $\log V-\cH_{\mathcal A}>0$, define
\begin{align*}
 g_{\mathcal A}(\Theta)=\frac{\log V-\cR_{\mathcal A}(\Theta)}{\log V-\cH_{\mathcal A}}
 =1-\frac{|\mathcal A|^{-1}\sum_{i\in\mathcal A}\KL(q_i\|P^\Theta_{i,:})}{\log V-\cH_{\mathcal A}}.
\end{align*}
\end{definition}
The uniform predictor has gain zero and the Bayes predictor has gain
one. For example, gain $7/8$ means that training removes seven eighths of the uniform predictor's excess risk.

\paragraph{Ground-truth class.}
Adding the same constant to every logit in a row leaves its softmax
unchanged. For any real matrix $Z$ with $V$ columns, define its
row-centering by
\(
 \mathcal{L}(Z)=Z\left(I_V-\frac1V\1\1^\top\right).
\)
Thus the ground truth's centered logits are $\mathcal{L}(\log Q)$, where the
logarithm is entrywise. We consider the class
\begin{align*}
 \mathcal Q_{r,B,\Delta_0}
 =\left\{Q\in\mathbb{R}^{V\times V}:\begin{array}{l}
 Q_{ij}>0,\ Q\1=\1,\ \rank \mathcal{L}(\log Q)\le r,\\
 \|\mathcal{L}(\log Q)\|_\infty\le B,\ \log V-\cH_{[V]}\ge\Delta_0
 \end{array}\right\}.
\end{align*}
Here $r$ is a fixed positive integer and $B,\Delta_0>0$ are fixed
independently of $V$. The matrix infinity norm is the maximum absolute
entry. We impose the rank condition on the centered logits, because centering does not change the model or the risk, and the rank determines whether the model can represent the ground truth. In particular, when $r\le d$, the ground truth is representable by the untied model.
The bound on the logits keeps every transition probability positive. The last condition, which lower bounds the information gap $\log V-\cH_{[V]}$, excludes a trivial uniform prediction problem.

\paragraph{Training algorithm.}
All entries of $\Theta_0$ are independent $\mathcal N(0,1/d)$ variables.
We train all entries simultaneously on the weight-decayed objective and use perturbed gradient descent~\citep{jin2019nonconvex} as follows.
\begin{align*}
 &\cF_{\lambda,\mathcal A}(\Theta)
 =\cR_{\mathcal A}(\Theta)+\frac\lambda2\|\Theta\|_\fro^2,
 \\
 &\Theta_{t+1}=\Theta_t-h(\nabla \cF_{\lambda,\mathcal A}(\Theta_t)+\xi_t),
 \qquad \xi_t\stackrel{\rm iid}{\sim}
 \mathcal N\left(0,\frac{\sigma^2}{D_\Theta}I_{D_\Theta}\right).
\end{align*}
Here $D_\Theta$ is the number of parameters and
$\|(E,U)\|_\fro^2=\|E\|_\fro^2+\|U\|_\fro^2$.
The perturbations are independent of initialization. The weight decay
$\lambda$, step size $h$, perturbation scale $\sigma$, and number of iterations $T$
are deterministic choices made before initialization. The step size and
the perturbation scale stay fixed throughout the run. Unless stated otherwise, constants in our bounds may depend on the fixed assumptions of the corresponding theorem, but are uniform in \(V,N,d\), and \(T\).

\paragraph{Technical challenges.}
As discussed in \cref{sec:iclr-introduction}, the difficulty comes from the combination of \textit{$d<V$, cross-entropy loss, Gaussian initialization,
and simultaneous gradient-based training of all embeddings}. Specifically, (i) the low-dimensional
embeddings make the optimization nonconvex; (ii) the Hessian of the
cross-entropy loss depends on the learned probabilities and can become
nearly singular; and (iii) random initialization
provides no alignment with the ground truth and no local convergence guarantee. Therefore, analyses based on fixed embeddings, staged training, or convex logit reformulations~\citep{thrampoulidis2024implicit,huang2024nonasymptotic,zhao2024implicitgeometry}
do not directly apply to our setting. Moreover, the generic saddle-escaping analysis~\citep{jin2019nonconvex} only guarantees that the algorithm visits approximate second-order stationary points, which does not imply accurate prediction by itself. To obtain a prediction guarantee, we need to (a) relate stationarity to low prediction loss in the
region that is actually reached by training, and (b) control the iterates between consecutive visits to such stationary points.

\section{Training Gain of Untied Model}
\label{sec:iclr-training}

In this section, we study the untied model trained on all current states, i.e.,
$\mathcal A=[V]$. We are interested in whether an embedding dimension much smaller than $V$ can support useful prediction over many consecutive training steps, despite the nonconvex objective and perturbations at every update. The theorem below gives a finite-time answer: with high probability, the training achieves at least 7/8 RPG within sufficiently long intervals. Roughly speaking, the training loss is ``close'' to the optimal loss, and therefore the prediction is accurate.

\begin{theorem}[Untied constant gain]
\label{thm:iclr-untied}
Fix $r,B,\Delta_0$. There are positive constants $c,C,V_0$ such that,
for every $V\ge V_0$, $Q\in\mathcal Q_{r,B,\Delta_0}$, and
$\max\{2r,C\sqrt V\}\le d<V-1$, setting the weight decay
\(
 \lambda=\frac{\Delta_0}{8B\sqrt r\,V}
\), there exist a step size $h$, a perturbation scale $\sigma$, and a number of iterations $T$ such that
\begin{equation*}
 \Pr\left\{g_{[V]}(\Theta_t)>\frac78
 \text{ for every integer }t\in[T/d,T]\right\}\ge1-e^{-cd}.
\end{equation*}
The quantities $h^{-1},\sigma^{-1}$, and $T$ are bounded by
polynomials in $V$. The probability is over both initialization and
perturbations.
\end{theorem}


\Cref{thm:iclr-untied} states that, with high probability, every iterate
after the first $1/d$ fraction of training, including the final one,
has RPG above $7/8$. Moreover, an embedding dimension of order
$\sqrt V$, which is much smaller than $V$, is sufficient.

\subsection{Proof Overview}
\label{sec:untied-overview}
\begin{definition}[Approximate SOSP~\citep{jin2019nonconvex}]
Let $f$ be an objective whose Hessian is $\rho$-Lipschitz on the region under consideration. We call $\Theta$ an $\epsilon$-approximate second-order stationary point (SOSP)
of $f$ if
\[
 \|\nabla f(\Theta)\|\le\epsilon,
 ~\text{and}~
 \nabla^2 f(\Theta)\succeq-\sqrt{\rho\epsilon}\,I.
\]
Throughout, we take $f=\cF_{\lambda,[V]}$, i.e., the training objective.
\end{definition}

\begin{wrapfigure}{r}{0.5\textwidth}
\centering
\vspace{-10pt}
\includegraphics[width=\linewidth]{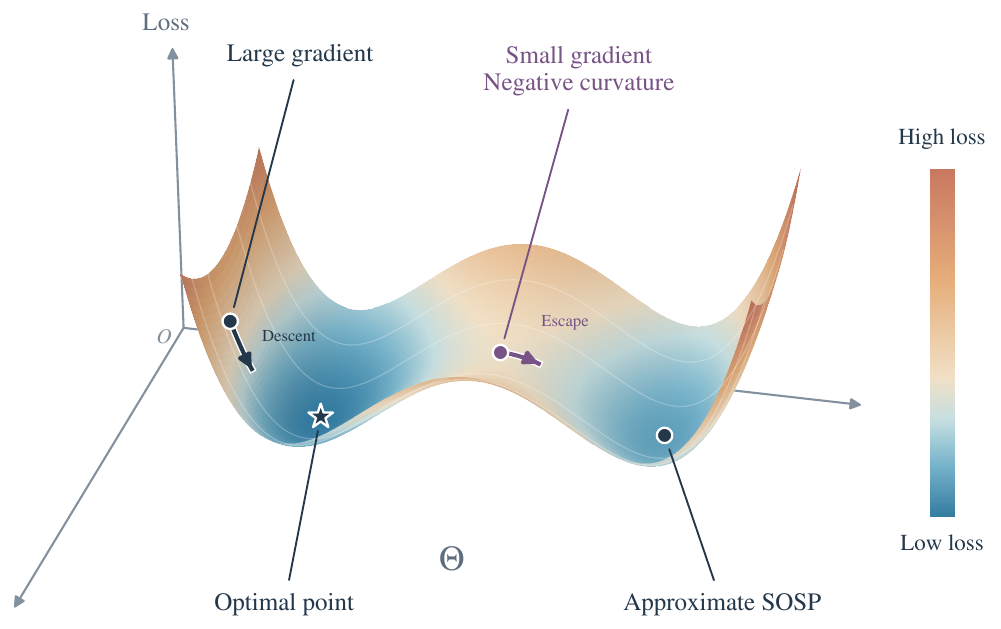}
\caption{Illustration of an approximate SOSP. Our Step 2 shows that all approximate SOSPs ``of interest'' must have low loss.}
\label{fig:landscape}
\end{wrapfigure}
At a high level, the goal is to prove that the dynamics of $\Theta$ will enter into a low-loss region. The difficulty is that $\Theta$ may fall into an approximate SOSP. Because otherwise, either the gradient $\nabla \cF(\Theta)$ is sufficiently large to make the loss decrease, or the Hessian $\nabla^2\cF(\Theta)$ has a sufficiently negative eigenvalue that allows $\Theta$ to escape. To handle this difficulty, the key idea is to show that approximate SOSPs ``of interest'' have high prediction gain. To this end, we construct three main steps. First, we establish that the training trajectory of $\Theta$ remains in a ``good'' region, namely localize the training trajectory. Second, we show that in this region, every approximate SOSP must have high prediction gain. Finally, the conclusion follows from combining repeated approximate-SOSP visits and induction on the error control. Appendix~\ref{sec:positive} gives the parameter choices and complete proof.



\paragraph{Step 1: Localize the training trajectory.}
Positive weight decay creates a region where the squared parameter norm is at most \(O(V)\). Leaving this region requires a substantial increase in the objective loss. 
More importantly, concentration inequalities guarantee that, with high probability, Gaussian initialization lies inside this region (\cref{lem:untied-gaussian-initialization}), 
and the perturbed gradient descent ensures that the objective cannot rise by more than a fixed small amount during training. 
Therefore, with high probability, the training trajectory remains in the region where the parameter norm is bounded.

\paragraph{Step 2: Convert approximate stationarity into gain.}
In this step, we show that every $\epsilon$-approximate SOSP $\Theta$ in the localized region satisfies $\cF(\Theta)<\cH_{[V]}+\Delta_0/8-2b$ for some constant $b>0$, which implies $g_{[V]}(\Theta)>7/8$. To this end, we first construct a comparator $\Theta_\star$, whose logits are the slightly shrunken ground-truth logits, such that $\|\Theta_\star\|_\fro^2=O(V)$ and $\cF(\Theta_\star)\le\cH_{[V]}+\Delta_0/8-4b$ (\cref{lem:shrunken-comparator}). Suppose, for contradiction, that $\cF(\Theta)\ge\cH_{[V]}+\Delta_0/8-2b$, so that $\cF(\Theta)-\cF(\Theta_\star)\ge2b$. The key step lies in analyzing the Hessian $\nabla^2\cF(\Theta)$ along a direction $\dot\Theta$ obtained by rotating $\Theta_\star$ among the $d$ embedding directions, which satisfies $\|\dot\Theta\|_\fro^2=O(V)$ and (\cref{lem:fisher-packing,lem:highgain-collar})
\[
 \nabla^2\cF(\Theta)[\dot\Theta,\dot\Theta]
 \le\underbrace{O\Bigl(\frac{V}{d^2}+\frac1d\Bigr)}_{\le b\text{ for }d\ge C\sqrt V}
 +\underbrace{\langle\nabla\cF(\Theta),\Theta\rangle}_{\le\epsilon\|\Theta\|_\fro\le b}
 -2\underbrace{\bigl(\cF(\Theta)-\cF(\Theta_\star)\bigr)}_{\ge2b}
 \le-2b.
\]
Here, the first term bounds the curvature of the softmax, using that the Hessian of the loss of each row with respect to its logits has trace at most one, and the last two terms follow from the convexity of the loss in the logits and $\|\Theta\|_\fro=O(\sqrt V)$. Since $\|\dot\Theta\|_\fro^2=O(V)$, we obtain $\lambda_{\min}(\nabla^2\cF(\Theta))\le\nabla^2\cF(\Theta)[\dot\Theta,\dot\Theta]/\|\dot\Theta\|_\fro^2\le-2b/\|\dot\Theta\|_\fro^2\le-c/V$ for some constant $c>0$. However, an $\epsilon$-approximate SOSP satisfies $\lambda_{\min}(\nabla^2\cF(\Theta))\ge-\sqrt{\rho\epsilon}$, where $\sqrt{\rho\epsilon}\le c/(2V)$ since $\rho=O(\sqrt V)$ and $\epsilon$ is a sufficiently small constant times $V^{-5/2}$. This is a contradiction.

\paragraph{Step 3: Refresh the gain throughout the final window.}
Finally, we extend the guarantee in Step 2 from approximate SOSPs to all iterates in the final window. We partition the $T$ iterations into $d$ blocks, which are only used in the analysis. By the result of \citet{jin2019nonconvex} on perturbed gradient descent, with high probability, every block contains an \(\epsilon\)-approximate SOSP, and the objective increases by at most $b$ within each block, i.e., $\cF(\Theta_v)-\cF(\Theta_u)\le b$ for any $u<v$ in the same block (\cref{thm:tied-fixed-noise-consequence}). Combining this with Step 2 and using induction over the blocks, we obtain $\cF(\Theta_t)<\cH_{[V]}+\Delta_0/8$, and hence $g_{[V]}(\Theta_t)>7/8$, for every integer
$t\in[T/d,T]$.


\section{Tied Model Training and Conditional Transfer}
\label{sec:iclr-generalization}

In this section, we study the training dynamics when the input and output embeddings are tied. Let $\mathcal S\subset[V]$ be the observed current states, with
$N=|\mathcal S|=\Theta(V)$, and let
$\mathcal U=[V]\setminus\mathcal S$ be the full unseen set. We assume $d<N<V$. Every state remains a possible output label during training, including the states in $\mathcal U$. Hence, tying causes the embedding of an unseen input state to be updated through its output role. We establish two results: a training 
guarantee on the observed states and, under additional conditions, a  prediction guarantee on the complete unseen set.

\subsection{Tied Model Training}

\begin{definition}[Tied-compatible base class]
\label{def:iclr-tied}
For fixed $B_Y\ge1$, let $\mathcal Q^{\rm tied}_{r,B,\Delta_0}$
consist of tables $Q\in\mathcal Q_{r,B,\Delta_0}$ that admit
\(
 Q=\softmax(YY^\top)
\)
for some $Y\in\R^{V\times r_Y}$ with $r_Y\le r$ and
$\max_i\|y_i\|_2^2\le B_Y$, where $y_i$ is the column-vector form of
row $i$ of $Y$.
\end{definition}


Tying the input and output embeddings restricts which transition
laws the model can represent. For the class above, the next theorem
shows that training achieves a gain close to one on the observed
states throughout a long final portion of the run.

\begin{theorem}[Tied training]
\label{thm:iclr-tied-training}
Let $Q\in\mathcal Q^{\rm tied}_{r,B,\Delta_0}$ and suppose
$\log V-\cH_{\mathcal S}\ge\delta_{\rm tr}>0$. For sufficiently
large $V$ and
\(
 \max\{2r,C\sqrt V\}\le d<N,
\)
setting the weight decay $\lambda=C_\lambda/d^2$, there exist a step size $h$, a perturbation scale $\sigma$, and a number of iterations $T$ divisible by $d$ such that, with probability at least
$1-e^{-cd}$, \( g_{\mathcal S}(X_t)
 \ge1-\frac{CV}{d^2\delta_{\rm tr}}\), for every integer $t\in[T/d,T]$.
The quantities $h^{-1},\sigma^{-1}$, and $T$ are bounded by polynomials in $V$.
\end{theorem}

\Cref{thm:iclr-tied-training} indicates that the gap from gain one scales
as $V/d^2$, so the guarantee strengthens as the embedding dimension
grows beyond $\Omega(\sqrt V)$.

\subsubsection{Proof Overview}
\label{sec:tied-overview}
The proof follows the similar structure as the untied argument
in \cref{sec:iclr-training}, i.e., we localize the trajectory, obtain a loss guarantee at approximate SOSPs,
and then propagate this guarantee throughout the final window.
The key difference is that the weight decay absorbs a curvature term arising from the tied embeddings, which depends on the norm $\|X\|_\fro^2$ of the embeddings, referred to as their \emph{energy}.

\paragraph{Step 1: Localize the training trajectory.}
Gaussian initialization bounds the initial objective.
Since cross entropy is bounded below by $\cH_{\mathcal{S}}$, the penalty creates an
objective barrier outside a deterministic ball. Modifying the objective
outside this ball provides the global smoothness bounds required by the analysis of perturbed gradient descent.
The control of the objective increase keeps the modified and original training trajectories inside the ball
and identical under the same noise (\cref{lem:tied-localization-inputs}).

\paragraph{Step 2: Convert approximate stationarity into gain.}
We show that every sufficiently accurate approximate SOSP $X$ in the ball satisfies $\|X\|_\fro^2=O(V)$ and $\cF_{\lambda,\mathcal S}(X)-\cH_{\mathcal S}=O(V/d^2)$. Similarly to Step 2 in \cref{sec:untied-overview}, the key step lies in analyzing the Hessian $\nabla^2\cF_{\lambda,\mathcal S}(X)$ along a direction that represents the ground-truth Gram matrix $YY^\top$ and has small softmax curvature. Together with the convexity of the loss in the Gram matrix, this gives
$\cF_{\lambda,\mathcal S}(X)\lesssim\cH_{\mathcal S}+\frac\lambda2\|Y\|_\fro^2+C\|X\|_\fro^2/d^2$
up to small stationarity errors (\cref{lem:tied-teacher-packing}), where the last term depends on the energy of $X$. Since $\cF_{\lambda,\mathcal S}(X)\ge\cH_{\mathcal S}+\frac\lambda2\|X\|_\fro^2$, choosing $\lambda=C_\lambda/d^2$ with a sufficiently large $C_\lambda$ lets the weight decay penalty absorb this term, which yields $\|X\|_\fro^2=O(V)$, since $\|Y\|_\fro^2=O(V)$, and then the claim (\cref{lem:tied-sosp-energy}).

\paragraph{Step 3: Refresh the gain throughout the final window.}
Finally, as in Step 3 of \cref{sec:untied-overview}, we partition the iterations into $d$ blocks. With high probability, every block contains an accurate approximate SOSP, and the objective increases by at most $V/(4d^2)$ within each block. Combining this with Step 2 and using induction over the blocks, we obtain $\cR_{\mathcal S}(X_t)-\cH_{\mathcal S}\le\cF_{\lambda,\mathcal S}(X_t)-\cH_{\mathcal S}=O(V/d^2)$ for every integer $t\in[T/d,T]$.
Dividing by the training information gap gives \cref{thm:iclr-tied-training}.
Appendix~\ref{app:tied-proofs} provides the complete argument.




\subsection{Conditional Transfer to Unseen States}

Training on $\mathcal S$ does not by itself determine predictions on
$\mathcal U$. We therefore impose a coverage condition on the
ground-truth representation.

\begin{definition}[Coverage for transfer]
\label{def:iclr-transfer-coverage}
A representative $Y$ from \cref{def:iclr-tied} satisfies centered
coverage over $\mathcal S$ if it has linearly independent columns and 
for some fixed $\kappa>0$, we have
\begin{align*}
 Y^\top\1=0,
 \qquad
 \frac1N\sum_{i\in\mathcal S}(y_i-\bar y_{\mathcal S})
 (y_i-\bar y_{\mathcal S})^\top\succeq\kappa^2 I,
 \qquad
 \bar y_{\mathcal S}=\frac1N\sum_{i\in\mathcal S}y_i.
\end{align*}
\end{definition}


The coverage condition says that the observed ground-truth embeddings
capture every ground-truth feature direction. This lets information
from the observed rows constrain predictions when unseen states appear
as inputs. The theorem below uses a {weight decay} chosen for this
transfer argument and requires a larger embedding dimension than tied
training alone.

\begin{theorem}[Conditional tied transfer]
\label{thm:iclr-transfer}
Let $Q\in\mathcal Q^{\rm tied}_{r,B,\Delta_0}$ admit a representative
$Y$ satisfying \cref{def:iclr-transfer-coverage}. Suppose
\(
 \log V-\cH_{\mathcal S}\ge\delta_{\rm tr}>0,\) and \(
 \log V-\cH_{\mathcal U}\ge\delta_{\rm te}>0.
\)
For sufficiently large $V$, let
\(
 \max\{2r,CV/N^{1/3}\}\le d<N,\) and \(
 \lambda=C_\lambda\sqrt{\frac{V}{Nd^3}}.
\)
Then, there exist a step size $h$, a perturbation scale $\sigma$, and a number of iterations $T$, with $h^{-1},\sigma^{-1}$, and $T$ bounded by polynomials in $V$, such that, with probability at
least $1-e^{-cd}$, every integer $t\in[T/d,T]$ satisfies
\begin{align*}
 g_{\mathcal S}(X_t)
 \ge1-\frac{CV^3}{Nd^3\delta_{\rm tr}}.
\end{align*}
Moreover, at every such iterate satisfying \( \max_{j\in[V]}\|x_j(t)\|_2^2\le R_0\)
for a fixed $R_0$, the complete unseen set satisfies
\begin{align*}
 g_{\mathcal U}(X_t)
 \ge1-\frac{CV^4}{Nd^3(V-N)\delta_{\rm te}}.
\end{align*}
\end{theorem}

\Cref{thm:iclr-transfer} bounds the gain on the observed states
throughout the final window and, at any of those iterates with
uniformly bounded learned embeddings, also the gain on the complete
unseen set. The bound on the learned embeddings is an additional
condition that our optimization analysis does not establish, so the
conclusion on unseen states is conditional.

\subsubsection{Proof Overview}
\label{sec:transfer-overview}

The proof consists of three main steps. The key idea is that, since the Gram matrix $XX^\top$ is positive semidefinite, its rows indexed by the observed states together with its trace control the entire matrix.
First, we compare the objective at every iterate in the final window with the objective at the ground truth.
Second, we use the coverage condition to turn this comparison into bounds on the training loss and on the trace of the Gram matrix.
Finally, under the bound on the learned embeddings, these two bounds control the entire Gram matrix, which
yields the gain bound on the unseen states. Appendix~\ref{app:transfer-proof} gives the parameter choices and complete proof.

\paragraph{Step 1: Compare the objective with the ground truth in the final window.}

We show that, with high probability, every iterate $X_t$ in the final window satisfies $\cF_{\lambda,\mathcal S}(X_t)\le\cH_{\mathcal S}+\frac\lambda2\|Y\|_\fro^2+C(\lambda V)^2$, where the first two terms are the objective at the ground truth, and $(\lambda V)^2$ is of order $V^3/(Nd^3)$ under the weight decay in \cref{thm:iclr-transfer}. This follows from the same three steps as in \cref{sec:tied-overview}. Specifically, by Step 2 there, every accurate approximate SOSP satisfies this comparison up to the energy-dependent term $C\|X\|_\fro^2/d^2$ (\cref{lem:tied-teacher-packing}), which is absorbed by the weight decay. Then, by Steps 1 and 3 there, the comparison extends to every iterate in the final window (\cref{lem:tied-final-window-comparison}).

\paragraph{Step 2: Bound the training loss and the trace of the Gram matrix.}
Let $K=XX^\top$ and $K^\star=YY^\top$ be the learned and ground-truth Gram matrices. By Step 1, every iterate $X$ in the final window satisfies $\{\cR_{\mathcal S}(X)-\cH_{\mathcal S}\}+\frac\lambda2\{\tr K-\tr K^\star\}\le C(\lambda V)^2$, which does not immediately bound the excess loss $\cR_{\mathcal S}(X)-\cH_{\mathcal S}$, because $\tr K-\tr K^\star$ can be negative. Using the coverage condition, we establish the following trace certificate (\cref{lem:tied-anchor-trace-certificate,lem:tied-kl-lone}):
\[
 \tr K-\tr K^\star\ge-CV\Bigl(\cR_{\mathcal S}(X)-\cH_{\mathcal S}+\sqrt{\cR_{\mathcal S}(X)-\cH_{\mathcal S}}\Bigr).
\]
Substituting it into the above inequality, and using that $\lambda V$ is small, we obtain $\cR_{\mathcal S}(X)-\cH_{\mathcal S}=O((\lambda V)^2)$ and $\tr K-\tr K^\star\le O(\lambda V^2)$.

\paragraph{Step 3: Transfer the bounds to unseen states.}
Finally, we transfer the bounds in Step 2 to the unseen states at any iterate $X$ in the final window with $\max_j\|x_j\|_2^2\le R_0$. In this case, all logits are bounded, so the softmax probabilities are of order $1/V$, and the excess loss of each row is equivalent, up to constants, to $1/V$ times the squared error of its centered logits. Hence, the training loss controls the error of $K$ on the observed rows. The remaining error consists of a part determined by the observed rows and a positive semidefinite residual, whose norm is at most its trace. This gives (\cref{lem:tied-psd-completion})
\[
 \|K-K^\star\|_\fro\lesssim\sqrt{NV\bigl(\cR_{\mathcal S}(X)-\cH_{\mathcal S}\bigr)}+\max\{\tr K-\tr K^\star,0\}=O(\lambda V^2).
\]
{For} the unseen rows, the same equivalence gives $\cR_{\mathcal U}(X)-\cH_{\mathcal U}\lesssim{\|K-K^\star\|_\fro^2/((V-N)V)}$, which is $O({(\lambda V)^2V/(V-N)})$ (\cref{lem:iclr-transfer-mechanism}). Substituting the choice of $\lambda$ and dividing by the training and test information gaps gives the two gain bounds in \cref{thm:iclr-transfer}.

\section{Numerical Experiments}
\label{sec:iclr-experiments}
In this section, we present two sets of experiments. The first set uses population gradients, and
examines the persistence of the training gain, the transfer of the tied model, and
the bounded-embedding condition in \cref{thm:iclr-transfer}. The second set uses
SGD with sampled transitions at a larger scale, and tests whether these behaviors
persist under stochastic training and across larger state spaces,
{ground-truth} ranks, and embedding {dimensions}. Both studies use
practical parameter values.

\subsection{Population(Full batch)-Gradient Training}

\begin{figure}[thb]
 \centering
 \includegraphics[width=0.90\linewidth]{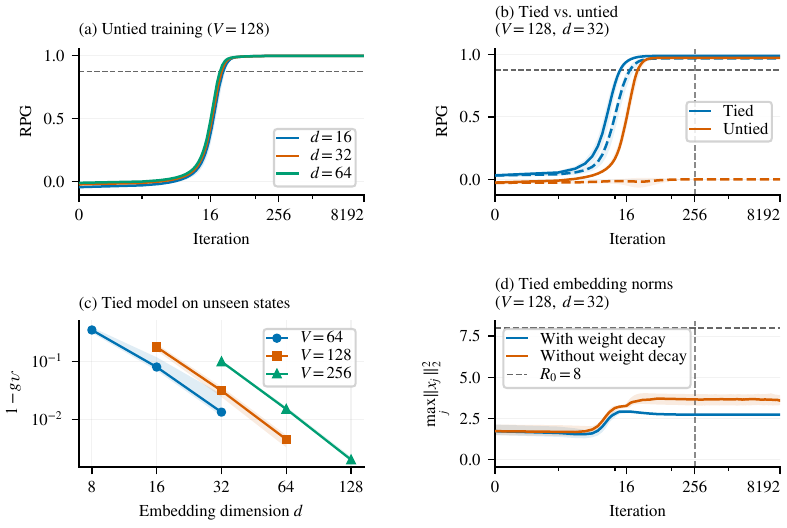}
 \caption{Training and transfer in the {population-gradient} experiments.
 Curves show medians and shaded full ranges over three ground truths
 and five initializations each. (a) Untied training gain at $V=128$.
 (b) Tied and untied models on the same observed/unseen split at
 $V=128,d=32$ (solid: training; dashed: test). (c) Final tied test deficit ($1-$ gain) as {the embedding dimension} and {the corresponding weight} decay change together. (d) Maximum tied embedding-row norm squared.
 The vertical lines in (b,d) mark $T/d$.
 No trajectories are smoothed; statistics use every update.}
 \label{fig:training-validation}
\end{figure}

We use ground truths with rank-four centered logits bounded by 3,
$V\in\{64,128,256\}$, and $d\in\{V/8,V/4,V/2\}$. Untied training
uses all {current} states; tied training uses $N=3V/4$ {observed} states. Every run
uses Gaussian initialization, full population gradients evaluated in FP32, fixed step
$h=V/4$, and $T=8192$ updates. The perturbation scale is fixed at
$\sigma=0.01/\sqrt V$.
We report all 375 runs, including paired no-noise and no-decay controls.
Appendix~\ref{app:experiments} gives the full protocol, data, and checks.

\paragraph{Main results.} We examine three behaviors predicted by our theorems.
(i) {\it The training gain remains high throughout the final window $[T/d,T]$} (Theorems 4.1 and 5.2). The gain stays above $7/8$ in all 135 runs of the untied model and in 120 of the 135 runs of the tied model. The exceptions are the runs with the smallest embedding dimension ($V=64$, $d=8$).
- (ii) {\it The tied model transfers to unseen states} (Theorem 5.4). When both models are trained on the same observed states, the final gain of the tied model on the unseen states is above $0.96$, while that of the untied model remains near zero.
- (iii) {\it The condition on the learned embeddings in Theorem 5.4 holds in the experiments.} The squared norms of the learned embeddings stay below the bound $R_0=8$, which is fixed in advance, in all runs of the tied model.

\if{0}
All 135 main runs of the untied model maintain gain above $7/8$ at every
$t\in[T/d,T]$; the minimum over these runs and times is $0.9951$.
For tied training, 120 of 135 main runs remain above the same $7/8$
reference level. The exception is the
setting with the smallest embedding dimension, $V=64,d=8$, whose final training gains range from
$0.8610$ to $0.8753$.
On the same observed/unseen split in \cref{fig:training-validation}(b), final tied
test gains range from $0.9638$ to $0.9758$, while the untied model's
unseen gain remains near zero. All tied runs satisfy the bound $\max_j\|x_j(t)\|^2\le8$, which was fixed in advance,
throughout the final window $[T/d,T]$.
These results show that high gain persists throughout late training, and that the tied model also predicts well on unseen current
states while satisfying the bounded-embedding condition in
\cref{thm:iclr-transfer}.
\fi
\paragraph{Effects of perturbation and weight decay.} The paired control runs show that removing the Gaussian gradient perturbations yields comparable gain, indicating that the strong performance on these instances does not rely on the injected
perturbations. Removing weight decay lowers final training prediction
error, reflecting a tradeoff between regularization and training fit.
In \cref{fig:training-validation}(c), the embedding dimension and the weight decay vary together, so the curves reflect their joint effect.

\subsection{Scaling Optimization with Sampled SGD}
\begin{figure}[h]
 \centering
 \includegraphics[width=\linewidth, trim=0 2mm 0 2mm, clip]{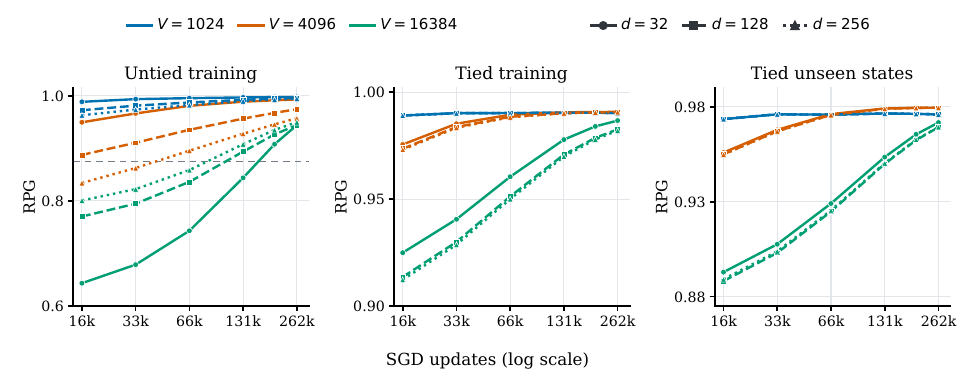}
 \caption{Relative prediction gain during sampled-transition SGD at
 ground-truth rank $r=8$. Each panel contains nine curves:
 $V\in\{1024,4096,16384\}$ (colors) and $d\in\{32,128,256\}$
 (line and marker styles). Lines join medians of three initialization
 seeds at each checkpoint, without smoothing. Training uses all
 current states in the untied model and $3V/4$ observed states in the
 tied model; tied unseen gain uses the remaining $V/4$ states. The
 horizontal axis is logarithmic and the RPG scales differ by panel.
 The dashed horizontal reference is $7/8$ wherever visible.}
 \label{fig:scaling-r8-trajectories}
\end{figure}

To test whether our qualitative findings extend beyond exact population
gradients, we evaluate sampled-transition training over larger state
spaces, ground-truth ranks, and embedding dimensions.
\Cref{fig:scaling-r8-trajectories} follows nine $(V,d)$ settings at
{ground-truth} rank $r=8$ from 16{,}384 to 262{,}144 SGD updates. {For larger $V$}, some untied runs begin well below $7/8$ but continue to
improve with training. By the final checkpoint, every one of the 78
$r=8$ runs exceeds $7/8$ RPG: the lowest individual gains are 0.934
for untied training, 0.982 for tied training, and 0.968 for tied unseen
states.At larger ground-truth ranks, some configurations remain below $7/8$ at the final checkpoint, but are
still improving. \Cref{app:scaling-sgd} gives
the full protocol and the wider configuration grid.

\section{Conclusion}
In this work, we investigated state-transition models, i.e., next-token prediction with a
one-token context, in which low-dimensional embeddings are jointly trained from random initialization. We
first used the bounded trace of the Hessian of the cross-entropy loss to show that
accurate approximate second-order stationary points reached by the
untied model have high prediction gain. Combining this result with
{the localization of the trajectory and repeated visits to such points}, we proved that perturbed gradient
descent with weight decay achieves constant gain in polynomial
time and maintains it throughout a {final window of training}. We then extended this guarantee to tied embeddings, and, under additional
conditions on the coverage and the learned embeddings, obtained a prediction
guarantee for the complete unseen set. Our experiments illustrate
the persistence of the training gain and the transfer benefit of tying. An important future direction is to extend our analysis to models with attention.

\subsection*{AI use statement}




In this work, we used generative AI tools to assist in proving mathematical claims, including checking and refining intermediate proof steps. We did not use generative AI tools for other research tasks requiring disclosure where applicable. Additionally, we used generative AI tools to aid and polish the writing of the manuscript, including improving clarity, grammar, and presentation.

All AI-assisted mathematical reasoning was independently reviewed and verified by the authors, and all AI-assisted writing was reviewed and edited to ensure that it accurately reflects the authors' intended arguments and findings. We take responsibility for the final content of this work, including all text, mathematical claims, proofs, and other artifacts produced with the aid of generative AI.




\subsection*{Reproducibility statement}
The main text states the model assumptions and training algorithms,
and the appendices provide complete proofs and parameter choices.
\Cref{app:experiments} and \cref{app:scaling-sgd}
document the synthetic data construction, experimental protocols,
and evaluation procedures. The accompanying \texttt{experiments/}
directory provides the code, dependencies, fixed configurations,
and instructions to rerun the experiments and regenerate the
reported figures and table. No external dataset is required.

\bibliography{references}

@article{wu2026attention,
  title={Attention with Trained Embeddings Provably Selects Important Tokens},
  author={Wu, Diyuan and Shevchenko, Aleksandr and Oymak, Samet and Mondelli, Marco},
  journal={Advances in Neural Information Processing Systems},
  volume={38},
  pages={28956--28997},
  year={2026}
}

@article{negahban2018learning,
  title={Learning from comparisons and choices},
  author={Negahban, Sahand and Oh, Sewoong and Thekumparampil, Kiran K and Xu, Jiaming},
  journal={Journal of Machine Learning Research},
  volume={19},
  number={40},
  pages={1--95},
  year={2018}
}

@article{sankagiri2025recommendations,
  title={Recommendations with Sparse Comparison Data: Provably Fast Convergence for Nonconvex Matrix Factorization},
  author={Sankagiri, Suryanarayana and Etesami, Jalal and Grossglauser, Matthias},
  journal={arXiv preprint arXiv:2502.20033},
  year={2025}
}

@article{ouyang2025burer,
  title={{Burer--Monteiro} Factorizability of Nuclear Norm Regularized Optimization},
  author={Ouyang, Wenqing and Pong, Ting Kei and Yue, Man-Chung},
  journal={arXiv preprint arXiv:2505.00349},
  year={2025},
  note={Version 1},
  url={https://arxiv.org/abs/2505.00349v1}
}

@inproceedings{press2017using,
  title={Using the Output Embedding to Improve Language Models},
  author={Press, Ofir and Wolf, Lior},
  booktitle={Proceedings of the 15th Conference of the European Chapter of
    the Association for Computational Linguistics: Volume 2, Short Papers},
  pages={157--163},
  publisher={Association for Computational Linguistics},
  year={2017},
  url={https://aclanthology.org/E17-2025/}
}

@inproceedings{makkuva2024attention,
  title={Local to Global: Learning Dynamics and Effect of Initialization for Transformers},
  author={Makkuva, Ashok Vardhan and Bondaschi, Marco and Ekbote, Chanakya
    and Girish, Adway and Nagle, Alliot and Kim, Hyeji and Gastpar, Michael},
  booktitle={Advances in Neural Information Processing Systems},
  volume={37},
  year={2024},
  url={https://arxiv.org/abs/2406.03072}
}

@inproceedings{huang2025regular,
  title={How Transformers Learn Regular Language Recognition:
    A Theoretical Study on Training Dynamics and Implicit Bias},
  author={Huang, Ruiquan and Liang, Yingbin and Yang, Jing},
  booktitle={Proceedings of the 42nd International Conference on Machine Learning},
  volume={267},
  pages={25516--25549},
  year={2025},
  url={https://proceedings.mlr.press/v267/huang25s.html}
}

@article{jin2019nonconvex,
  title={On Nonconvex Optimization for Machine Learning: Gradients,
         Stochasticity, and Saddle Points},
  author={Jin, Chi and Netrapalli, Praneeth and Ge, Rong and Kakade, Sham M.
          and Jordan, Michael I.},
  journal={arXiv preprint arXiv:1902.04811},
  year={2019},
  note={arXiv version 2},
  url={https://arxiv.org/abs/1902.04811v2}
}

@inproceedings{ge2015escaping,
  title={Escaping From Saddle Points---Online Stochastic Gradient for Tensor
         Decomposition},
  author={Ge, Rong and Huang, Furong and Jin, Chi and Yuan, Yang},
  booktitle={Proceedings of the 28th Conference on Learning Theory},
  volume={40},
  pages={797--842},
  year={2015},
  url={https://proceedings.mlr.press/v40/Ge15.html}
}

@inproceedings{laurent2018deep,
  title={Deep Linear Networks with Arbitrary Loss: All Local Minima Are Global},
  author={Laurent, Thomas and von Brecht, James},
  booktitle={Proceedings of the 35th International Conference on Machine Learning},
  volume={80},
  pages={2902--2907},
  year={2018},
  url={https://proceedings.mlr.press/v80/laurent18a.html}
}

@inproceedings{ge2017nospurious,
  title={No Spurious Local Minima in Nonconvex Low Rank Problems: A Unified
         Geometric Analysis},
  author={Ge, Rong and Jin, Chi and Zheng, Yi},
  booktitle={Proceedings of the 34th International Conference on Machine Learning},
  volume={70},
  pages={1233--1242},
  year={2017},
  url={https://proceedings.mlr.press/v70/ge17a.html}
}

@article{li2017factorednuclear,
  title={Geometry of Factored Nuclear Norm Regularization},
  author={Li, Qiuwei and Zhu, Zhihui and Tang, Gongguo},
  journal={arXiv preprint arXiv:1704.01265},
  year={2017},
  url={https://arxiv.org/abs/1704.01265}
}

@inproceedings{zhu2021neuralcollapse,
  title={A Geometric Analysis of Neural Collapse with Unconstrained Features},
  author={Zhu, Zhihui and Ding, Tianyu and Zhou, Jinxin and Li, Xiao and
          You, Chong and Sulam, Jeremias and Qu, Qing},
  booktitle={Advances in Neural Information Processing Systems},
  volume={34},
  year={2021},
  url={https://proceedings.neurips.cc/paper_files/paper/2021/hash/f92586a25bb3145facd64ab20fd554ff-Abstract.html}
}

@inproceedings{ye2021asymmetric,
  title={Global Convergence of Gradient Descent for Asymmetric Low-Rank Matrix
         Factorization},
  author={Ye, Tian and Du, Simon S.},
  booktitle={Advances in Neural Information Processing Systems},
  volume={34},
  pages={1429--1439},
  year={2021}
}

@inproceedings{li2024mechanics,
  title={Mechanics of Next Token Prediction with Self-Attention},
  author={Li, Yingcong and Huang, Yixiao and Ildiz, Muhammed E. and
          Singh Rawat, Ankit and Oymak, Samet},
  booktitle={Proceedings of the 27th International Conference on Artificial
             Intelligence and Statistics},
  volume={238},
  pages={685--693},
  year={2024},
  url={https://proceedings.mlr.press/v238/li24f.html}
}

@inproceedings{thrampoulidis2024implicit,
  title={Implicit Optimization Bias of Next-Token Prediction in Linear Models},
  author={Thrampoulidis, Christos},
  booktitle={Advances in Neural Information Processing Systems},
  year={2024},
  url={https://arxiv.org/abs/2402.18551}
}

@inproceedings{varre2025ngrams,
  title={Learning In-context $n$-grams with Transformers: Sub-$n$-grams Are
         Near-Stationary Points},
  author={Varre, Aditya and Y{\"u}ce, Gizem and Flammarion, Nicolas},
  booktitle={Proceedings of the 42nd International Conference on Machine Learning},
  volume={267},
  pages={60924--60963},
  year={2025},
  url={https://proceedings.mlr.press/v267/varre25a.html}
}

@inproceedings{yang2024cooccurrence,
  title={Training Dynamics of Transformers to Recognize Word Co-occurrence via
         Gradient Flow Analysis},
  author={Yang, Hongru and Kailkhura, Bhavya and Wang, Zhangyang and Liang, Yingbin},
  booktitle={Advances in Neural Information Processing Systems},
  volume={37},
  year={2024}
}

@INPROCEEDINGS{guo2008exponential,
  author={Guo, Yuhong and Schuurmans, Dale},
  booktitle={2008 46th Annual Allerton Conference on Communication, Control, and Computing}, 
  title={Efficient global optimization for exponential family PCA and low-rank matrix factorization}, 
  year={2008},
  volume={},
  number={},
  pages={1100--1107},
  doi={10.1109/ALLERTON.2008.4797683}}

@inproceedings{yang2018softmax,
  title={Breaking the Softmax Bottleneck: A High-Rank {RNN} Language Model},
  author={Yang, Zhilin and Dai, Zihang and Salakhutdinov, Ruslan and
          Cohen, William W.},
  booktitle={International Conference on Learning Representations},
  year={2018},
  url={https://openreview.net/forum?id=HkwZSG-CZ}
}

@inproceedings{ganea2019softmax,
  title={Breaking the Softmax Bottleneck via Learnable Monotonic Pointwise
         Non-linearities},
  author={Ganea, Octavian and Gelly, Sylvain and B{\'e}cigneul, Gary and
          Severyn, Aliaksei},
  booktitle={Proceedings of the 36th International Conference on Machine
             Learning},
  volume={97},
  pages={2073--2082},
  year={2019},
  url={https://proceedings.mlr.press/v97/ganea19a.html}
}

@inproceedings{huang2024nonasymptotic,
  title={Non-asymptotic Convergence of Training Transformers for Next-token
         Prediction},
  author={Huang, Ruiquan and Liang, Yingbin and Yang, Jing},
  booktitle={Advances in Neural Information Processing Systems},
  volume={37},
  pages={80634--80673},
  year={2024},
  url={https://arxiv.org/abs/2409.17335}
}

@misc{inan2017tyingwordvectorsword,
      title={Tying Word Vectors and Word Classifiers: A Loss Framework for Language Modeling}, 
      author={Hakan Inan and Khashayar Khosravi and Richard Socher},
      year={2017},
      eprint={1611.01462},
      archivePrefix={arXiv},
      primaryClass={cs.LG},
      url={https://arxiv.org/abs/1611.01462}, 
}

@inproceedings{
zhao2024implicitgeometry,
title={Implicit Geometry of Next-token Prediction: From Language Sparsity Patterns to Model Representations},
author={Yize Zhao and Tina Behnia and Vala Vakilian and Christos Thrampoulidis},
booktitle={First Conference on Language Modeling},
year={2024},
url={https://openreview.net/forum?id=qyilOnIRHI}
}

@Inbook{John2014,
author="John, Fritz",
title="Extremum Problems with Inequalities as Subsidiary Conditions",
bookTitle="Traces and Emergence of Nonlinear Programming",
year="2014",
publisher="Springer Basel",
address="Basel",
pages="197--215",
isbn="978-3-0348-0439-4",
doi="10.1007/978-3-0348-0439-4_9",
url="https://doi.org/10.1007/978-3-0348-0439-4_9"
}

@InProceedings{liu2021noisyfactorization,
  title = 	 { Noisy Gradient Descent Converges to Flat Minima for Nonconvex Matrix Factorization },
  author =       {Liu, Tianyi and Li, Yan and Wei, Song and Zhou, Enlu and Zhao, Tuo},
  booktitle = 	 {Proceedings of The 24th International Conference on Artificial Intelligence and Statistics},
  pages = 	 {1891--1899},
  year = 	 {2021},
  editor = 	 {Banerjee, Arindam and Fukumizu, Kenji},
  volume = 	 {130},
  series = 	 {Proceedings of Machine Learning Research},
  month = 	 {13--15 Apr},
  publisher =    {PMLR},
  url = 	 {https://proceedings.mlr.press/v130/liu21e.html}
}

@inproceedings{bengio2003neural,
 author = {Bengio, Yoshua and Ducharme, R\'{e}jean and Vincent, Pascal},
 booktitle = {Advances in Neural Information Processing Systems},
 editor = {T. Leen and T. Dietterich and V. Tresp},
 pages = {},
 publisher = {MIT Press},
 title = {A Neural Probabilistic Language Model},
 url = {https://proceedings.neurips.cc/paper_files/paper/2000/file/728f206c2a01bf572b5940d7d9a8fa4c-Paper.pdf},
 volume = {13},
 year = {2000}
}
\bibliographystyle{iclr2027_conference}

\appendix
\section*{Proof organization}

The appendices are organized as follows.
Appendix~\ref{sec:model} restates the models and the training algorithm,
and introduces the additional notation used in the proofs.
Appendix~\ref{sec:positive} proves \cref{thm:iclr-untied} for the untied
model. Appendix~\ref{app:tied-proofs} proves
\cref{thm:iclr-tied-training} for the tied model, and
Appendix~\ref{app:transfer-proof} proves \cref{thm:iclr-transfer} on the
transfer to unseen states. Appendix~\ref{app:auxiliary} collects the
auxiliary results that are shared by these proofs. Finally,
Appendices~\ref{app:experiments} and~\ref{app:scaling-sgd} provide the
details of the experiments. Each proof follows the three steps of the
corresponding proof overview in the main text, which we summarize below.

\paragraph{Untied training.}
The proof of \cref{thm:iclr-untied} follows the three steps in
\cref{sec:untied-overview}. In Step 1, we show that the Gaussian
initialization lies in a ball of squared radius $O(V)$, and that the
objective outside this ball is larger than its initial value by a
constant. In Step 2, we show that every sufficiently accurate
approximate SOSP in this ball has gain above $7/8$. The key fact is that
the Hessian of the cross-entropy loss of each row with respect to its
logits has trace at most one. In Step 3, we specify the step size, the
perturbation scale, and the number of iterations, and prove by induction
over $d$ blocks of iterations that the trajectory stays in the ball and
that the gain is above $7/8$ at every iterate in the final window. After
the proof, we compare our analysis with a generic bound for factorized
optimization.

\paragraph{Tied training.}
The proof of \cref{thm:iclr-tied-training} has the same three steps. The
main difference is in Step 2, where the comparison with the ground-truth
Gram matrix has an error term proportional to $\|X\|_\fro^2/d^2$, and we
use the weight decay to absorb this term.

\paragraph{Conditional transfer.}
The proof of \cref{thm:iclr-transfer} follows the three steps in
\cref{sec:transfer-overview}. In Step 1, we compare the objective at
every iterate in the final window with the objective at the ground
truth. In Step 2, we use the coverage condition to turn this comparison
into bounds on the training loss and on the trace of the learned Gram
matrix. In Step 3, we use these two bounds and the positive
semidefiniteness of the Gram matrix to control the entire Gram matrix,
and thus the loss on the unseen states. The bound on the learned
embeddings is only used in Step 3, which is deterministic.

\section{Setup for the Proofs}
\label{sec:model}

In this section, we restate the models and the training algorithm in
\cref{sec:iclr-setup}, and introduce the additional notation that is
used in the proofs.

\paragraph{Notation.}
All logarithms are natural, and the softmax function is applied to a
matrix row-wise. For probability vectors $p,q\in\R^V$ with positive
entries and a logit vector $z\in\R^V$, we define
\[
 \softmax(z)_j=\frac{e^{z_j}}{\sum_{k=1}^V e^{z_k}},\qquad
 \KL(q\|p)=\sum_{j=1}^Vq_j\log\frac{q_j}{p_j},\qquad
 \CE(q,z)=\log\sum_{j=1}^V e^{z_j}-q^\top z.
\]
Recall that $\mathcal{L}$ is the row-centering operator. In particular,
the centered logits of the ground truth are
\begin{equation*}
 \mathcal{L}(\log Q)=(\log Q)\left(I_V-\frac1V\1\1^\top\right),
 \qquad \mathcal{L}(\log Q)\1=0.
\end{equation*}
We refer to the squared Frobenius norm of the parameters, e.g.,
$\|\Theta\|_\fro^2$, simply as their squared norm, which is also called
the energy in the main text.

\paragraph{Models.}
The untied model has parameters
\begin{equation*}
 \Theta=(E,U),\qquad E,U\in\R^{V\times d},\qquad
 P^{E,U}=\softmax(EU^\top),\qquad d<V.
\end{equation*}
Its centered logits are $\mathcal{L}(EU^\top)$, which have the same
row-wise softmax as $EU^\top$. The tied model has parameters
$\Theta=X\in\R^{V\times d}$, with logits $XX^\top$, centered logits
$\mathcal{L}(XX^\top)$, and $P^X=\softmax(XX^\top)$.
Thus, the number of parameters is $D_\Theta=2Vd$ for the untied model
and $D_\Theta=Vd$ for the tied model. In both models, every conditional
distribution is over all $V$ states. The untied model is trained on all
current states, while the tied model may be trained only on a subset
$\mathcal S\subset[V]$ of current states, where $N=|\mathcal S|$ and
$\mathcal U=[V]\setminus\mathcal S$.

\paragraph{Risk and gain.}
For any nonempty $\mathcal A\subseteq[V]$, the risk, the Bayes risk,
and their difference are given by
\begin{align*}
 \cR_{\mathcal A}(\Theta)
 &=-\frac1{|\mathcal A|}\sum_{i\in\mathcal A}\sum_{j=1}^V
 Q_{ij}\log P^\Theta_{ij},
 &\cH_{\mathcal A}
 &=-\frac1{|\mathcal A|}\sum_{i\in\mathcal A}\sum_{j=1}^V
 Q_{ij}\log Q_{ij},\\
 \cR_{\mathcal A}(\Theta)-\cH_{\mathcal A}
 &=\frac1{|\mathcal A|}\sum_{i\in\mathcal A}
 \KL(q_i\|P^\Theta_{i,:})\ge0.
\end{align*}
We refer to $\cR_{\mathcal A}(\Theta)-\cH_{\mathcal A}$ as the excess
loss, and to $\log V-\cH_{\mathcal A}$ as the information gap.
Whenever $\log V-\cH_{\mathcal A}>0$, the relative prediction gain is
\[
 g_{\mathcal A}(\Theta)
 =\frac{\log V-\cR_{\mathcal A}(\Theta)}{\log V-\cH_{\mathcal A}}
 =1-\frac{\cR_{\mathcal A}(\Theta)-\cH_{\mathcal A}}
 {\log V-\cH_{\mathcal A}}.
\]
By definition, the uniform predictor has gain zero and the ground-truth
predictor has gain one. Note that the gain only depends on the
prediction loss, and does not include the weight decay penalty.

\paragraph{Ground-truth classes.}
We restate the classes of ground-truth transition tables in the main
text.

\begin{definition}[Base class]
\label{def:teacher-class}
Fix a positive integer $r$ and constants $B,\Delta_0>0$ independent
of $V$. The class $\mathcal Q_{r,B,\Delta_0}$ consists of transition
tables with positive entries and row sums one such that
\[
 \rank \mathcal{L}(\log Q)\le r,\qquad \|\mathcal{L}(\log Q)\|_\infty\le B,\qquad
 \log V-\cH_{[V]}\ge\Delta_0,
\]
where $\|\cdot\|_\infty$ denotes the largest absolute entry of a matrix.
\end{definition}
When $r\le d$, every ground truth in $\mathcal Q_{r,B,\Delta_0}$ can be
represented by the untied model. The bound on the centered logits
implies that every transition probability is within constant multiples
of $1/V$, and the condition on the information gap excludes the case
where the uniform predictor is asymptotically optimal.

\begin{definition}[Tied-compatible class]
\label{def:appendix-tied-class}
For a fixed constant $B_Y\ge1$, the tied-compatible subclass
$\mathcal Q^{\rm tied}_{r,B,\Delta_0}$ consists of tables
$Q\in\mathcal Q_{r,B,\Delta_0}$ admitting a representative
\[
 Q=\softmax(YY^\top),\qquad
 Y\in\R^{V\times r_Y},\quad r_Y\le r,\quad
 \max_i\|y_i\|_2^2\le B_Y,
\]
where $y_i$ denotes the $i$-th row of $Y$ as a column vector.
\end{definition}

\begin{definition}[Centered coverage for transfer]
\label{def:appendix-transfer-coverage}
A representative $Y$ in \cref{def:appendix-tied-class} satisfies
centered coverage on $\mathcal S$ if its columns are linearly
independent and, for a fixed $\kappa>0$,
\[
 Y^\top\1=0,\qquad
 \bar y_{\mathcal S}=\frac1N\sum_{i\in\mathcal S}y_i,\qquad
 \frac1N\sum_{i\in\mathcal S}
 (y_i-\bar y_{\mathcal S})(y_i-\bar y_{\mathcal S})^\top
 \succeq\kappa^2 I.
\]
\end{definition}
The coverage condition is only used in the proof of
\cref{thm:iclr-transfer}, and is not needed for
\cref{thm:iclr-tied-training}.

\paragraph{Training algorithm.}
All entries of the initial parameters $\Theta_0$ are independent
$\mathcal N(0,1/d)$ random variables. The regularized objective is
\begin{equation*}
 \cF_{\lambda,\mathcal A}(\Theta)
 =\cR_{\mathcal A}(\Theta)+\frac\lambda2\|\Theta\|_\fro^2,
 \qquad \|(E,U)\|_\fro^2=\|E\|_\fro^2+\|U\|_\fro^2.
\end{equation*}
For the untied model trained on all states, we write
$\cF_\lambda=\cF_{\lambda,[V]}$, and denote its global minimum by
$\cF_\lambda^*=\min_{E,U}\cF_\lambda(E,U)$. All parameters are trained by perturbed gradient
descent:
\begin{equation}
 \Theta_{t+1}=\Theta_t-h\{\nabla\cF_{\lambda,\mathcal A}(\Theta_t)+\xi_t\},
 \qquad
 \xi_t\stackrel{\rm iid}{\sim}
 \mathcal N\!\left(0,\frac{\sigma^2}{D_\Theta}I_{D_\Theta}\right).
 \label{eq:appendix-noisy-gd}
\end{equation}
The perturbations $\{\xi_t\}$ are independent of the initialization. The
weight decay $\lambda$, the step size $h$, the perturbation scale
$\sigma$, and the number of iterations $T$ are deterministic and chosen
before training, and $h$ and $\sigma$ are fixed throughout training.
In the proof of \cref{thm:iclr-untied}, we use
$\lambda=\Delta_0/(8BV\sqrt r)$ and $\mathcal A=[V]$.
In the proofs of \cref{thm:iclr-tied-training,thm:iclr-transfer}, the
weight decay is specified by each theorem, and the quantity $\lambda V$
always refers to the weight decay of the theorem being proved.

\paragraph{Two facts on the untied objective.}
We end this section with two facts on the untied objective
$\cF_\lambda$. They are not used in the proofs of the main theorems.
Instead, they relate the factorized objective to a nuclear-norm
regularized problem over the logit matrix, which is the formulation
used in the comparison with existing results in
Appendix~\ref{app:comparison}. The first fact is that the mean of the
rows of the output embedding matrix $U$ does not affect the
predictions, and is decoupled from the rest of the parameters.

\begin{lemma}[Decoupling of the output mean]
\label{lem:output-mean-decoupling}
For $U\in\R^{V\times d}$,
\begin{equation}
 \begin{aligned}
 U&=\left(I_V-\frac{\1\1^\top}{V}\right)U
       +\frac{\1\1^\top U}{V},\\
 \|U\|_\fro^2
 &=\left\|\left(I_V-\frac{\1\1^\top}{V}\right)U\right\|_\fro^2
       +\frac{\|\1^\top U\|_2^2}{V}.
 \end{aligned}
 \label{eq:output-mean-orthogonal-split}
\end{equation}
Consequently,
\[
 \cF_{\lambda,[V]}(E,U)
 =\cF_{\lambda,[V]}\!\left(E,
       \left(I_V-\frac{\1\1^\top}{V}\right)U\right)
       +\frac{\lambda}{2V}\|\1^\top U\|_2^2.
\]
The predictions and the dynamics of the centered factors are
independent of $\1^\top U$. Under \cref{eq:appendix-noisy-gd},
\begin{equation}
 \1^\top U_{t+1}
 =(1-h\lambda)\1^\top U_t-h\1^\top\xi_t^U,
 \label{eq:output-mean-noisy-update}
\end{equation}
where $\xi_t^U$ is the component of $\xi_t$ that corresponds to $U$.
Moreover, at every critical point with $\lambda>0$, we have
$\1^\top U=0$.
\end{lemma}

\begin{proof}
The two components of $U$ in \cref{eq:output-mean-orthogonal-split} are
orthogonal in the Frobenius inner product, which gives the second
identity in \cref{eq:output-mean-orthogonal-split}. Moreover, replacing
$U$ by $(I_V-\1\1^\top/V)U$ subtracts the same constant from all entries
in each row of $EU^\top$, which does not change the row-wise softmax.
Combining these two facts gives the decomposition of the objective.
Next, since each row of $P^{E,U}-Q$ sums to zero, the gradient of the
prediction loss with respect to $U$ satisfies
$\1^\top\nabla_U\cR_{[V]}(\Theta)=0$. Thus, multiplying the update of
$U$ in \cref{eq:appendix-noisy-gd} by $\1^\top$ from the left gives
\cref{eq:output-mean-noisy-update}. The remaining statements follow
from the decomposition of the objective and
\cref{eq:output-mean-noisy-update}.
\end{proof}

The second fact is the following reformulation of the global minimum.

\begin{remark}[Bayes optimum and regularized optimum]
\label{def:three-optima}
The Bayes risk is $\cH_{[V]}$, and a predictor is Bayes-optimal if and
only if $P^{E,U}=Q$, or equivalently, $\cR_{[V]}(\Theta)=\cH_{[V]}$ and
$g_{[V]}(\Theta)=1$. Thus, if $d\ge\rank \mathcal{L}(\log Q)$, the
minimum of the risk over the untied model without weight decay is
$\cH_{[V]}$. On the other hand, the global minimum of the regularized
objective satisfies
\begin{equation*}
 \begin{aligned}
 \cF_\lambda^*
 &=\min_{E,U\in\R^{V\times d}}\cF_\lambda(E,U)\\
 &=\min_{\substack{M\in\R^{V\times V},\ M\1=0\\\rank M\le d}}
 \left\{\frac1V\sum_{i=1}^V\CE(q_i,M_{i,:})+\lambda\|M\|_*\right\},
 \end{aligned}
\end{equation*}
where $\|M\|_*$ is the nuclear norm of $M$. The second equality follows
from \cref{lem:output-mean-decoupling} and the identity
\[
 \min_{EU^\top=M}\frac{\|E\|_\fro^2+\|U\|_\fro^2}{2}
 =\|M\|_*.
\]
When $\log V-\cH_{[V]}>0$ and $\lambda>0$, every global minimizer of
$\cF_\lambda$ has positive excess loss and gain strictly smaller than
one, as shown in \cref{rem:which-global-optimum}.
\end{remark}

\section{Proof of Theorem~\ref{thm:iclr-untied}}
\label{sec:positive}

In this section, we present the full proof of \cref{thm:iclr-untied}.
The proof consists of three main steps.

\textbf{Step 1 (Appendix~\ref{sec:untied-step1}).}
We show that, with high probability, the Gaussian initialization has
squared norm at most $3V$ and bounded loss
(\cref{lem:untied-gaussian-initialization}). Due to the weight decay,
the objective outside a ball of squared radius $O(V)$ is then larger
than its initial value by a constant. This allows us to replace the
objective outside the ball by a globally smooth extension
(\cref{lem:localization}), to which the results on perturbed gradient
descent apply.

\textbf{Step 2 (Appendix~\ref{sec:untied-step2}).}
We show that every sufficiently accurate approximate SOSP in the ball
has objective below $\cH_{[V]}+\Delta_0/8$, which implies gain above
$7/8$. To this end, we construct a comparator whose objective is below
this level by a fixed margin (\cref{lem:shrunken-comparator}), and a
direction that represents the comparator and has small softmax
curvature (\cref{lem:fisher-packing}). Then, we show that any point in
the ball whose objective is not below this level has either a large
gradient or a direction of negative curvature
(\cref{lem:highgain-collar,cor:highgain-sosp}).

\textbf{Step 3 (Appendix~\ref{sec:untied-step3}).}
We partition the iterations into $d$ blocks. By the results of
\citet{jin2019nonconvex} (\cref{thm:jin-visit,lem:jin-descent}), with
high probability, each block contains an approximate SOSP, and the
objective cannot increase much within a block
(\cref{thm:tied-fixed-noise-consequence}). We then prove by induction
over the blocks that the trajectory stays in the ball and that the gain
is above $7/8$ at every iterate after the first block.

\paragraph{Notation and parameters.}
Throughout this section, $\Theta=(E,U)$,
$\cF_\lambda=\cF_{\lambda,[V]}$, and $\cR=\cR_{[V]}$. All $2Vd$ entries
of $\Theta$ are updated by \cref{eq:appendix-noisy-gd}. The weight decay
is
\begin{equation*}
 \lambda=\frac{\Delta_0}{8BV\sqrt r},
\end{equation*}
and we define $t_\star=1-\Delta_0/(8B^2)$, which is the factor by which
the ground-truth logits are shrunk in the comparator of Step 2.
Unless otherwise stated, the constants hidden in $O(\cdot)$ do not
depend on $V$, and their dependence on the fixed parameters
$(B,r,\Delta_0)$ is made explicit through the named constants below.

\subsection{Step 1: Localize the Trajectory}
\label{sec:untied-step1}

We first bound the energy and loss at initialization. This fixes a
deterministic objective budget before the perturbations are drawn.
\begin{lemma}[Ordinary-Gaussian initialization]
\label{lem:untied-gaussian-initialization}
Under the initialization in \cref{sec:model}, there is
a universal $c_0>0$ and a universal $C_{\rm init}\ge1$ such that, whenever
$d\ge C_{\rm init}\log(eV)$,
\begin{equation}
 \Pr\left\{\|E_0\|_{\mathrm F}^2 + \|U_0\|_{\mathrm F}^2\le 3V,\ \|E_0U_0^\top\|_\infty\le1 \right\}
 \ge1-e^{-c_0d}.
 \label{eq:untied-initialization-event}
\end{equation}
On this event,
\begin{equation*}
 \cF_\lambda(\Theta_0)-\cH_{[V]}\le 2B+2+\frac{3\Delta_0}{8B\sqrt r},
 \qquad \cF_\lambda(\Theta_0)-\cF_\lambda^*\le 2B+2+\frac{3\Delta_0}{8B\sqrt r}.
\end{equation*}
\end{lemma}

\begin{proof}
All entries of $E_0, U_0$ are independent $\mathcal N(0,1/d)$.  Hence
$d \left(\|E_0\|_{\mathrm F}^2 + \|U_0\|_{\mathrm F}^2\right)$ is chi-square with $2Vd$ degrees of freedom.  A
standard Chernoff bound gives
$\Pr\{\|E_0\|_{\mathrm F}^2 + \|U_0\|_{\mathrm F}^2>3V\}\le e^{-cdV}$ for a universal constant $c>0$. Applying the same bound to each
row of $E_0$ and taking a union bound yields
\begin{equation*}
 \Pr\left\{\max_i\|E_{0,i,:}\|_2^2>2\right\}
 \le Ve^{-cd}.
\end{equation*}

Conditional on $E_0$, each entry $(E_0U_0^\top)_{ij}$ is Gaussian
with mean zero and variance $\|E_{0,i,:}\|_2^2/d$. Thus, whenever
$\max_i\|E_{0,i,:}\|_2^2\le2$, each variance is at most $2/d$.
Gaussian tail bounds and a union bound over all $V^2$ entries give
\[
 \Pr\left\{
 \|E_0U_0^\top\|_\infty>1,\ 
 \max_i\|E_{0,i,:}\|_2^2\le2
 \right\}
 \le 2V^2e^{-d/4}.
\]
Combining these estimates, the failure probability in
\cref{eq:untied-initialization-event} is at most
\[
 e^{-cVd}+Ve^{-cd}+2V^2e^{-d/4}
 \le e^{-c_0d},
\]
provided $d\ge C_{\rm init}\log(eV)$ for a sufficiently large
universal $C_{\rm init}$ and a sufficiently small universal $c_0>0$.

On the event in \cref{eq:untied-initialization-event}, all initial
logits lie in $[-1,1]$, so each row's cross-entropy loss is at most
$\log V+2$. Hence $\cR(\Theta_0)\le\log V+2$.
Using $\lambda = \frac{\Delta_0}{8BV\sqrt r}$ and
$\cH_{[V]}\ge\log V-B$ from
\cref{eq:teacher-information-quadratic}, we obtain
\[
 \begin{aligned}
 \cF_\lambda(\Theta_0)-\cH_{[V]}
 &=
 \cR(\Theta_0)-\cH_{[V]}
 +\frac{\lambda}{2}
 \left(\|E_0\|_{\mathrm F}^2+\|U_0\|_{\mathrm F}^2\right)\\
 &\le B+2+\frac{\Delta_0}{16BV\sqrt r} \times 3V
 \le 2B+2+\frac{3\Delta_0}{8B\sqrt r}.
 \end{aligned}
\]
Finally, $\cR(\Theta_0)\ge\cH_{[V]}$ and the regularization term is
nonnegative, so $\cF_\lambda^*\ge\cH_{[V]}$. Therefore,
\[
 \cF_\lambda(\Theta_0)-\cF_\lambda^*
 \le \cF_\lambda(\Theta_0)-\cH_{[V]}
 \le 2B+2+\frac{3\Delta_0}{8B\sqrt r}.
\]
\end{proof}

On the initialization event in \cref{lem:untied-gaussian-initialization},
any $\Theta=(E,U)$ satisfying
\[
 \|E\|_\fro^2+\|U\|_\fro^2
 \ge\left(6+\frac{32B\sqrt r(B+3)}{\Delta_0}\right)V
\]
has regularized loss at least four above its initial value. Indeed,
$\cR_{[V]}(\Theta)\ge\cH_{[V]}$ and
$\lambda=\Delta_0/(8BV\sqrt r)$ give
\begin{equation}
 \begin{aligned}
 \cF_\lambda(\Theta)
 &\ge\cH_{[V]}+\frac\lambda2
       \bigl(\|E\|_\fro^2+\|U\|_\fro^2\bigr)\\
 &\ge\cH_{[V]}+2B+6+\frac{3\Delta_0}{8B\sqrt r}
 \ge\cF_\lambda(\Theta_0)+4.
 \end{aligned}
 \label{eq:highgain-wall-separation}
\end{equation}
Since $\lambda>0$, the objective is coercive and attains its global
minimum. Every global minimizer lies strictly inside this ball:
its objective is at most $\cF_\lambda(\Theta_0)$, whereas
\cref{eq:highgain-wall-separation} holds on and outside the boundary.
The initialization also lies inside because its squared norm is at most
$3V$. Thus \cref{lem:localization} gives a smooth extension that agrees
with $\cF_\lambda$ inside the ball and preserves the exterior loss
barrier. Step 3 will show that the noisy trajectory stays inside.

The extension has global gradient- and Hessian-Lipschitz bounds of
orders $O(1)$ and $O(\sqrt V)$, respectively, with constants depending
only on $(B,r,\Delta_0)$. To verify these orders, the first three
derivatives of population cross-entropy with respect to logits have
norms $O(V^{-1/2})$, $O(V^{-1})$, and $O(V^{-1})$.
On the factor ball above, the first two derivatives of
$\Theta\mapsto EU^\top$ have norms $O(\sqrt V)$ and $O(1)$.
A radial clipping map with a fixed relative buffer has first, second,
and third derivative norms $O(1)$, $O(V^{-1/2})$, and $O(V^{-1})$.
The chain rule gives the stated global smoothness bounds.

\subsection{Step 2: Convert Approximate Stationarity into Gain}
\label{sec:untied-step2}

In this step, we show that every sufficiently accurate approximate SOSP
in the ball $\|\Theta\|_\fro^2\le C_EV$ satisfies
$\cF_\lambda(\Theta)<\cH_{[V]}+\Delta_0/8$, where $C_E$ is a constant
that will be specified in Step 3. Since
$\cR_{[V]}(\Theta)\le\cF_\lambda(\Theta)$ and
$\log V-\cH_{[V]}\ge\Delta_0$, any such point has gain above $7/8$.

The proof is based on the Hessian of the objective. Recall from
\cref{lem:factor-calculus} in Appendix~\ref{app:factor-derivatives} that,
for any direction $(\dot E,\dot U)$ with the same dimensions as $(E,U)$,
\begin{equation*}
 \begin{aligned}
 \nabla^2\cF_\lambda(E,U)[(\dot E,\dot U),(\dot E,\dot U)]
 &=\underbrace{\frac1V\sum_{i=1}^V\operatorname{Var}_{j\sim P^{E,U}_{i,:}}
 \bigl[(\dot E U^\top+E\dot U^\top)_{ij}\bigr]}_{\text{softmax curvature term}}\\
 &\quad+\underbrace{\frac2V\langle P^{E,U}-Q,\dot E\dot U^\top\rangle}_{\text{bilinear term}}
 +\underbrace{\lambda(\|\dot E\|_\fro^2+\|\dot U\|_\fro^2)}_{\text{weight decay term}}.
 \end{aligned}
\end{equation*}
The softmax curvature term is nonnegative, while the bilinear term can
be negative. Our goal is to find a direction along which the sum of the
three terms is negative, whenever the objective at $\Theta$ is not
sufficiently below $\cH_{[V]}+\Delta_0/8$ and the gradient at $\Theta$
is small. We proceed in three parts. First, we construct a comparator
$(E_\star,U_\star)$ whose objective is below $\cH_{[V]}+\Delta_0/8$ by a
fixed margin (\cref{lem:shrunken-comparator}). Second, we construct a
direction $(\dot E,\dot U)$ with
$\dot E\dot U^\top=E_\star U_\star^\top$ whose softmax curvature term is
small (\cref{lem:fisher-packing}). Third, we show that along this
direction, the bilinear term and the weight decay term are negative
enough to dominate the softmax curvature term
(\cref{lem:highgain-collar}).

\paragraph{Part 1: A comparator below the target level.}
We first construct the comparator. A natural candidate is a
factorization of the ground-truth logits $\mathcal{L}(\log Q)$, whose
prediction loss equals $\cH_{[V]}$. By \cref{lem:maxnorm-factor} in
Appendix~\ref{app:factor-tools}, such a factorization exists with
squared norm at most $2B\sqrt r\,V$, so that its weight decay penalty is
at most $\Delta_0/8$ under our choice of $\lambda$. However, this only
shows that its objective is at most $\cH_{[V]}+\Delta_0/8$, without any
margin. To obtain a margin, we shrink the logits by the factor
$t_\star=1-\Delta_0/(8B^2)<1$. The shrinkage reduces the penalty by an
amount that is linear in $1-t_\star$, while it increases the prediction
loss only by an amount that is quadratic in $1-t_\star$. The following
lemma makes this precise.
\begin{lemma}[Comparator with shrunken ground-truth logits]
\label{lem:shrunken-comparator}
For every ground truth in $\mathcal Q_{r,B,\Delta_0}$ with
$\rank \mathcal{L}(\log Q)\le d$,
\begin{equation}
 (\log V-\cH_{[V]})\le\min\left\{B,\frac{B^2}{2}\right\}.
 \label{eq:teacher-information-quadratic}
\end{equation}
In particular, $0<1-t_\star\le1/16$. Moreover, there exist factors
$E_\star,U_\star\in\R^{V\times d}$ satisfying
\[
 \begin{gathered}
 E_\star U_\star^\top=t_\star \mathcal{L}(\log Q),\qquad U_\star^\top\1=0,\\
 \max_i\|(E_\star)_{i,:}\|_2^2\le t_\star B\sqrt{r},\qquad
 \max_j\|(U_\star)_{j,:}\|_2^2\le t_\star B\sqrt{r},
 \end{gathered}
\]
where all but $\rank \mathcal{L}(\log Q)$ columns of $E_\star$ and
$U_\star$ are zero. These factors satisfy
\begin{equation}
 \cF_\lambda(E_\star,U_\star)
 =\cR_{[V]}(E_\star,U_\star)
   +\frac\lambda2(\|E_\star\|_\fro^2+\|U_\star\|_\fro^2)
 \le \cH_{[V]}+\frac{\Delta_0}{8}-\frac{\Delta_0^2}{128B^2}.
 \label{eq:shrunken-comparator}
\end{equation}
\end{lemma}

\begin{proof}
We first prove \cref{eq:teacher-information-quadratic}. For every row
$i$ and every real number $s$, the second derivative of the following
function of $s$ is a variance, which is at most $B^2$ because the
centered logits are bounded by $B$:
\[
 \frac{d^2}{ds^2}\log\left(\frac1V\sum_j e^{s\mathcal{L}(\log Q)_{ij}}\right)
 =\operatorname{Var}_{j\sim\operatorname{softmax}(s\mathcal{L}(\log Q)_{i,:})}
 (\mathcal{L}(\log Q)_{ij})\le B^2.
\]
Since $\mathcal{L}(\log Q)$ is row-centered, both this function and its
first derivative vanish at $s=0$. Thus, integrating by parts, the KL
divergence between $q_i$ and the uniform distribution satisfies
\[
 \operatorname{KL}(q_i\|\1/V)
 =\int_0^1s\frac{d^2}{ds^2}
       \log\left(\frac1V\sum_j e^{s\mathcal{L}(\log Q)_{ij}}\right)\,ds
 \le\frac{B^2}{2}.
\]
On the other hand, since $\mathcal{L}(\log Q)$ is row-centered, Jensen's
inequality implies that the logarithm in the following display is
nonnegative, and thus
\[
 \operatorname{KL}(q_i\|\1/V)
 =\sum_jq_{ij}\mathcal{L}(\log Q)_{ij}
    -\log\left(\frac1V\sum_j e^{\mathcal{L}(\log Q)_{ij}}\right)
 \le\max_j\mathcal{L}(\log Q)_{ij}\le B.
\]
Since $\log V-\cH_{[V]}=\frac1V\sum_i\KL(q_i\|\1/V)$, averaging the above
two bounds over $i$ proves \cref{eq:teacher-information-quadratic}. In
particular, $\Delta_0\le\log V-\cH_{[V]}\le B^2/2$ gives
$0<1-t_\star=\Delta_0/(8B^2)\le1/16$.

Next, we construct the comparator. We apply \cref{lem:maxnorm-factor}
to the matrix $t_\star\mathcal{L}(\log Q)$, which has rank at most $r$,
entries bounded by $t_\star B$, and zero row sums. This gives factors
with $\rank\mathcal{L}(\log Q)$ columns that satisfy the product
identity, the centering condition, and the bounds on the row norms in
the lemma. Appending zero columns gives
$E_\star,U_\star\in\R^{V\times d}$.

It remains to prove \cref{eq:shrunken-comparator}. For the prediction
loss, we view $\KL(q_i\|\softmax(t\mathcal{L}(\log Q)_{i,:}))$ as a
function of $t$. This function
and its first derivative vanish at $t=1$, and its second derivative is
the same variance as above, which is at most $B^2$. Thus, Taylor's
theorem gives
\[
 \operatorname{KL}\!\left(
 q_i\middle\|\operatorname{softmax}(t_\star \mathcal{L}(\log Q)_{i,:})
 \right)
 \le\frac{B^2}{2}(1-t_\star)^2.
\]
Consequently,
\begin{equation}
 \cR_{[V]}(E_\star,U_\star)-\cH_{[V]}
 \le\frac{B^2(1-t_\star)^2}{2}=\frac{\Delta_0^2}{128B^2}.
 \label{eq:shrunken-risk}
\end{equation}
For the weight decay penalty, the bounds on the row norms of
$E_\star$ and $U_\star$ imply that
$\frac\lambda2(\|E_\star\|_\fro^2+\|U_\star\|_\fro^2)$ is at most
\[
 \lambda t_\star B\sqrt r\,V=t_\star\frac{\Delta_0}{8}.
\]
Since $(1-t_\star)\Delta_0/8=\Delta_0^2/(64B^2)$, adding this penalty to
\cref{eq:shrunken-risk} proves \cref{eq:shrunken-comparator}.
\end{proof}

We remark that the margin $\frac{\Delta_0^2}{128B^2}$ in
\cref{eq:shrunken-comparator} is what allows the objective to increase
slightly between two visits to approximate SOSPs in Step 3.

\paragraph{Part 2: A direction with small softmax curvature.}
Next, we construct the direction $(\dot E,\dot U)$. Since the comparator
has at most $r$ nonzero columns, we can place each of its columns along
any one of $d$ orthonormal directions in the embedding space without
changing the product $E_\star U_\star^\top$. We refer to this
construction as packing. The key observation is that the total softmax
curvature over all embedding directions is limited. Specifically, the
Hessian of the cross-entropy loss of row $i$ with respect to its logits
is the covariance matrix
$\operatorname{diag}(P^{E,U}_{i,:})
 -(P^{E,U}_{i,:})^\top P^{E,U}_{i,:}$,
which is positive semidefinite and has trace at most one. Hence, the
sum of the traces over the $V$ rows is at most $V$, and most of the $d$
directions have softmax curvature of order $V/d$ relative to their
squared norm. The following lemma chooses $r$ such directions, along
which the squared norm of $\Theta$ is also small.

\begin{lemma}[Packing the comparator into directions with small softmax curvature]
\label{lem:fisher-packing}
Suppose $d\ge2r$ and $\|\Theta\|_\fro^2\le C_EV$.
Then, there exists a direction $(\dot E,\dot U)$ with
$\dot U^\top\1=0$ such that
\begin{align}
 \dot E\dot U^\top&=t_\star \mathcal{L}(\log Q),
 \label{eq:packing-product}\\
 \|\dot E\|_\fro^2+\|\dot U\|_\fro^2&\le2B\sqrt{r} V,
 \label{eq:packing-energy}\\
 \frac1V\sum_i
 \operatorname{Var}_{j\sim P^{E,U}_{i,:}}
 \bigl((\dot E U^\top+E\dot U^\top)_{ij}\bigr)
 &\le\frac{8B r^{3/2}C_E}{d}
       \left(\frac{4V}{d}+1\right).
 \label{eq:packing-fisher}
\end{align}
\end{lemma}

\begin{proof}
We first select the directions in the embedding space. Let
$w_1,\ldots,w_d$ be an orthonormal eigenbasis of $U^\top U$. Then, the
vectors $Uw_j/\|Uw_j\|_2$ with $Uw_j\ne0$ are orthonormal in $\R^V$.
For a vector $v\in\R^V$, we write
$\operatorname{Var}_{P^{E,U}_{i,:}}(v)$ for the variance of its
coordinate $v_j$ when $j$ is drawn from the distribution
$P^{E,U}_{i,:}$. We have
\begin{align*}
 \sum_{j=1}^d\bigl(\|Uw_j\|_2^2+\|Ew_j\|_2^2\bigr)
 &=\|U\|_\fro^2+\|E\|_\fro^2\le C_EV,\\
 \sum_{j:\,Uw_j\ne0}
 \frac{\sum_{i=1}^V\operatorname{Var}_{P^{E,U}_{i,:}}(Uw_j)}
      {\|Uw_j\|_2^2}
 &\le\sum_{i=1}^V\left(1-\sum_{\ell=1}^V(P^{E,U}_{i\ell})^2\right)
 \le V,
\end{align*}
where the first line follows from the orthonormality of $\{w_j\}$ and
the assumption $\|\Theta\|_\fro^2\le C_EV$, and the second line follows
from the fact that the sum of the quadratic forms of a covariance matrix
over orthonormal vectors is at most its trace. Therefore, at most $d/4$
indices $j$ violate the first of the following two bounds, and at most
$d/4$ indices violate the second one:
\[
 \|Uw_j\|_2^2+\|Ew_j\|_2^2\le\frac{4C_EV}{d},\qquad
 \sum_{i=1}^V\operatorname{Var}_{P^{E,U}_{i,:}}(Uw_j)
 \le\frac{4V}{d}\|Uw_j\|_2^2.
\]
Here, the second bound holds trivially when $Uw_j=0$. Thus, at least
$d/2\ge r$ indices satisfy both bounds.

Next, we construct the direction. Let $(E_\star,U_\star)$ be the
comparator in \cref{lem:shrunken-comparator}, whose nonzero columns are
the first $\rank \mathcal{L}(\log Q)$ columns. We choose distinct indices
$j_1,\ldots,j_{\rank \mathcal{L}(\log Q)}$ that satisfy both bounds
above, and signs $\varepsilon_k\in\{-1,1\}$ that will be specified
later. Define
\[
 \dot E=\sum_{k=1}^{\rank \mathcal{L}(\log Q)}\varepsilon_k(E_\star)_{:,k}w_{j_k}^\top,
 \qquad
 \dot U=\sum_{k=1}^{\rank \mathcal{L}(\log Q)}\varepsilon_k(U_\star)_{:,k}w_{j_k}^\top.
\]
Since $\{w_{j_k}\}$ are orthonormal and $\varepsilon_k^2=1$, for every
choice of the signs we have
$\dot E\dot U^\top=E_\star U_\star^\top=t_\star \mathcal{L}(\log Q)$ and
\begin{equation}
 \|\dot E\|_\fro^2+\|\dot U\|_\fro^2
 =\|E_\star\|_\fro^2+\|U_\star\|_\fro^2
 \le2t_\star B\sqrt{r} V.
 \label{eq:packing-comparator-energy}
\end{equation}
In addition, $U_\star^\top\1=0$ implies $\dot U^\top\1=0$. This proves
\cref{eq:packing-product,eq:packing-energy} and $\dot U^\top\1=0$.

It remains to choose the signs such that \cref{eq:packing-fisher} holds.
By the bounds on the row norms of the comparator in
\cref{lem:shrunken-comparator}, for every column $k$ and every row $i$,
\[
 (E_\star)_{ik}^2\le B\sqrt{r},\qquad
 \operatorname{Var}_{j\sim P^{E,U}_{i,:}}((U_\star)_{jk})
 \le\sum_j P^{E,U}_{ij}(U_\star)_{jk}^2\le B\sqrt{r}.
\]
By the definition of $(\dot E,\dot U)$, the $i$-th row of
$\dot EU^\top+E\dot U^\top$ is the sum over $k$ of
$\varepsilon_k\{(E_\star)_{ik}Uw_{j_k}+(Ew_{j_k})_i(U_\star)_{:,k}\}^\top$.
Suppose that the signs are independent and uniformly distributed on
$\{-1,1\}$. Then, in the expectation of the softmax curvature term over
the signs, the cross terms between different columns vanish, and the
expectation is the sum of the contributions of the individual columns.
Using $\operatorname{Var}(a+b)\le
2\operatorname{Var}(a)+2\operatorname{Var}(b)$, the contribution of
column $k$ is bounded as
\[
 \begin{aligned}
 &\frac1V\sum_{i=1}^V
 \operatorname{Var}_{P^{E,U}_{i,:}}\!\left(
 (E_\star)_{ik}Uw_{j_k}+(Ew_{j_k})_i(U_\star)_{:,k}\right)\\
 &\quad\le\frac{2B\sqrt r}{V}\left\{
 \sum_{i=1}^V\operatorname{Var}_{P^{E,U}_{i,:}}(Uw_{j_k})
 +\|Ew_{j_k}\|_2^2\right\}
 \le\frac{8B\sqrt r\,C_E}{d}\left(\frac{4V}{d}+1\right),
 \end{aligned}
\]
where the last inequality follows from the two bounds satisfied by the
index $j_k$. Since there are at most $r$ columns, the expectation of the
softmax curvature term is at most the right-hand side of
\cref{eq:packing-fisher}. Therefore, there exists a choice of the signs
for which \cref{eq:packing-fisher} holds.
\end{proof}

We note that \cref{lem:fisher-packing} only uses an upper bound on the
traces of the covariance matrices. Thus, it also holds when some rows
of $P^{E,U}$ are nearly deterministic and have very small curvature.

\paragraph{Part 3: Large gradient or negative curvature.}
We now evaluate the Hessian along the direction in
\cref{lem:fisher-packing}. By \cref{eq:packing-fisher}, the softmax
curvature term is small when $d\ge C\sqrt V$. For the bilinear term and
the weight decay term, we use the convexity of the cross-entropy loss
in the logits to relate them to the gap between the objective at
$\Theta$ and the objective at the comparator, which is positive when
the objective at $\Theta$ is large. The remaining error is the inner
product between the gradient and $\Theta$, which is small when the
gradient is small. This leads to the following lemma.
\begin{lemma}[Large gradient or negative curvature above the target level]
\label{lem:highgain-collar}
Fix a constant $C_E>0$. Suppose $d\ge2r$ and
\begin{equation}
 d^2\ge \frac{20480B^3r^{3/2}C_E}{\Delta_0^2}V.
 \label{eq:highgain-size-body}
\end{equation}
Then, every point $\Theta$ with $\|\Theta\|_\fro^2\le C_EV$ and
\begin{equation*}
 \cF_\lambda(\Theta)-\cH_{[V]}
 \ge\frac{\Delta_0}{8}-\frac{\Delta_0^2}{256B^2}
\end{equation*}
satisfies
\begin{equation}
 \|\nabla \cF_\lambda(\Theta)\|_\fro
 >\frac{\Delta_0^2}{512B^2\sqrt{C_EV}}
 \quad\hbox{or}\quad
 \lambda_{\min}(\nabla^2\cF_\lambda(\Theta))
 \le-\frac{\Delta_0^2}{512B^3\sqrt r\,V}.
 \label{eq:highgain-alternative}
\end{equation}
\end{lemma}

\begin{proof}
Suppose that the first inequality in \cref{eq:highgain-alternative} does
not hold, i.e.,
$\|\nabla\cF_\lambda(\Theta)\|_\fro\le\Delta_0^2/(512B^2\sqrt{C_EV})$.
We show that the second inequality must hold.

First, we bound the inner product between the gradient and $\Theta$.
By \cref{eq:factor-gradients}, we have
$\langle\nabla\cF_\lambda(\Theta),\Theta\rangle
=\frac2V\langle P^{E,U}-Q,EU^\top\rangle+\lambda\|\Theta\|_\fro^2$.
Thus, the Cauchy--Schwarz inequality and
$\|\Theta\|_\fro\le\sqrt{C_EV}$ give
\begin{equation}
 \left|
 \frac1V\langle P^{E,U}-Q,EU^\top\rangle
       +\frac\lambda2\|\Theta\|_\fro^2
 \right|
 \le\frac12\|\nabla\cF_\lambda(\Theta)\|_\fro\|\Theta\|_\fro
 \le\frac{\Delta_0^2}{1024B^2}.
 \label{eq:highgain-radial}
\end{equation}
Next, let $(\dot E,\dot U)$ be the direction in
\cref{lem:fisher-packing}. Since the cross-entropy loss is convex in
the logit matrix, and its gradient with respect to the logits at
$EU^\top$ is $(P^{E,U}-Q)/V$, we have
$\cR_{[V]}(E_\star,U_\star)\ge\cR_{[V]}(\Theta)
+\frac1V\langle P^{E,U}-Q,E_\star U_\star^\top-EU^\top\rangle$.
Since $E_\star U_\star^\top=t_\star\mathcal{L}(\log Q)$, this is
equivalent to
\[
 \frac1V\langle P^{E,U}-Q,t_\star \mathcal{L}(\log Q)\rangle
 \le\cR_{[V]}(E_\star,U_\star)-\cR_{[V]}(\Theta)
       +\frac1V\langle P^{E,U}-Q,EU^\top\rangle.
\]
We now bound the sum of the bilinear term and the weight decay term in
\cref{eq:factor-hessian} along $(\dot E,\dot U)$:
\begin{align*}
 &\frac2V\langle P^{E,U}-Q,\dot E\dot U^\top\rangle
       +\lambda(\|\dot E\|_\fro^2+\|\dot U\|_\fro^2)\\
 &\quad\overset{(a)}{\le}
 2\left\{\frac1V\langle P^{E,U}-Q,EU^\top\rangle
               +\frac\lambda2\|\Theta\|_\fro^2\right\}
 -2\{\cF_\lambda(\Theta)-\cF_\lambda(E_\star,U_\star)\}\\
 &\quad\overset{(b)}{\le}\frac{\Delta_0^2}{512B^2}-\frac{\Delta_0^2}{128B^2}
 =-\frac{3\Delta_0^2}{512B^2},
\end{align*}
where $(a)$ follows from
\cref{eq:packing-product,eq:packing-comparator-energy} and the
convexity inequality above, and $(b)$ follows from
\cref{eq:highgain-radial}, the assumption
$\cF_\lambda(\Theta)\ge\cH_{[V]}+\Delta_0/8-\Delta_0^2/(256B^2)$, and
\cref{eq:shrunken-comparator}.
Finally, we bound the softmax curvature term. Since $d<V$, we have
$4V/d+1\le5V/d$. Thus, \cref{eq:packing-fisher,eq:highgain-size-body}
imply that the softmax curvature term is at most
$40Br^{3/2}C_EV/d^2\le\Delta_0^2/(512B^2)$. Combining the above bounds
with \cref{eq:factor-hessian}, we obtain
\begin{equation*}
 \nabla^2\cF_\lambda(\Theta)
       [(\dot E,\dot U),(\dot E,\dot U)]
 \le-\frac{\Delta_0^2}{256B^2}.
\end{equation*}
Since the squared norm of $(\dot E,\dot U)$ is at most $2B\sqrt{r} V$ by
\cref{eq:packing-energy}, the minimum eigenvalue of the Hessian is at
most $-\Delta_0^2/(512B^3\sqrt r\,V)$, which is the second inequality in
\cref{eq:highgain-alternative}.
\end{proof}

Recall that an $\epsilon$-approximate SOSP is defined with a single
accuracy parameter $\epsilon$ for both the gradient and the Hessian.
The following corollary restates \cref{lem:highgain-collar} in terms of
$\epsilon$.
\begin{corollary}[Approximate SOSPs in the ball have small objective]
\label{cor:highgain-sosp}
Let $\rho\ge1$ be an upper bound on the Lipschitz constant of the Hessian, and suppose
\begin{equation}
 0<\epsilon\le\min\left\{
 \frac{\Delta_0^2}{512B^2\sqrt{C_EV}},
 \frac{\Delta_0^4}{2^{20}\rho B^6rV^2},1
 \right\}.
 \label{eq:highgain-epsilon}
\end{equation}
Then, under \cref{eq:highgain-size-body}, every point $\Theta$ with $\|\Theta\|_\fro^2\le C_EV$ and
\[
 \|\nabla \cF_\lambda(\Theta)\|\le\epsilon,\qquad
 \nabla^2\cF_\lambda(\Theta)\succeq-\sqrt{\rho\epsilon}\,I
\]
satisfies
\begin{equation*}
 \cF_\lambda(\Theta)<\cH_{[V]}+\frac{\Delta_0}{8}-\frac{\Delta_0^2}{256B^2}.
\end{equation*}
\end{corollary}

\begin{proof}
We prove the claim by contradiction. Suppose that
$\cF_\lambda(\Theta)\ge\cH_{[V]}+\Delta_0/8-\Delta_0^2/(256B^2)$. Then,
by \cref{lem:highgain-collar}, one of the two inequalities in
\cref{eq:highgain-alternative} holds. The first one contradicts
$\|\nabla\cF_\lambda(\Theta)\|\le\epsilon$ and the first bound in
\cref{eq:highgain-epsilon}. The second one contradicts
$\nabla^2\cF_\lambda(\Theta)\succeq-\sqrt{\rho\epsilon}\,I$, because the
second bound in \cref{eq:highgain-epsilon} implies
$\sqrt{\rho\epsilon}\le\Delta_0^2/(1024B^3\sqrt r\,V)$.
\end{proof}

In summary, Step 2 shows that every $\epsilon$-approximate SOSP in the
ball has objective below $\cH_{[V]}+\Delta_0/8$ by the margin
$\Delta_0^2/(256B^2)$.

\subsection{Step 3: Refresh the Gain throughout the Final Window}
\label{sec:untied-step3}

In this step, we combine Step 1 and Step 2 to prove
\cref{thm:iclr-untied}. The idea is as follows. By Step 2, whenever the
trajectory visits an approximate SOSP in the ball, the objective is
below $\cH_{[V]}+\Delta_0/8$ by the margin $\Delta_0^2/(256B^2)$. We
partition the iterations into $d$ blocks, and choose the step size, the
perturbation scale, and the length of the blocks such that, with high
probability, each block contains a visit to an approximate SOSP, and
the objective increases by at most half of this margin within the block
(\cref{thm:tied-fixed-noise-consequence} in
Appendix~\ref{app:optimization-tools}). Then, the first block
establishes the margin at its last iterate, and each later block keeps
the objective below $\cH_{[V]}+\Delta_0/8$ and restores the margin at
its last iterate. Since the results on perturbed gradient descent
require global smoothness, they are applied to the smooth extension in
Step 1. The induction also shows that the trajectory never leaves the
ball, so that it coincides with the trajectory under the extension.
For convenience, we restate \cref{thm:iclr-untied} below.

\begin{theorem}[Restatement of \Cref{thm:iclr-untied}: Untied constant gain]
Fix $r,B,\Delta_0$. There are positive constants $c,C,V_0$ such that,
for every $V\ge V_0$, $Q\in\mathcal Q_{r,B,\Delta_0}$, and
$\max\{2r,C\sqrt V\}\le d<V-1$, setting the weight decay
$\lambda=\Delta_0/(8B\sqrt r\,V)$, there exist a step size $h$, a perturbation scale $\sigma$, and a number of iterations $T$
divisible by $d$ such that
\begin{equation*}
 \Pr\left\{g_{[V]}(\Theta_t)>\frac78
 \text{ for every integer }t\in[T/d,T]\right\}\ge1-e^{-cd}.
\end{equation*}
The quantities $h^{-1},\sigma^{-1}$ and $T$ are bounded by polynomials
in $V$. The probability is over both initialization and perturbations.
\end{theorem}

\begin{proof}
\textbf{Parameter choices.}
Let $\ell,\rho\ge1$ be the Lipschitz constants of the gradient and the
Hessian of the smooth extension in Step 1, which satisfy $\ell=O(1)$
and $\rho=O(\sqrt V)$ uniformly over the class
$\mathcal Q_{r,B,\Delta_0}$. Let $C_E$ be the constant in Step 1, i.e.,
\[
 C_E=6+\frac{32B\sqrt r(B+3)}{\Delta_0},
\]
so that, by \cref{eq:highgain-wall-separation}, the ball
$\|\Theta\|_\fro^2<C_EV$ contains the initialization, and the objective
outside the ball is at least $\cF_\lambda(\Theta_0)+4$. For the
accuracy, we choose
\[
 \epsilon=c_{\rm acc}V^{-5/2},
\]
where $0<c_{\rm acc}=c_{\rm acc}(B,r,\Delta_0)\le1$ is a sufficiently
small constant such that \cref{eq:highgain-epsilon} holds. Such a
constant exists since $\rho/\sqrt V$ is uniformly bounded. For the
embedding dimension, we choose
$C_{\rm wid}=C_{\rm wid}(B,r,\Delta_0)\ge1$ sufficiently large such
that, for all $V\ge2$,
\[
 C_{\rm wid}\sqrt V\ge
 \max\left\{2r,C_{\rm init}\log(eV),
 \sqrt{\frac{20480B^3r^{3/2}C_EV}{\Delta_0^2}}\right\}.
\]
Such a choice exists because $\log(eV)/\sqrt V$ is bounded. Then, the
condition $d\ge C_{\rm wid}\sqrt V$ guarantees that both
\cref{lem:untied-gaussian-initialization} and \cref{cor:highgain-sosp}
are applicable.

Next, we specify the quantities in
\cref{thm:tied-fixed-noise-consequence}. We set
\[
 \begin{gathered}
 D_\Theta=2Vd,\qquad
 A=2B+2+\frac{3\Delta_0}{8B\sqrt r},\\
 \delta_c=\frac{\Delta_0^2}{512B^2},\qquad
 \zeta=\frac14e^{-d}.
 \end{gathered}
\]
Here, $A$ is the bound on the initial objective gap in
\cref{lem:untied-gaussian-initialization}, and $\delta_c$ is half of the
margin in \cref{cor:highgain-sosp}. These quantities satisfy the
assumptions of \cref{thm:tied-fixed-noise-consequence}, i.e., $A\ge1$,
$0<\epsilon\le1$, and $0<\delta_c\le1/512$, where the last one is due
to $\Delta_0\le B$. The step size, the perturbation scale, and the
number of iterations are chosen as
\begin{equation}
 \begin{aligned}
 \iota&=C_{\rm sched}d,
 &h&=(\iota^9\ell)^{-1},
 &\sigma&=\iota\epsilon,\\
 T_{\rm blk}
 &=\left\lceil
 C_J\ell\left(2B+2+\frac{3\Delta_0}{8B\sqrt r}\right)
 \epsilon^{-2}\iota^{15}
 \right\rceil,
 &&T=dT_{\rm blk},
 \end{aligned}
 \label{eq:highgain-schedule}
\end{equation}
where $T_{\rm blk}$ is the length of each block. With the above
choices, the lower bound on $\iota$ required in
\cref{eq:predeclared-schedule} is $O(d+\log(eV))=O(d)$, and
$\log(T_{\rm blk}+1)=O(\log(eV)+\log C_{\rm sched})$. Hence, a
sufficiently large constant
$C_{\rm sched}=C_{\rm sched}(B,r,\Delta_0)$ satisfies both this lower
bound and $4(T_{\rm blk}+1)^2e^{-\iota}\le e^{-d}/4$. Note that all the
above quantities are deterministic and chosen before training, and the
same $h$ and $\sigma$ are used in all blocks.

\textbf{Coupling with the smooth extension.}
We condition on the event in \cref{eq:untied-initialization-event}. Let
$\widetilde{\cF}_\lambda$ be the smooth extension of $\cF_\lambda$ in
Step 1. Recall that $\widetilde{\cF}_\lambda$ has the same minimum as
$\cF_\lambda$, coincides with $\cF_\lambda$ inside the ball
$\|E\|_\fro^2+\|U\|_\fro^2<C_EV$, and satisfies
\begin{equation}
 \widetilde{\cF}_\lambda
 \ge\cH_{[V]}+2B+2+\frac{3\Delta_0}{8B\sqrt r}+4
 \quad\text{outside this ball}.
 \label{eq:highgain-extended-wall}
\end{equation}
We define an auxiliary trajectory $\{\widetilde\Theta_t\}$, which runs
perturbed gradient descent on $\widetilde{\cF}_\lambda$ with the same
initialization and the same perturbations $\{\xi_t\}$ as the original
trajectory in \cref{eq:appendix-noisy-gd}:
\[
 \widetilde\Theta_0=\Theta_0,\qquad
 \widetilde\Theta_{t+1}
 =\widetilde\Theta_t-h\{
 \nabla\widetilde{\cF}_\lambda(\widetilde\Theta_t)+\xi_t\}.
\]
Since the gradients of $\widetilde{\cF}_\lambda$ and $\cF_\lambda$
coincide inside the ball, a simple induction on $t$ shows that
$\widetilde\Theta_t=\Theta_t$ as long as the auxiliary trajectory has
not left the ball.

\textbf{One-block estimate.}
We partition the iterations $\{0,1,\ldots,T\}$ into $d$ blocks of length
$T_{\rm blk}$, where consecutive blocks share their boundary iterate.
Consider any block, and condition on the history up to the first
iterate of the block. Suppose that the objective gap
$\widetilde{\cF}_\lambda(\widetilde\Theta_t)-\cF_\lambda^*$ at the first
iterate of the block is at most $A=2B+2+3\Delta_0/(8B\sqrt r)$. Then,
by \cref{thm:tied-fixed-noise-consequence}, with conditional
probability at least $1-e^{-d}/2$, the block contains an iterate that
is an $\epsilon$-approximate SOSP of $\widetilde{\cF}_\lambda$, and
\begin{equation}
 \widetilde{\cF}_\lambda(\widetilde\Theta_v)
 -\widetilde{\cF}_\lambda(\widetilde\Theta_u)
 \le\frac{\Delta_0^2}{512B^2}
 \quad\text{for every integer pair }u<v\text{ in the block}.
 \label{eq:highgain-block-heating}
\end{equation}
We say that a block is \emph{successful} if both events occur. Since
\cref{eq:highgain-block-heating} holds for all pairs of iterates in the
block, it can be applied with $u$ being the random iteration at which
the approximate SOSP is visited. In the following, we show by induction
that, as long as all blocks so far are successful, the condition on the
objective gap holds at the first iterate of the next block, and the two
trajectories coincide.

\textbf{Base case: the first block.}
By \cref{lem:untied-gaussian-initialization}, the objective gap at
$\Theta_0$ is at most $A$, and thus the one-block estimate applies to
the first block. Suppose that the first block is successful. Applying
\cref{eq:highgain-block-heating} with $u=0$, every auxiliary iterate in
the block satisfies
$\widetilde{\cF}_\lambda(\widetilde\Theta_t)<\cF_\lambda(\Theta_0)+1
\le\cH_{[V]}+A+1$, which is strictly smaller than the lower bound in
\cref{eq:highgain-extended-wall}. Hence, every auxiliary iterate in the
block lies inside the ball, and $\widetilde\Theta_t=\Theta_t$ for all
$t$ in the block. In particular, the $\epsilon$-approximate SOSP visited
by the auxiliary trajectory, say at iteration $\tau_0$, is also an
$\epsilon$-approximate SOSP of the original objective $\cF_\lambda$ in
the ball. By \cref{cor:highgain-sosp}, we have
\[
 \cF_\lambda(\Theta_{\tau_0})
 <\cH_{[V]}+\frac{\Delta_0}{8}-\frac{\Delta_0^2}{256B^2}.
\]
Applying \cref{eq:highgain-block-heating} with $u=\tau_0$ and $v=T_{\rm blk}$, we obtain
\begin{equation}
 \cF_\lambda(\Theta_{jT_{\rm blk}})
 <\cH_{[V]}+\frac{\Delta_0}{8}-\frac{\Delta_0^2}{512B^2}
 \qquad (j=1).
 \label{eq:highgain-refresh-margin}
\end{equation}
If $\tau_0=T_{\rm blk}$, \cref{eq:highgain-refresh-margin} follows
directly from the bound at $\tau_0$.

\textbf{Induction step: the later blocks.}
Suppose that the first $j$ blocks are successful for some $1\le j<d$.
The induction hypothesis consists of two parts: (H1) the two
trajectories stay inside the ball and coincide up to iteration
$jT_{\rm blk}$; (H2) \cref{eq:highgain-refresh-margin} holds at
iteration $jT_{\rm blk}$. Both parts hold for $j=1$ by the base case.
We show that, if the $(j+1)$-th block is also successful, then (H1) and
(H2) hold with $j$ replaced by $j+1$, and the gain is above $7/8$ at
every iterate in the $(j+1)$-th block.

First, by (H2) and $\cF_\lambda^*\ge\cH_{[V]}$, the objective gap at the
first iterate of the $(j+1)$-th block satisfies
\[
 \widetilde{\cF}_\lambda(\widetilde\Theta_{jT_{\rm blk}})
 -\cF_\lambda^*
 <\frac{\Delta_0}{8}
 \le 2B+2+\frac{3\Delta_0}{8B\sqrt r}.
\]
Thus, the one-block estimate applies to the $(j+1)$-th block
conditionally on the history. Suppose that this block is successful.
Then, by \cref{eq:highgain-block-heating} with $u=jT_{\rm blk}$ and
(H2), every iteration $t$ in the block satisfies
\[
 \widetilde{\cF}_\lambda(\widetilde\Theta_t)
 \le\cF_\lambda(\Theta_{jT_{\rm blk}})
       +\frac{\Delta_0^2}{512B^2}
 <\cH_{[V]}+\frac{\Delta_0}{8}.
\]
Since $\Delta_0\le B$, the right-hand side is smaller than the lower
bound in \cref{eq:highgain-extended-wall}. Hence, the auxiliary
trajectory stays inside the ball, and $\widetilde\Theta_t=\Theta_t$ for
all $t$ in the block, which proves (H1) for $j+1$. Moreover, since the
weight decay penalty is nonnegative,
\[
 \cR_{[V]}(\Theta_t)-\cH_{[V]}
 \le\cF_\lambda(\Theta_t)-\cH_{[V]}
 <\frac{\Delta_0}{8}
 \le\frac{\log V-\cH_{[V]}}8,
\]
which implies $g_{[V]}(\Theta_t)>7/8$ for all $t$ in the block. Finally,
by the same argument as in the base case, i.e., applying
\cref{cor:highgain-sosp} at the approximate SOSP visited in this block
and then \cref{eq:highgain-block-heating}, we conclude that
\cref{eq:highgain-refresh-margin} holds with $j$ replaced by $j+1$,
which proves (H2) for $j+1$. By induction, if all $d$ blocks are
successful, then $g_{[V]}(\Theta_t)>7/8$ for every integer
$t\in[T_{\rm blk},T]$.

\textbf{Failure probability and running time.}
Conditional on the initialization event and on the success of all
previous blocks, each block fails with probability at most $e^{-d}/2$.
Thus, by the tower property and the union bound, which do not require
the blocks to be independent, we have
\[
 \Pr\left\{\exists t\in[T_{\rm blk},T]:
                 g_{[V]}(\Theta_t)\le\frac78\right\}
 \le e^{-c_0d}+\frac d2e^{-d}\le e^{-c_{\rm u}d},
\]
for a universal constant $c_{\rm u}>0$, provided that $V\ge V_0$ for a
sufficiently large $V_0=V_0(B,r,\Delta_0)$. Since $T_{\rm blk}=T/d$,
this proves the probability bound in the theorem with $c=c_{\rm u}$ and
$C=C_{\rm wid}$. For the running time, \cref{eq:highgain-schedule} and
$\ell=O(1)$ give
\[
 h^{-1}=O(d^9),\qquad
 \sigma^{-1}=O(V^{5/2}/d),\qquad
 T_{\rm blk}\le C_TV^5d^{15},\qquad
 T\le C_TV^5d^{16}\le C_TV^{21},
\]
where $C_T$ and the hidden constants depend only on $(B,r,\Delta_0)$.
Thus, $h^{-1}$, $\sigma^{-1}$, and $T$ are bounded by polynomials in
$V$. We also note that, on the same event, only the first
$T_{\rm blk}$ iterates may have gain below $7/8$. Hence, the fraction of
iterates with gain above $7/8$ satisfies
\[
 \frac1{T+1}\sum_{t=0}^T
 \mathbf 1\!\left\{g_{[V]}(\Theta_t)>\frac78\right\}
 \ge1-\frac{T_{\rm blk}}{dT_{\rm blk}+1}>1-\frac1d.
\]
\end{proof}

\subsection{Comparison with a Generic Factorized Bound}
\label{app:comparison}

In this subsection, we compare our analysis in Step 2 with a direct
application of the generic theory for factorized optimization
\citep{ouyang2025burer}, and explain why the latter requires a larger
embedding dimension. For simplicity, consider an exact SOSP
$\Theta=(E,U)$ of the untied objective $\cF_\lambda$ with $\lambda>0$
and $\|\Theta\|_\fro^2\le C_EV$. By \cref{lem:factor-calculus}, at such
a point, we have $\1^\top U=0$, and the weight decay penalty equals
$\lambda\|EU^\top\|_*$. Then, for any comparator matrix $M$ with
$M\1=0$, Theorem~4.1 of \citet{ouyang2025burer}, together with the
convexity of the loss and the stationarity of $\Theta$, gives
\begin{align*}
 &\cF_\lambda(\Theta)
 -\left\{\frac1V\sum_i\CE(q_i,M_{i,:})+\lambda\|M\|_*\right\}\\
 &\quad\le
 \left\{\frac1V
 \sup_{\|H\|_\fro=1,\,\rank H=2}
 \sum_i\operatorname{Var}_{j\sim P^{E,U}_{i,:}}(H_{ij})\right\}
 \sigma_d(EU^\top)\|M\|_*\\
 &\quad\le\frac{\sigma_d(EU^\top)\|M\|_*}{2V}.
\end{align*}
Here, the expression in the braces is the largest curvature of the loss
along a rank-two direction in the logit space, which is at most
$1/(2V)$ for categorical distributions. Moreover, we have
$\sigma_d(EU^\top)\le C_EV/(2d)$, and when the comparator is given by
the ground-truth logits, which have rank at most $r$ and entries bounded
by $B$, we have $\|M\|_*\le B\sqrt r\,V$. Therefore, the above bound
only gives a comparison error of $O(V/d)$, which is not small when
$d\asymp\sqrt V$.

In contrast, our analysis uses the sum of the traces of the Hessians
over all rows, instead of a uniform bound on the curvature along each
direction. This leads to the bound $O(V/d^2+1/d)$ in
\cref{eq:packing-fisher}, which is sufficiently small for the argument
in Step 2 as long as $d\ge C\sqrt V$. We note that this comparison only
explains the improvement in our sufficient condition on $d$, and it is
not a lower bound on the embedding dimension required by other
methods.

\begin{remark}[Prediction loss versus the regularized objective]
\label{rem:which-global-optimum}
We note that a global minimizer of $\cF_\lambda$ is not Bayes-optimal.
Indeed, when $\lambda>0$ and $\log V-\cH_{[V]}>0$, the centered
ground-truth logits $\mathcal{L}(\log Q)$ are nonzero, and the
derivative of
$t\mapsto V^{-1}\sum_i\CE(q_i,t\mathcal{L}(\log Q)_{i,:})+\lambda t\|\mathcal{L}(\log Q)\|_*$
at $t=1$ equals $\lambda\|\mathcal{L}(\log Q)\|_*>0$. Thus, slightly
shrinking the ground-truth logits decreases the regularized objective.
For this reason, \cref{thm:iclr-untied} is stated as a guarantee on the
prediction gain, and not as the convergence to the Bayes predictor.
\end{remark}

\section{Tied Training and Transfer}

\subsection{Proof of Theorem~\ref{thm:iclr-tied-training}}
\label{app:tied-proofs}

In this section, we present the full proof of
\cref{thm:iclr-tied-training}. The proof consists of three main steps,
which are parallel to those in Appendix~\ref{sec:positive}.

\textbf{Step 1 (Appendix~\ref{sec:tied-step1}).}
We bound the objective at the Gaussian initialization, and construct a
smooth extension of the objective that coincides with it in a ball and
is large outside the ball (\cref{lem:tied-localization-inputs}).

\textbf{Step 2 (Appendix~\ref{sec:tied-step2}).}
We compare the objective at an approximate SOSP with the objective at
the ground truth. Specifically, we construct a direction that
represents the ground-truth Gram matrix, along which the softmax
curvature term of the Hessian is at most $O(\|X\|_\fro^2/d^2)$
(\cref{lem:tied-teacher-packing}). Then, we use the weight decay to
absorb this term, which gives $\|X\|_\fro^2=O(V)$ and a comparison
error of $O(V/d^2)$ (\cref{lem:tied-sosp-energy}).

\textbf{Step 3 (Appendix~\ref{sec:tied-step3}).}
We apply the one-block estimate
(\cref{thm:tied-fixed-noise-consequence}) to each of the $d$ blocks of
iterations, and prove by induction that the comparison in Step 2 holds
at every iterate in the final window
(\cref{lem:tied-final-window-comparison}). Finally, we specify the
weight decay and prove \cref{thm:iclr-tied-training}.

We remark that the lemmas in this section are stated for a general
weight decay $\lambda$, with the bounds expressed in terms of
$\lambda V$. Thus, they can also be used in the proof of
\cref{thm:iclr-transfer} in Appendix~\ref{app:transfer-proof}, where a
different $\lambda$ is chosen.

\paragraph{Notation.}
Recall that $\mathcal S\subset[V]$ is the set of observed current
states, and $\mathcal U=[V]\setminus\mathcal S$ is the set of unseen
states. Throughout this section, we assume
\[
 |\mathcal S|=N=\Theta(V),\qquad |\mathcal U|=V-N,\qquad 1\le d<N<V,
\]
where the constants in $N=\Theta(V)$ are fixed. In particular, $d<V-1$.
Note that all $V$ states, including those in $\mathcal U$, are possible
next states of every observed current state. For a matrix $A$, the
subscripts $_{\mathcal S,:}$ and $_{:,\mathcal S}$ select the rows and
the columns indexed by $\mathcal S$, respectively. In particular,
$(I_V)_{:,\mathcal S}$ consists of the columns of $I_V$ indexed by
$\mathcal S$, and multiplying an $N$-row matrix by it from the left
extends the matrix by zero rows outside $\mathcal S$. We use
$\|A\|_\infty$ to denote the maximum absolute entry of $A$, and
$\|A\|_1=\sum_{ij}|A_{ij}|$. For the tied model, we write $x_i$ for the
$i$-th row of $X$ as a column vector, and $K=XX^\top$ for the learned
Gram matrix, so that $P^X=\softmax(XX^\top)=\softmax(\mathcal{L}(K))$.
All $Vd$ entries of $X$ are trained. The risk, the gain, and the
objective are defined as in Appendix~\ref{sec:model}. In particular,
\[
 \cR_{\mathcal A}(X)-\cH_{\mathcal A}
 =\frac1{|\mathcal A|}\sum_{i\in\mathcal A}
       \KL(q_i\|P^X_{i,:}),\qquad
 \cF_{\lambda,\mathcal S}(X)
 =\cR_{\mathcal S}(X)+\frac\lambda2\|X\|_\fro^2.
\]
We assume that the information gap on the observed states satisfies
$\log V-\cH_{\mathcal S}\ge\delta_{\rm tr}>0$ and, when the gain on the
unseen states is considered,
$\log V-\cH_{\mathcal U}\ge\delta_{\rm te}>0$. Note that these
conditions are imposed on the averages over $\mathcal S$ and
$\mathcal U$, and not on each individual state.

\paragraph{Ground truth.}
Let $Q\in\mathcal Q^{\rm tied}_{r,B,\Delta_0}$ with the constants
$B_Y\ge1$ and $r\ge1$ in \cref{def:appendix-tied-class}. We fix a
representative $Y$ of $Q$ with linearly independent columns, i.e.,
\begin{equation*}
 \begin{gathered}
 K^\star=YY^\top,\qquad \mathcal{L}(\log Q)=\mathcal{L}(K^\star),
 \\ Y\in\R^{V\times r_Y},\quad
 1\le r_Y\le r,\quad \max_i\|y_i\|_2^2\le B_Y.
 \end{gathered}
\end{equation*}
Such a representative always exists. Indeed, if the columns of a
representative $Y$ in \cref{def:appendix-tied-class} are linearly
dependent, we can write $Y=Y'R^\top$, where $Y'$ has linearly
independent columns and $R$ has orthonormal columns. Then, $Y'$ has the
same Gram matrix and the same row norms as $Y$. The bound on the row
norms implies $\|\mathcal{L}(\log Q)\|_\infty\le2B_Y$. Note that the
centering condition $Y^\top\1=0$ and the coverage condition are not
required in this section. The objective at the ground-truth embedding
matrix $Y$, padded with zero columns, is
$\cH_{\mathcal S}+\frac\lambda2\|Y\|_\fro^2$, which serves as the
reference value in the comparisons below.

\paragraph{Parameters and algorithm.}
Let $C_p=40B_Yr$. Throughout this section, we assume
$2r_Y\le d<N<V$, $d\ge C\log(eV)$, $\lambda\ge2C_p/d^2$, and
$0<\lambda V\le1$. The entries of $X_0$ are independent
$\mathcal N(0,1/d)$ random variables, and the embeddings are updated by
\begin{equation*}
 X_{t+1}=X_t-h\{\nabla \cF_{\lambda,\mathcal S}(X_t)+\xi_t\},\qquad
 \operatorname{vec}(\xi_t)\stackrel{\rm iid}{\sim}
 \mathcal N\left(0,\frac{\sigma^2}{Vd}I_{Vd}\right).
\end{equation*}
The step size $h$ and the perturbation scale $\sigma$ are chosen before
training and fixed throughout training.

\subsubsection{Step 1: Localize the Trajectory}
\label{sec:tied-step1}

In this step, we show that the Gaussian initialization has bounded
objective with high probability, and that the objective is large
outside a ball with a deterministic radius $R_w$, due to the weight
decay. Since the results on perturbed gradient descent require global
smoothness, we also construct a smooth extension of the objective
outside the ball. We note that the radius $R_w$ in the following lemma
depends on $\lambda V$, and can be much larger than $\sqrt V$. A sharper
bound $\|X\|_\fro^2=O(V)$ at approximate SOSPs will be established in
Step 2.
\begin{lemma}[Initialization and smooth extension]
\label{lem:tied-localization-inputs}
For $d\ge C\log(eV)$, with probability at least $1-e^{-cd}$,
\begin{equation}
 \max_i\|(X_0)_i\|_2^2\le2,\qquad
 \|X_0\|_\fro^2\le2V,\qquad
 \cF_{\lambda,\mathcal S}(X_0)-\cH_{\mathcal S}\le2B_Y+8+(\lambda V).
 \label{eq:initialization-event}
\end{equation}
Moreover, let $A=4B_Y+20+C_p^{-1}$ and
\begin{equation*}
 C_w=1+\max\{2,B_Y,2(A+4)/(\lambda V)\},\qquad R_w^2=C_wV.
\end{equation*}
Then, there exists a smooth function
$\widetilde{\cF}_{\lambda,\mathcal S}:\R^{Vd}\to\R$ that coincides with
$\cF_{\lambda,\mathcal S}$ on the ball $\|X\|_\fro\le R_w$, has the same
global minimum as $\cF_{\lambda,\mathcal S}$, and satisfies
$\widetilde{\cF}_{\lambda,\mathcal S}>\cH_{\mathcal S}+A+4$ outside the
ball. Moreover, the Lipschitz constants
$\ell_{\rm ext},\rho_{\rm ext}\ge1$ of its gradient and its Hessian can
be chosen to be polynomial in $V$, $N$, and $(\lambda V)^{-1}$.
\end{lemma}

\begin{proof}
We first prove \cref{eq:initialization-event}. For each row $i$, the
random variable $d\|(X_0)_i\|_2^2$ follows the $\chi_d^2$ distribution.
Thus, the Chernoff bound and the union bound over the $V$ rows give
\[
 \Pr\{\max_i\|(X_0)_i\|^2>2\}\le Ve^{-c_0d},
\]
for a universal constant $c_0>0$. Since $d\ge C\log(eV)$, the right-hand
side is at most $e^{-cd}$ after adjusting the constants. On this event,
we have $\|X_0\|_\fro^2\le2V$, and the entries of $X_0X_0^\top$ are at
most two in absolute value. Thus, the entries of the centered logits
are at most four in absolute value, which implies
$\cR_{\mathcal S}(X_0)\le\log V+8$. Moreover, since the entries of
$YY^\top$ lie in $[-B_Y,B_Y]$, we have $q_{ij}\le e^{2B_Y}/V$, and thus
$\log V+\sum_jq_{ij}\log q_{ij}\le2B_Y$ for every $i$, i.e.,
$\log V-\cH_{\mathcal S}\le2B_Y$. Combining these two bounds with the
penalty $\frac\lambda2\|X_0\|_\fro^2\le\lambda V$ gives the last bound
in \cref{eq:initialization-event}.

Next, we show that the objective is large outside the ball. Since the
cross-entropy loss is lower bounded by $\cH_{\mathcal S}$ and the weight
decay penalty is coercive, a global minimizer of
$\cF_{\lambda,\mathcal S}$ exists. Comparing it with the ground-truth
embedding matrix $Y$ padded with zero columns, whose squared norm is at
most $B_YV$, we conclude that every global minimizer has squared norm
at most $B_YV<R_w^2$, and thus lies in the ball. On the other hand, for
any $X$ with $\|X\|_\fro\ge R_w$, the definition of $C_w$ gives
$\cF_{\lambda,\mathcal S}(X)\ge \cH_{\mathcal S}+\lambda R_w^2/2>\cH_{\mathcal S}+A+4$.

We now construct the extension. Let $s:[0,\infty)\to[0,2R_w]$ be a
smooth function such that $s(a)=a$ for $a\le R_w$, $s(a)\ge R_w$ for
$a\ge R_w$, and $s$ is constant for $a\ge3R_w$. Such a function can be
chosen with its $k$-th derivative bounded by a constant multiple of
$R_w^{1-k}$. For $X\ne0$, we define
\[
 \widetilde{\cF}_{\lambda,\mathcal S}(X)
 =\cF_{\lambda,\mathcal S}
       \left(s(\|X\|_\fro)\frac{X}{\|X\|_\fro}\right),
\]
and set $\widetilde{\cF}_{\lambda,\mathcal S}(0)
=\cF_{\lambda,\mathcal S}(0)$.
By construction, $\widetilde{\cF}_{\lambda,\mathcal S}$ coincides with
$\cF_{\lambda,\mathcal S}$ on the ball $\|X\|_\fro\le R_w$. Outside the
ball, $\widetilde{\cF}_{\lambda,\mathcal S}(X)$ is the value of
$\cF_{\lambda,\mathcal S}$ at a point with norm at least $R_w$, which is
larger than $\cH_{\mathcal S}+A+4$. Since all global minimizers of
$\cF_{\lambda,\mathcal S}$ lie in the ball,
$\widetilde{\cF}_{\lambda,\mathcal S}$ has the same global minimum.

It remains to bound the Lipschitz constants. For any directions
$Z,W\in\R^{V\times d}$, the first and second variations of the centered
logits of the observed rows satisfy
\[
 \begin{aligned}
 \|\mathcal{L}(XZ^\top+ZX^\top)_{\mathcal S,:}\|_\fro
 &\le2\|X\|_\fro\|Z\|_\fro,\\
 \|\mathcal{L}(ZW^\top+WZ^\top)_{\mathcal S,:}\|_\fro
 &\le2\|Z\|_\fro\|W\|_\fro.
 \end{aligned}
\]
In addition, the first, second, and third derivatives of the average
cross-entropy loss with respect to the logits have norms at most
$\sqrt{2/N}$, $1/(2N)$, and $C/N$, respectively. Thus, by the chain
rule, on the ball of radius $2R_w$, which contains the image of the map
$X\mapsto s(\|X\|_\fro)X/\|X\|_\fro$, the Lipschitz constants of the
gradient and the Hessian of $\cF_{\lambda,\mathcal S}$ are at most
$C(1+R_w^2/N)+\lambda$ and $C(1+(R_w^3+R_w)/N)$, respectively. Applying
the chain rule again to the composition with this map, whose
derivatives are bounded as above, we conclude that $\ell_{\rm ext}$ and
$\rho_{\rm ext}$ are polynomial in $V$, $N$, and $(\lambda V)^{-1}$.
\end{proof}

\subsubsection{Step 2: Convert Approximate Stationarity into Gain}
\label{sec:tied-step2}

In this step, we compare the objective at an approximate SOSP with the
objective at the ground truth, i.e.,
$\cH_{\mathcal S}+\frac\lambda2\|Y\|_\fro^2$. The argument is parallel
to Step 2 in Appendix~\ref{sec:untied-step2}, with two differences.
First, since the input and output embeddings are tied, a perturbation
of $X$ changes both of them, and thus the gradient and the Hessian take
a different form. Second, the softmax curvature term along the
constructed direction is proportional to $\|X\|_\fro^2/d^2$, where
$\|X\|_\fro^2$ is not known to be $O(V)$ in advance. We handle the
second issue in \cref{lem:tied-sosp-energy} by using the weight decay.

The following lemma provides the gradient and the Hessian of the tied
objective, and constructs the direction by the same packing argument as
in \cref{lem:fisher-packing}.
\begin{lemma}[Derivatives of the tied objective and comparison with the ground truth]
\label{lem:tied-teacher-packing}
Let $e_i$ denote the $i$th coordinate vector in $\R^V$ and let
$x_i$ denote row $i$ of $X$ as a column vector. For every direction $Z$,
\begin{align}
 \nabla\cF_{\lambda,\mathcal S}(X)
 &=\frac1N\sum_{i\in\mathcal S}\sum_{j=1}^V
      (P^X_{ij}-q_{ij})(e_ix_j^\top+e_jx_i^\top)+\lambda X,
      \nonumber\\
 \nabla^2\cF_{\lambda,\mathcal S}(X)[Z,Z]
 &=\frac1N\sum_{i\in\mathcal S}\operatorname{Var}_{P^X_{i,:}}
              ((XZ^\top+ZX^\top)_{i,:})\nonumber\\
 &\quad+\frac2N\sum_{i\in\mathcal S}\sum_{j=1}^V
      (P^X_{ij}-q_{ij})(ZZ^\top)_{ij}+\lambda\|Z\|_\fro^2,
      \nonumber\\
 \frac1N\sum_{i\in\mathcal S}\sum_{j=1}^V(P^X_{ij}-q_{ij})K_{ij}
       +\frac\lambda2\|X\|_\fro^2
 &=\tfrac12\langle\nabla\cF_{\lambda,\mathcal S}(X),X\rangle.
 \label{eq:tied-differential-identities}
\end{align}
We refer to the first term on the right-hand side of the second
identity as the softmax curvature term. Under
\cref{def:appendix-tied-class}, if $d\ge2r_Y$, there exists a direction
$Z\in\R^{V\times d}$ with $ZZ^\top=K^\star$ and
$\|Z\|_\fro^2\le B_YV$, along which the softmax curvature term is at
most $C_p\|X\|_\fro^2/d^2$, where $C_p=40B_Yr$.
Consequently, if $\|\nabla \cF_{\lambda,\mathcal S}(X)\|_\fro\le\eta$ and
$\nabla^2\cF_{\lambda,\mathcal S}(X)\succeq-\gamma I$, then
\begin{equation}
 \cF_{\lambda,\mathcal S}(X)-\left(\cH_{\mathcal S}+\frac\lambda2\|Y\|_\fro^2\right)
 \le\frac{C_p\|X\|_\fro^2}{2d^2}
      +\frac{\eta\|X\|_\fro}{2}+\frac{\gamma B_YV}{2}.
 \label{eq:tied-teacher-sosp-comparison}
\end{equation}
\end{lemma}

\begin{proof}
We first prove \cref{eq:tied-differential-identities}. The first and
second variations of $XX^\top$ along $Z$ are $XZ^\top+ZX^\top$ and
$2ZZ^\top$, respectively. Since the gradient of the cross-entropy loss
of row $i$ with respect to its logits is $P^X_{i,:}-q_i^\top$, and its
Hessian is the covariance matrix of $P^X_{i,:}$, the chain rule gives
the first two identities. Here, we use the fact that the variance is
invariant to adding a constant, so that the centering of the logits
does not affect the Hessian. The third identity follows from taking the
inner product of the gradient with $X$, since
$\langle e_ix_j^\top+e_jx_i^\top,X\rangle=2K_{ij}$.

Next, we construct the direction $Z$ by the same packing argument as in
the proof of \cref{lem:fisher-packing}. Let $w_1,\ldots,w_d$ be an
orthonormal eigenbasis of $X^\top X$. Then, the vectors
$Xw_j/\|Xw_j\|_2$ with $Xw_j\ne0$ are orthonormal. Since the covariance
matrix of each row has trace at most one, we have
\[
 \sum_{j=1}^d\|Xw_j\|_2^2=\|X\|_\fro^2,\qquad
 \sum_{j:Xw_j\ne0}\sum_{i\in\mathcal S}
 \frac{\operatorname{Var}_{P^X_{i,:}}(Xw_j)}{\|Xw_j\|_2^2}
 \le N.
\]
Consequently, at least $d/2$ indices $j$ satisfy both of the following bounds:
\[
 \|Xw_j\|_2^2\le\frac{4\|X\|_\fro^2}{d},\qquad
 \sum_{i\in\mathcal S}\operatorname{Var}_{P^X_{i,:}}(Xw_j)
 \le\frac{4N}{d}\|Xw_j\|_2^2.
\]
Since $d\ge2r_Y$, we can choose $r_Y$ distinct indices
$j_1,\ldots,j_{r_Y}$ among them. Let $y^{(k)}$ be the $k$-th column of
$Y$, and define
$Z=\sum_{k=1}^{r_Y}\varepsilon_k y^{(k)}w_{j_k}^\top$ with signs
$\varepsilon_k\in\{-1,1\}$. For every choice of the signs, the
orthonormality of $\{w_{j_k}\}$ gives $ZZ^\top=YY^\top=K^\star$ and
$\|Z\|_\fro^2=\|Y\|_\fro^2\le B_YV$. To choose the signs, suppose that
they are independent and uniformly distributed. Then, in the
expectation of the softmax curvature term
$\frac1N\sum_{i\in\mathcal S}\operatorname{Var}_{P^X_{i,:}}((XZ^\top+ZX^\top)_{i,:})$,
the cross terms between different columns vanish, and the contribution
of the $k$-th column is at most
\[
 \begin{aligned}
 &\frac{2B_Y}{N}\left\{\|Xw_{j_k}\|_2^2
       +\sum_{i\in\mathcal S}
          \operatorname{Var}_{P^X_{i,:}}(Xw_{j_k})\right\}\\
 &\hspace{2cm}\le\frac{8B_Y\|X\|_\fro^2}{Nd}(1+4N/d).
 \end{aligned}
\]
Here, we use
$\operatorname{Var}(a+b)\le2\operatorname{Var}(a)+2\operatorname{Var}(b)$,
$\operatorname{Var}_{P^X_{i,:}}(y^{(k)})\le B_Y$,
$|y_i^{(k)}|^2\le B_Y$, and the two bounds satisfied by the index
$j_k$. Summing over the $r_Y\le r$ columns and using $d<N$, the
expectation of the softmax curvature term is at most
$C_p\|X\|_\fro^2/d^2$. Hence, there exists a choice of the signs for
which the softmax curvature term is at most $C_p\|X\|_\fro^2/d^2$.

Finally, we prove \cref{eq:tied-teacher-sosp-comparison}. Note that the
function
$K\mapsto N^{-1}\sum_{i\in\mathcal S}\CE(q_i,\mathcal{L}(K)_{i,:})+\lambda\tr K/2$
is convex in the symmetric matrix $K$. It equals
$\cF_{\lambda,\mathcal S}(X)$ at $K=XX^\top$, and
$\cH_{\mathcal S}+\frac\lambda2\|Y\|_\fro^2$ at $K=K^\star$. Thus, we
have
\[
 \begin{aligned}
 &\frac1N\sum_{i\in\mathcal S}\sum_{j=1}^V
        (P^X_{ij}-q_{ij})K^\star_{ij}+\frac\lambda2\|Y\|_\fro^2\\
 &\quad\overset{(a)}{\le}\cH_{\mathcal S}+\frac\lambda2\|Y\|_\fro^2
      -\cF_{\lambda,\mathcal S}(X)
      +\tfrac12\langle\nabla\cF_{\lambda,\mathcal S}(X),X\rangle\\
 &\quad\overset{(b)}{\le}\cH_{\mathcal S}+\frac\lambda2\|Y\|_\fro^2
      -\cF_{\lambda,\mathcal S}(X)+\eta\|X\|_\fro/2,
 \end{aligned}
\]
where $(a)$ follows from the first-order condition of convexity at
$K=XX^\top$ and the last identity in
\cref{eq:tied-differential-identities}, and $(b)$ follows from
$\|\nabla\cF_{\lambda,\mathcal S}(X)\|_\fro\le\eta$ and the
Cauchy--Schwarz inequality. On the other hand, applying the second
identity in \cref{eq:tied-differential-identities} to the direction $Z$
constructed above, and using $ZZ^\top=K^\star$,
$\|Z\|_\fro^2=\|Y\|_\fro^2\le B_YV$, and
$\nabla^2\cF_{\lambda,\mathcal S}(X)\succeq-\gamma I$, we obtain
\[
 -\gamma B_YV\le\nabla^2\cF_{\lambda,\mathcal S}(X)[Z,Z]
 \le\frac{C_p\|X\|_\fro^2}{d^2}
 +2\left\{\frac1N\sum_{i\in\mathcal S}\sum_{j=1}^V
        (P^X_{ij}-q_{ij})K^\star_{ij}+\frac\lambda2\|Y\|_\fro^2\right\}.
\]
Combining the above two inequalities and rearranging the terms proves
\cref{eq:tied-teacher-sosp-comparison}.
\end{proof}

Note that the right-hand side of \cref{eq:tied-teacher-sosp-comparison}
depends on the squared norm $\|X\|_\fro^2$ of the learned embeddings,
which can be as large as $R_w^2$ in the ball. On the other hand, the
objective on the left-hand side contains the weight decay penalty
$\frac\lambda2\|X\|_\fro^2$. Thus, when $\lambda\ge2C_p/d^2$, the term
$C_p\|X\|_\fro^2/(2d^2)$ can be absorbed by the penalty. The following
lemma shows that this first gives $\|X\|_\fro^2=O(V)$, and then a
comparison error of order $V/d^2$.
\begin{lemma}[Approximate SOSPs have squared norm $O(V)$ and small objective]
\label{lem:tied-sosp-energy}
Let $\lambda\ge2C_p/d^2$, and let $R_w$
be the radius in \cref{lem:tied-localization-inputs}.
If $\|X\|_\fro\le R_w$,
$\|\nabla\cF_{\lambda,\mathcal S}(X)\|_\fro\le V/(4R_wd^2)$, and
$\nabla^2\cF_{\lambda,\mathcal S}(X)\succeq-I/(4B_Yd^2)$, then
\[
 \|X\|_\fro^2\le\left(2B_Y+\frac1{2C_p}\right)V,\qquad
 \cF_{\lambda,\mathcal S}(X)-\left(\cH_{\mathcal S}+\frac\lambda2\|Y\|_\fro^2\right)\le(C_pB_Y+1/2)\frac V{d^2}.
\]
\end{lemma}

\begin{proof}
We apply \cref{eq:tied-teacher-sosp-comparison} with
$\eta=V/(4R_wd^2)$ and $\gamma=1/(4B_Yd^2)$. Since $\|X\|_\fro\le R_w$,
each of the last two terms in \cref{eq:tied-teacher-sosp-comparison} is
at most $V/(8d^2)$, i.e., their sum is at most $V/(4d^2)$. Thus, using
$\|Y\|_\fro^2\le B_YV$, we have
\[
 \begin{aligned}
 \cF_{\lambda,\mathcal S}(X)-\cH_{\mathcal S}-\lambda VB_Y/2
 &\overset{(a)}{\le}\lambda\|X\|_\fro^2/4+V/(4d^2)\\
 &\overset{(b)}{\le}\tfrac12\bigl(\cF_{\lambda,\mathcal S}(X)-\cH_{\mathcal S}\bigr)
       +V/(4d^2),
 \end{aligned}
\]
where $(a)$ follows from \cref{eq:tied-teacher-sosp-comparison} and
$\lambda\ge2C_p/d^2$, and $(b)$ follows from
$\cF_{\lambda,\mathcal S}(X)-\cH_{\mathcal S}\ge\lambda\|X\|_\fro^2/2$.
Rearranging the terms, we have $\cF_{\lambda,\mathcal S}(X)-\cH_{\mathcal S}
\le\lambda VB_Y+V/(2d^2)$ and
\[
 \|X\|_\fro^2\le2B_YV+\frac{V}{\lambda d^2}
       \le\left(2B_Y+\frac1{2C_p}\right)V.
\]
Substituting this bound on $\|X\|_\fro^2$ back into
\cref{eq:tied-teacher-sosp-comparison} gives
\begin{equation*}
 \cF_{\lambda,\mathcal S}(X)-\left(\cH_{\mathcal S}+\frac\lambda2\|Y\|_\fro^2\right)\le(C_pB_Y+1/2)\frac V{d^2}.
\end{equation*}
Note that the final comparison error does not depend on the radius
$R_w$, and thus has no factor of $(\lambda V)^{-1}$.

\end{proof}

\subsubsection{Step 3: Refresh the Gain throughout the Final Window}
\label{sec:tied-step3}

In this step, we extend the comparison in Step 2, which holds at
approximate SOSPs, to every iterate in the final window. The argument
is the same as Step 3 in Appendix~\ref{sec:untied-step3}. We partition
the iterations into $d$ blocks, and apply the one-block estimate
(\cref{thm:tied-fixed-noise-consequence} in
Appendix~\ref{app:optimization-tools}) to the smooth extension in
Step 1. With high probability, each block contains a visit to an
approximate SOSP, at which \cref{lem:tied-sosp-energy} applies, and the
objective increases by at most $V/(4d^2)$ within the block. An induction
over the blocks then gives the following lemma.

\begin{lemma}[Final-window objective comparison]
\label{lem:tied-final-window-comparison}
Let \cref{def:appendix-tied-class} hold, with
$\log V-\cH_{\mathcal S}\ge\delta_{\rm tr}>0$,
$2r_Y\le d<N<V$, and $d\ge C\log(eV)$. Suppose that
\begin{equation*}
 C_p=40B_Yr,\qquad
 \lambda\ge2C_p/d^2,\quad 0<\lambda V\le1.
\end{equation*}
Then, there exist deterministic $h,\sigma>0$ and an integer
$T_{\rm blk}$, which are specified in the proof, such that $h^{-1}$ and
$\sigma^{-1}$ are bounded by polynomials in
$(V,N,d,(\lambda V)^{-1})$, and
\begin{equation}
 T=dT_{\rm blk}\le p_{\rm time}(V,N,d,(\lambda V)^{-1}),
 \label{eq:tied-polynomial-time-bound}
\end{equation}
where $p_{\rm time}$ is a fixed polynomial whose coefficients depend
only on the parameters of the class. Moreover, with probability at
least $1-e^{-cd}$ over the initialization and the perturbations, every
integer
$T_{\rm blk}\le t\le T$ satisfies
\begin{equation*}
 \begin{gathered}
 \cF_{\lambda,\mathcal S}(X_t)-\left(\cH_{\mathcal S}+\frac\lambda2\|Y\|_\fro^2\right)\le\frac{(C_pB_Y+1)V}{d^2},
 \\
 g_{\mathcal S}(X_t)\ge
 1-\frac{B_Y(\lambda V)/2+\frac{(C_pB_Y+1)V}{d^2}}{\delta_{\rm tr}}.
 \end{gathered}
\end{equation*}
In particular, both bounds hold at the final iterate $t=T$.
\end{lemma}

\begin{proof}
The proof follows the same structure as the proof of
\cref{thm:iclr-untied} in Appendix~\ref{sec:untied-step3}.

\textbf{Parameter choices.}
Let $A$, $R_w$, $\widetilde{\cF}_{\lambda,\mathcal S}$,
$\ell_{\rm ext}$, and $\rho_{\rm ext}$ be as in
\cref{lem:tied-localization-inputs}. We set
\begin{equation*}
 \begin{gathered}
 \epsilon=\min\left\{\frac{V}{4R_wd^2},
             \frac1{16B_Y^2d^4\rho_{\rm ext}},1\right\},\\
 \delta_c=\min\{V/(4d^2),1\},\qquad
 \zeta=\frac{e^{-d}}{8d},
 \end{gathered}
\end{equation*}
and let $h$, $\sigma$, and $T_{\rm blk}$ be given by
\cref{thm:tied-fixed-noise-consequence} applied to
$f=\widetilde{\cF}_{\lambda,\mathcal S}$ with $D_\Theta=Vd$,
$\ell=\ell_{\rm ext}$, $\rho=\rho_{\rm ext}$, and the above
$A,\epsilon,\delta_c,\zeta$. The algorithm runs $T=dT_{\rm blk}$
iterations with these fixed $h$ and $\sigma$. Since
$\sqrt{\rho_{\rm ext}\epsilon}\le1/(4B_Yd^2)$, the choice of $\epsilon$
guarantees that every $\epsilon$-approximate SOSP $X$ of
$\cF_{\lambda,\mathcal S}$ with $\|X\|_\fro\le R_w$ satisfies the
assumptions of \cref{lem:tied-sosp-energy}. Thus, at every such point,
\begin{equation}
 \cF_{\lambda,\mathcal S}(X)-\left(\cH_{\mathcal S}+\frac\lambda2\|Y\|_\fro^2\right)
 \le\left(C_pB_Y+\frac12\right)\frac V{d^2}.
 \label{eq:tied-visit-comparison}
\end{equation}

\textbf{Coupling and one-block estimate.}
We condition on the event in \cref{eq:initialization-event}. As in
Appendix~\ref{sec:untied-step3}, let $\{\widetilde X_t\}$ be the
auxiliary trajectory that runs the same recursion on
$\widetilde{\cF}_{\lambda,\mathcal S}$ with the same initialization
$\widetilde X_0=X_0$ and the same perturbations $\{\xi_t\}$. Since
$\widetilde{\cF}_{\lambda,\mathcal S}$ coincides with
$\cF_{\lambda,\mathcal S}$ in the ball $\|X\|_\fro\le R_w$, we have
$\widetilde X_t=X_t$ as long as the auxiliary trajectory has not left
the ball. We partition the iterations into $d$ blocks of length
$T_{\rm blk}$, where consecutive blocks share their boundary iterate.
By \cref{thm:tied-fixed-noise-consequence}, conditional on the history
up to the first iterate of a block, if the objective gap
$\widetilde{\cF}_{\lambda,\mathcal S}(\widetilde X_t)-\min_X\cF_{\lambda,\mathcal S}(X)$
at this iterate is at most $A$, then with conditional probability at
least $1-2\zeta$, the block is \emph{successful}, i.e., it contains an
$\epsilon$-approximate SOSP of $\widetilde{\cF}_{\lambda,\mathcal S}$,
and $\widetilde{\cF}_{\lambda,\mathcal S}$ increases by at most
$\delta_c\le V/(4d^2)$ between any two iterates $u<v$ in the block.

\textbf{Base case: the first block.}
By \cref{eq:initialization-event} and $\lambda V\le1$, we have
$\cF_{\lambda,\mathcal S}(X_0)\le\cH_{\mathcal S}+2B_Y+9$. Since
$\min_X\cF_{\lambda,\mathcal S}(X)\ge\cH_{\mathcal S}$ and
$2B_Y+9\le A$, the one-block estimate applies to the first block.
Suppose that the first block is successful. Then, every auxiliary
iterate in the block satisfies
$\widetilde{\cF}_{\lambda,\mathcal S}(\widetilde X_t)
\le\cF_{\lambda,\mathcal S}(X_0)+\delta_c\le\cH_{\mathcal S}+2B_Y+10$,
which is smaller than $\cH_{\mathcal S}+A+4$. By
\cref{lem:tied-localization-inputs}, every auxiliary iterate in the
block is inside the ball, and thus $\widetilde X_t=X_t$ for all $t$ in
the block. In particular, the approximate SOSP visited in the block is
an approximate SOSP of $\cF_{\lambda,\mathcal S}$ in the ball, at which
\cref{eq:tied-visit-comparison} holds. Since the objective increases by
at most $V/(4d^2)$ from this visit to the last iterate of the block, we
obtain
\begin{equation}
 \cF_{\lambda,\mathcal S}(X_{jT_{\rm blk}})
 -\left(\cH_{\mathcal S}+\frac\lambda2\|Y\|_\fro^2\right)
 \le\left(C_pB_Y+\frac34\right)\frac V{d^2}
 \label{eq:tied-refresh-margin}
\end{equation}
for $j=1$.

\textbf{Induction step: the later blocks.}
Suppose that the first $j$ blocks are successful for some $1\le j<d$,
that the two trajectories coincide up to iteration $jT_{\rm blk}$, and
that \cref{eq:tied-refresh-margin} holds. Since $\|Y\|_\fro^2\le B_YV$,
$\lambda V\le1$, and $V/d^2\le\lambda V/(2C_p)\le1/(2C_p)$,
\cref{eq:tied-refresh-margin} implies
\[
 \cF_{\lambda,\mathcal S}(X_{jT_{\rm blk}})-\cH_{\mathcal S}
 \le\frac{B_Y(\lambda V)}2+\left(C_pB_Y+\frac34\right)\frac V{d^2}
 \le B_Y+\frac3{8C_p}\le A.
\]
Thus, the one-block estimate applies to the $(j+1)$-th block. Suppose
that this block is successful. Then, every iteration $t$ in the block
satisfies
\[
 \widetilde{\cF}_{\lambda,\mathcal S}(\widetilde X_t)
 -\left(\cH_{\mathcal S}+\frac\lambda2\|Y\|_\fro^2\right)
 \le\left(C_pB_Y+\frac34\right)\frac V{d^2}+\delta_c
 \le\left(C_pB_Y+1\right)\frac V{d^2}.
\]
By the same calculation as above, this implies
$\widetilde{\cF}_{\lambda,\mathcal S}(\widetilde X_t)
\le\cH_{\mathcal S}+B_Y+1/(2C_p)<\cH_{\mathcal S}+A+4$.
Hence, the auxiliary trajectory stays inside the ball,
$\widetilde X_t=X_t$ for all $t$ in the block, and the first bound in
the lemma holds for all $t$ in the block. Moreover, applying
\cref{eq:tied-visit-comparison} at the approximate SOSP visited in this
block, together with the bound on the increase of the objective, we
conclude that \cref{eq:tied-refresh-margin} holds with $j$ replaced by
$j+1$. By induction, if all $d$ blocks are successful, the first bound
in the lemma holds for every integer $t\in[T_{\rm blk},T]$.

\textbf{Gain, failure probability, and running time.}
For the gain, since $\|Y\|_\fro^2\le B_YV$ and the weight decay penalty
of $X_t$ is nonnegative, the first bound in the lemma implies
\[
 \cR_{\mathcal S}(X_t)-\cH_{\mathcal S}
 \le\frac{B_Y(\lambda V)}2+\frac{(C_pB_Y+1)V}{d^2}.
\]
Dividing both sides by $\log V-\cH_{\mathcal S}\ge\delta_{\rm tr}$ gives
the second bound in the lemma. For the probability, conditional on the
initialization event and on the success of all previous blocks, each
block fails with probability at most $2\zeta$. Thus, by the union
bound, the total failure probability is at most the failure probability
of the initialization plus $2d\zeta=e^{-d}/4$, which is at most
$e^{-cd}$ after adjusting the constant $c$. For the running time, note
that $\lambda V\ge2C_pV/d^2>2C_p/V$, $d^2/V\le V$, and
$\log(1/\zeta)=d+\log(8d)$. Thus, $R_w$, $\ell_{\rm ext}$,
$\rho_{\rm ext}$, $\epsilon^{-1}$, and $\delta_c^{-1}$ are polynomial in
$(V,N,d,(\lambda V)^{-1})$, and so are $h^{-1}$, $\sigma^{-1}$, and
$T_{\rm blk}$ by \cref{eq:predeclared-schedule}. This proves
\cref{eq:tied-polynomial-time-bound}.
\end{proof}

We are now ready to prove \cref{thm:iclr-tied-training}. We choose
$\lambda=2C_p/d^2$, under which both the weight decay penalty of the
ground truth, i.e., $\frac\lambda2\|Y\|_\fro^2\le B_Y(\lambda V)/2$, and
the comparison error in \cref{lem:tied-final-window-comparison} are of
order $V/d^2$. For convenience, we restate the theorem below.

\begin{theorem}[Restatement of \Cref{thm:iclr-tied-training}: Tied training]
Let $Q\in\mathcal Q^{\rm tied}_{r,B,\Delta_0}$ and suppose
$\log V-\cH_{\mathcal S}\ge\delta_{\rm tr}>0$. For sufficiently
large $V$ and $\max\{2r,C\sqrt V\}\le d<N$, setting the weight decay
$\lambda=C_\lambda/d^2$, there exist a step size $h$, a perturbation scale $\sigma$, and a number of iterations $T$ divisible
by $d$ such that, with probability at least $1-e^{-cd}$,
\begin{equation*}
 g_{\mathcal S}(X_t)\ge1-\frac{CV}{d^2\delta_{\rm tr}}
 \quad\text{for every integer }t\in[T/d,T].
\end{equation*}
The quantities $h^{-1},\sigma^{-1}$, and $T$ are bounded by polynomials in $V$.
\end{theorem}

\begin{proof}
Let $C_\lambda=2C_p$, i.e., $\lambda=2C_p/d^2$. We first verify the
assumptions of \cref{lem:tied-final-window-comparison}. When the
constant $C$ in $d\ge C\sqrt V$ is sufficiently large, we have
$\lambda V=2C_pV/d^2\le1$, and $d\ge C'\log(eV)$ for all sufficiently
large $V$. In addition, $d\ge2r$ implies $d\ge2r_Y$. Thus,
\cref{lem:tied-final-window-comparison} applies with this $\lambda$.
With probability at least $1-e^{-cd}$ over the initialization and the
perturbations, every integer $t\in[T/d,T]$ satisfies
\[
 \bigl(\cR_{\mathcal S}(X_t)-\cH_{\mathcal S}\bigr)
 \le B_Y(\lambda V)/2+(C_pB_Y+1)\frac V{d^2}
 =(2C_pB_Y+1)\frac V{d^2}.
\]
Dividing both sides by $\log V-\cH_{\mathcal S}\ge\delta_{\rm tr}$
proves the gain bound. Finally, by
\cref{lem:tied-final-window-comparison}, $h$ and $\sigma$ are fixed,
$T=dT_{\rm blk}$ is divisible by $d$, and $h^{-1}$, $\sigma^{-1}$, and
$T$ are bounded by polynomials in $V$, since $N<V$, $d<V$, and
$(\lambda V)^{-1}=d^2/(2C_pV)<V$.
\end{proof}

\subsection{Proof of Theorem~\ref{thm:iclr-transfer}}
\label{app:transfer-proof}

In this section, we present the full proof of \cref{thm:iclr-transfer}.
The proof consists of three main steps.

\textbf{Step 1 (Appendix~\ref{sec:transfer-step1}).}
We apply \cref{lem:tied-final-window-comparison} with the weight decay
in \cref{thm:iclr-transfer}, and show that, with high probability, the
objective at every iterate in the final window exceeds the objective at
the ground truth by at most $O((\lambda V)^2)$.

\textbf{Step 2 (Appendix~\ref{sec:transfer-step2}).}
We convert this comparison of the objectives into a bound of
$O((\lambda V)^2)$ on the training excess loss, and an upper bound on
the trace of the learned Gram matrix. The key tool is an identity that
relates the trace of the Gram matrix to the logits on the observed rows
under the coverage condition
(\cref{lem:tied-anchor-trace-certificate}), together with a bound on
the error of the logits in terms of the KL divergence
(\cref{lem:tied-kl-lone}).

\textbf{Step 3 (Appendix~\ref{sec:transfer-step3}).}
We show that the error of the entire Gram matrix is controlled by the
error of the logits on the observed rows and the excess of the trace
(\cref{lem:tied-psd-completion}). Combining this result with Step 2 and
the bound on the learned embeddings, we bound the loss on the unseen
states (\cref{lem:iclr-transfer-mechanism}), and then prove
\cref{thm:iclr-transfer}.

We note that Steps 2 and 3 are deterministic, i.e., they hold for every
$X$ that satisfies the comparison in Step 1.

\paragraph{Notation and assumptions.}
We use the notation in Appendix~\ref{app:tied-proofs}. In addition, by
\cref{def:appendix-transfer-coverage}, the representative $Y$ satisfies
\[
 \begin{gathered}
 Y^\top\1=0,\qquad
 \sigma_{r_Y}(Y_{\mathcal S}-\1_N\bar y_{\mathcal S}^\top)
       \ge\kappa\sqrt N,\\
 \bar y_{\mathcal S}=\frac1N\sum_{i\in\mathcal S}y_i.
 \end{gathered}
\]
We collect some direct consequences of these conditions. First, since
$Y^\top\1=0$, the matrix $K^\star=YY^\top$ is row-centered, and thus
$\mathcal{L}(\log Q)=K^\star$. Second, since the diagonal entries of
$K^\star$ are $\|y_i\|_2^2$ and $\|\mathcal{L}(\log Q)\|_\infty\le B$,
we have $\max_i\|y_i\|_2^2\le B$, and hence $|K^\star_{ij}|\le B$.
Third, the coverage condition implies
$r_Y\kappa^2N\le\|Y_{\mathcal S}-\1_N\bar y_{\mathcal S}^\top\|_\fro^2\le NB$.
Throughout this section, the constants $c$ and $C$ may depend on the
parameters of the class, the coverage parameter $\kappa$, and the
constants in $N=\Theta(V)$, but not on $V$, $N$, or $d$. The dependence
on the bound $R_0$ on the learned embeddings is stated explicitly.

\subsubsection{Step 1: Compare the Objective with the Ground Truth in the Final Window}
\label{sec:transfer-step1}

In this step, we specify the weight decay and apply
\cref{lem:tied-final-window-comparison}. Recall that in
\cref{thm:iclr-transfer}, the weight decay and the embedding dimension
satisfy
\begin{equation}
 \lambda=C_\lambda\sqrt{\frac{V}{Nd^3}},\qquad
 d\ge C_{\rm wid}\frac{V}{N^{1/3}},\qquad 2r_Y\le d<N<V.
 \label{eq:tied-polynomial-decay}
\end{equation}
Note that this weight decay is different from $\lambda=2C_p/d^2$ in the
proof of \cref{thm:iclr-tied-training}. We choose the constant
$C_\lambda$ sufficiently large, and then the constant $C_{\rm wid}$
sufficiently large, both depending only on the parameters of the class,
such that the following three conditions hold:
\[
 \text{(i) }\lambda\ge\frac{2C_p}{d^2},\qquad
 \text{(ii) }\frac{(C_pB_Y+1)V}{d^2}\le C(\lambda V)^2,\qquad
 \text{(iii) }0<\lambda V\le c_{\rm decay}\le1,
\]
where $c_{\rm decay}$ is a sufficiently small constant that is required
by \cref{lem:tied-kl-lone} in Step 2. These conditions can be satisfied
because
\[
 \begin{gathered}
 \lambda d^2=C_\lambda\sqrt{\frac{Vd}{N}}\ge C_\lambda,\qquad
 \lambda V\le C_\lambda C_{\rm wid}^{-3/2},\\
 \frac{(C_pB_Y+1)V/d^2}{(\lambda V)^2}
 =\frac{(C_pB_Y+1)Nd}{C_\lambda^2V^2}\le\frac{C_pB_Y+1}{C_\lambda^2},
 \end{gathered}
\]
where we use $1\le d<N<V$ and $d\ge C_{\rm wid}V/N^{1/3}$.

Conditions (i) and (iii) are the assumptions on $\lambda$ in
\cref{lem:tied-final-window-comparison}, and its remaining assumptions
are verified in the proof of the theorem below. Thus,
\cref{lem:tied-final-window-comparison} provides deterministic $h$,
$\sigma$, and $T=dT_{\rm blk}$ such that, with probability at least
$1-e^{-cd}$, every integer $t\in[T/d,T]$ satisfies
\[
 \cF_{\lambda,\mathcal S}(X_t)-\left(\cH_{\mathcal S}+\frac\lambda2\|Y\|_\fro^2\right)
 \le\frac{(C_pB_Y+1)V}{d^2}\le C(\lambda V)^2,
\]
where the second inequality is condition (ii). The remaining steps are
deterministic, and apply to every such iterate. In particular, the
bound on the learned embeddings in \cref{thm:iclr-transfer} is not part
of this probability event.

\subsubsection{Step 2: Bound the Training Loss and the Trace of the Gram Matrix}
\label{sec:transfer-step2}

In Step 1, we obtained an upper bound on
$\cF_{\lambda,\mathcal S}(X)-\bigl(\cH_{\mathcal S}+\frac\lambda2\|Y\|_\fro^2\bigr)
=\{\cR_{\mathcal S}(X)-\cH_{\mathcal S}\}+\frac\lambda2\{\tr K-\tr K^\star\}$.
This does not immediately bound the excess loss
$\cR_{\mathcal S}(X)-\cH_{\mathcal S}$, because the second term can be
negative, i.e., an increase in the loss can be compensated by a
decrease in the weight decay penalty. Therefore, we need to show that
the trace of the learned Gram matrix $K$ cannot be much smaller than
that of $K^\star$, unless the logits on the observed rows have a large
error. This is where the coverage condition is used.

\paragraph{Linear relations among the ground-truth embeddings.}
By the coverage condition, the centered embeddings of the observed
states span $\R^{r_Y}$. Hence, the embedding $y_i$ of every state
$i\in[V]$, observed or unseen, is a linear combination of the embeddings
of the observed states, i.e., $y_i=Y_{\mathcal S}^\top a_i$ for some
coefficient vector $a_i\in\R^N$, whose coordinates are indexed by
$s\in\mathcal S$. Specifically, we use the following coefficient
vectors:
\begin{equation}
 \begin{aligned}
 a_i={}&\left[
 (I_N-\1_N\1_N^\top/N)(\log Q_{\mathcal S,\mathcal S})
       (I_N-\1_N\1_N^\top/N)
        \right]^\dagger\\
 &\quad\cdot(I_N-\1_N\1_N^\top/N)
       \mathcal{L}(\log Q_{\mathcal S,:})_{:,i}
       \in\R^N.
 \end{aligned}
 \label{eq:tied-observable-anchors}
\end{equation}
Here, the logarithm is entrywise, and $\dagger$ denotes the
Moore--Penrose inverse. \cref{lem:tied-anchor-trace-certificate} below
shows that these vectors satisfy $y_i=Y_{\mathcal S}^\top a_i$, and
provides a closed form in terms of $Y$. We note that
\cref{eq:tied-observable-anchors} only involves the rows of $Q$ indexed
by $\mathcal S$. This is possible because, although the state $i$ may
be unseen as a current state, it is still a possible next state, and
thus the $i$-th column of $Q_{\mathcal S,:}$ is available.

For the learned embeddings, the same linear relations do not hold in
general, and we measure their violation by the residuals
$x_i-X_{\mathcal S}^\top a_i$. The key identity in
\cref{lem:tied-anchor-trace-certificate} states that the sum of the
squared residuals equals the excess of the trace, $\tr(K-K^\star)$,
minus a linear function of the error of the logits on the observed
rows. Since the squared residuals are nonnegative, this identity lower
bounds $\tr(K-K^\star)$ by the error of the logits on the observed
rows, which is what we need. We refer to this identity as the trace
certificate.

In the following, we use
\[
 Q_\perp=I_V-Y(Y^\top Y)^{-1}Y^\top
\]
to denote the orthogonal projection onto the orthogonal complement of
the column space of $Y$, and write $\sym(H)=(H+H^\top)/2$. Note that the
results in this step hold for every $X\in\R^{V\times d}$, which is not
required to be centered or to be a stationary point.
\begin{lemma}[Linear relations among the ground-truth embeddings and the trace certificate]
\label{lem:tied-anchor-trace-certificate}
The coefficients in \cref{eq:tied-observable-anchors} satisfy
\begin{equation*}
 \begin{gathered}
 a_i=(Y_{\mathcal S}-\1_N\bar y_{\mathcal S}^\top)
       (Y_{\mathcal S}^\top Y_{\mathcal S}
                  -N\bar y_{\mathcal S}\bar y_{\mathcal S}^\top)^{-1}y_i,\\
 a_i^\top Y_{\mathcal S}=y_i^\top,\qquad
 a_i^\top K^\star_{\mathcal S,:}=K^\star_{i,:},\\
 \|a_i\|_2\le\frac{\sqrt B}{\kappa\sqrt N},\qquad
 \|a_i\|_\infty\le\frac{2B}{\kappa^2N}.
 \end{gathered}
\end{equation*}
In particular, every entry of $K^\star$ is determined by
$Q_{\mathcal S,:}$ through
\[
 K^\star_{ij}
 =a_i^\top \mathcal{L}(\log Q_{\mathcal S,:})_{:,j}.
\]
Moreover, there exists a matrix $Y_{\rm dual}\in\R^{N\times V}$ with
$\|Y_{\rm dual}\|_\infty\le C/V$ and $\|Y_{\rm dual}\|_\fro\le C$
such that, for every $X\in\R^{V\times d}$ with $K=XX^\top$,
\begin{equation}
 \begin{gathered}
 M_{\rm dual}=I_V-\sym((I_V)_{:,\mathcal S}\mathcal{L}(Y_{\rm dual})),\\
 M_{\rm dual}\succeq Q_\perp,\qquad M_{\rm dual}Y=0,\\
 \begin{aligned}
 \sum_i\|x_i-X_{\mathcal S}^\top a_i\|_2^2
 &=\tr(K-K^\star)\\
 &\quad-\langle Y_{\rm dual},\mathcal{L}(K-K^\star)_{\mathcal S,:}\rangle.
 \end{aligned}
 \end{gathered}
 \label{eq:tied-trace-certificate}
\end{equation}
\end{lemma}

\begin{proof}
We first prove the properties of $a_i$. Since $Y^\top\1=0$, we have
$\mathcal{L}(\log Q)=YY^\top$, and thus
\[
 \mathcal{L}(\log Q_{\mathcal S,:})=Y_{\mathcal S}Y^\top.
\]
Multiplying it by $I_N-\1_N\1_N^\top/N$ from the left gives
$(Y_{\mathcal S}-\1_N\bar y_{\mathcal S}^\top)Y^\top$.
Similarly, since the entries of $\log Q_{\mathcal S,\mathcal S}$ differ
from those of $Y_{\mathcal S}Y_{\mathcal S}^\top$ only by a constant in
each row, centering $\log Q_{\mathcal S,\mathcal S}$ on both sides gives
\[
 (Y_{\mathcal S}-\1_N\bar y_{\mathcal S}^\top)
 (Y_{\mathcal S}-\1_N\bar y_{\mathcal S}^\top)^\top.
\]
Since $Y_{\mathcal S}-\1_N\bar y_{\mathcal S}^\top$ has full column rank
by the coverage condition, the identity
$(AA^\top)^\dagger A=A(A^\top A)^{-1}$ for a matrix $A$ with full column
rank, applied to \cref{eq:tied-observable-anchors}, proves the closed
form of $a_i$.
Since
\[
 (Y_{\mathcal S}-\1_N\bar y_{\mathcal S}^\top)^\top Y_{\mathcal S}
 =Y_{\mathcal S}^\top Y_{\mathcal S}
                  -N\bar y_{\mathcal S}\bar y_{\mathcal S}^\top,
\]
the closed form gives $a_i^\top Y_{\mathcal S}=y_i^\top$. Multiplying
this identity by $Y^\top$ from the right gives
$a_i^\top K^\star_{\mathcal S,:}=K^\star_{i,:}$ and the formula for
$K^\star_{ij}$. For the norms, the coverage condition gives
$\|a_i\|_2\le\|y_i\|_2/(\kappa\sqrt N)\le\sqrt B/(\kappa\sqrt N)$, and
the bound $\|y_s-\bar y_{\mathcal S}\|_2\le2\sqrt B$ gives the bound on
$\|a_i\|_\infty$. In addition, the closed form implies
$a_i^\top\1_N=0$ and $\sum_{i=1}^Va_i=0$, which will be used below.

Next, we construct $Y_{\rm dual}$. For $s\in\mathcal S$ and $j\in[V]$,
define
\[
 (Y_{\rm dual})_{s,j}
 =\begin{cases}
 2(a_j)_s-\displaystyle\sum_{i=1}^V(a_i)_s(a_i)_j,&j\in\mathcal S,\\
 2(a_j)_s,&j\notin\mathcal S.
 \end{cases}
\]
By $a_i^\top\1_N=0$ and $\sum_ia_i=0$, every row of $Y_{\rm dual}$ sums
to zero, and thus $\mathcal{L}(Y_{\rm dual})=Y_{\rm dual}$.
Expanding the square, for every $v\in\R^V$ we have
\[
 v^\top M_{\rm dual}v
 =\sum_{i=1}^V\left(v_i-\sum_{s\in\mathcal S}(a_i)_s v_s\right)^2.
\]
We now verify the properties of $M_{\rm dual}$. First, when $v$ is a
column of $Y$, the identity $a_i^\top Y_{\mathcal S}=y_i^\top$ shows
that every summand is zero. Since the above identity also shows
$M_{\rm dual}\succeq0$, this implies $M_{\rm dual}Y=0$. Second, by the
closed form of $a_i$, the vector with entries
$\sum_{s\in\mathcal S}(a_i)_s v_s$ lies in the column space of $Y$.
Thus, for $v$ orthogonal to the column space of $Y$, the cross term in
the above sum of squares vanishes, which gives
$v^\top M_{\rm dual}v\ge\|v\|_2^2$. Together with $M_{\rm dual}Y=0$,
this proves $M_{\rm dual}\succeq Q_\perp$. Third, applying the above
identity to each column of $X$ and summing over the columns gives
$\langle M_{\rm dual},K\rangle=\sum_i\|x_i-X_{\mathcal S}^\top a_i\|_2^2$.
On the other hand, by the definition of $M_{\rm dual}$, the symmetry of
$K$, and $\langle M_{\rm dual},K^\star\rangle=0$, we have
$\langle M_{\rm dual},K\rangle=\tr(K-K^\star)
-\langle Y_{\rm dual},(K-K^\star)_{\mathcal S,:}\rangle$.
Since the rows of $Y_{\rm dual}$ sum to zero, $(K-K^\star)_{\mathcal S,:}$
can be replaced by $\mathcal{L}(K-K^\star)_{\mathcal S,:}$ in the last
inner product. This proves the last identity in
\cref{eq:tied-trace-certificate}.
Finally, the bound on $\|a_i\|_\infty$ gives
\[
 \|Y_{\rm dual}\|_\infty
 \le\frac{4B}{\kappa^2N}+\frac{4B^2V}{\kappa^4N^2}\le C/V.
\]
Multiplying this bound by $\sqrt{NV}$ gives $\|Y_{\rm dual}\|_\fro\le C$.
\end{proof}

The trace certificate in \cref{eq:tied-trace-certificate} is stated in
terms of the error of the logits on the observed rows, while the
comparison in Step 1 is stated in terms of the loss, i.e., the KL
divergence. The following lemma connects the two. Its first part bounds
the $\ell^1$ error of the centered logits by the KL divergence, which
holds without any bound on the learned logits. Its second part combines
this bound with the trace certificate and the comparison in Step 1.
Among the four conclusions of the second part, the bounds on the
training excess loss and on the trace are the ones used in Step 3.
\begin{lemma}[From the objective comparison to bounds on the training loss and the trace]
\label{lem:tied-kl-lone}
For every $X\in\R^{V\times d}$, without any bound on its logits,
\begin{equation}
 \begin{aligned}
 \frac{\|\mathcal{L}(K-K^\star)_{\mathcal S,:}\|_1}{NV}
 &\le2e^{2B}\bigl(\cR_{\mathcal S}(X)-\cH_{\mathcal S}\bigr)\\
 &\quad+2e^{2B}\sqrt{2\bigl(\cR_{\mathcal S}(X)-\cH_{\mathcal S}\bigr)}.
 \end{aligned}
 \label{eq:tied-kl-lone}
\end{equation}
Moreover, suppose that
$\cF_{\lambda,\mathcal S}(X)-\left(\cH_{\mathcal S}+\frac\lambda2\|Y\|_\fro^2\right)\le\nu$
for some $\nu\ge0$, and that $0<\lambda V\le c_{\rm decay}$ for a
sufficiently small constant $c_{\rm decay}$ depending only on the
parameters of the class. Then,
\begin{align*}
 \cR_{\mathcal S}(X)-\cH_{\mathcal S}&\le C(\nu+(\lambda V)^2),\\
 (\tr K-\tr K^\star)_+&\le2\nu/\lambda,\\
 \|(I_V-\1\1^\top/V) X\|_\fro^2/V&\le B+2\nu/(\lambda V),\\
 \sum_i\|x_i-X_{\mathcal S}^\top a_i\|^2&\le CV((\lambda V)+\sqrt\nu+\nu/(\lambda V)).
\end{align*}
\end{lemma}

\begin{proof}
We first prove \cref{eq:tied-kl-lone}. Fix an observed row
$i\in\mathcal S$. Using the inequality $\log z\le z-1$, we have
\[
 \sum_jQ_{ij}\left(\log\frac{P^X_{ij}}{Q_{ij}}\right)_+
 \le\sum_j(P^X_{ij}-Q_{ij})_+
 \le\sqrt{\frac{\KL(q_i\|P^X_{i,:})}{2}},
\]
where the last inequality is Pinsker's inequality. Since
$\sum_jQ_{ij}\log(P^X_{ij}/Q_{ij})=-\KL(q_i\|P^X_{i,:})$ and
$|a|=2a_+-a$ for any real number $a$, we obtain
\[
 \sum_jQ_{ij}\left|\log\frac{P^X_{ij}}{Q_{ij}}\right|
 \le\KL(q_i\|P^X_{i,:})+\sqrt{2\KL(q_i\|P^X_{i,:})}.
\]
Recall that the centered ground-truth logits lie in $[-B,B]$, which
implies $Q_{ij}\ge e^{-2B}/V$. In addition, since $\log P^X_{i,:}$ and
$\log Q_{i,:}$ differ from $K_{i,:}$ and $K^\star_{i,:}$ by constants,
respectively, the error of the centered logits is the centered
log-ratio of the probabilities:
\[
 \begin{aligned}
 &\mathcal{L}(K-K^\star)_{i,:}\\
 &\qquad=\mathcal{L}(\log P^X_{i,:}-\log Q_{i,:}).
 \end{aligned}
\]
Since centering increases the $\ell^1$ norm by at most a factor of two, we have
\[
 \begin{aligned}
 \|\mathcal{L}(K-K^\star)_{i,:}\|_1
 &\le2\sum_j\left|\log\frac{P^X_{ij}}{Q_{ij}}\right|\\
 &\le2Ve^{2B}\left\{\KL(q_i\|P^X_{i,:})
             +\sqrt{2\KL(q_i\|P^X_{i,:})}\right\}.
 \end{aligned}
\]
Averaging over $i\in\mathcal S$ and applying Jensen's inequality to the
square root proves \cref{eq:tied-kl-lone}.

Next, we prove the second part. Since the left-hand side of the last
identity in \cref{eq:tied-trace-certificate} is nonnegative, and
$\|Y_{\rm dual}\|_\infty\le C/V$ with $N=\Theta(V)$, we have
\[
 \begin{aligned}
 \tr(K-K^\star)
 &\ge\langle Y_{\rm dual},\mathcal{L}(K-K^\star)_{\mathcal S,:}\rangle\\
 &\ge-\frac C N\|\mathcal{L}(K-K^\star)_{\mathcal S,:}\|_1.
 \end{aligned}
\]
Note that $\|X\|_\fro^2=\tr K$ and $\|Y\|_\fro^2=\tr K^\star$, so the
assumption on the objective is equivalent to
$\cR_{\mathcal S}(X)-\cH_{\mathcal S}
+\lambda\tr(K-K^\star)/2\le\nu$. Combining it with the above bound and
\cref{eq:tied-kl-lone}, we have
\[
 \begin{aligned}
 \cR_{\mathcal S}(X)-\cH_{\mathcal S}
 &\le\nu+C\lambda V\left\{\cR_{\mathcal S}(X)-\cH_{\mathcal S}
          +\sqrt{\cR_{\mathcal S}(X)-\cH_{\mathcal S}}\right\}.
 \end{aligned}
\]
We choose $c_{\rm decay}$ such that the coefficient $C\lambda V$ is at
most $1/2$, and apply Young's inequality to the term with the square
root. This gives
$\cR_{\mathcal S}(X)-\cH_{\mathcal S}\le C(\nu+(\lambda V)^2)$, which is
the first bound. For the second bound, since
$\cR_{\mathcal S}(X)-\cH_{\mathcal S}\ge0$, the assumption on the
objective directly gives $\lambda\tr(K-K^\star)/2\le\nu$. The third
bound follows from $\|(I_V-\1\1^\top/V) X\|_\fro^2\le\tr K$, the second
bound, and $\tr K^\star\le BV$.
Finally, for the fourth bound, \cref{eq:tied-trace-certificate} gives
\[
 \begin{aligned}
 \sum_i\|x_i-X_{\mathcal S}^\top a_i\|^2
 &\le2\nu/\lambda+\frac C N\|\mathcal{L}(K-K^\star)_{\mathcal S,:}\|_1\\
 &\le CV\{\nu/(\lambda V)+\nu+(\lambda V)^2+\sqrt\nu+\lambda V\}.
 \end{aligned}
\]
Since $\lambda V\le1$, the terms $\nu$ and $(\lambda V)^2$ are dominated
by $\nu/(\lambda V)$ and $\lambda V$, respectively, which completes the
proof.
\end{proof}

In summary, applying \cref{lem:tied-kl-lone} with
$\nu\le C(\lambda V)^2$ from Step 1, every iterate $X=X_t$ in the final
window satisfies
\[
 \cR_{\mathcal S}(X)-\cH_{\mathcal S}\le C(\lambda V)^2,\qquad
 (\tr K-\tr K^\star)_+\le C(\lambda V) V.
\]
These are the two bounds used in Step 3. In addition,
\cref{lem:tied-kl-lone} also gives
$\frac{\|\mathcal{L}(K-K^\star)_{\mathcal S,:}\|_1}{NV}\le C(\lambda V)$,
$\|(I_V-\1\1^\top/V) X\|_\fro^2/V\le C$, and
$\sum_i\|x_i-X_{\mathcal S}^\top a_i\|^2\le C(\lambda V) V$, which are
not needed in the proof of \cref{thm:iclr-transfer}.

\subsubsection{Step 3: Transfer the Bounds to Unseen States}
\label{sec:transfer-step3}

In this step, we show that the bounds on the observed rows in Step 2
control the entire Gram matrix, and thus the loss on the unseen states.
The difficulty is that the rows of $K-K^\star$ indexed by $\mathcal S$
do not determine the whole matrix in general. To handle this, we
decompose the error as
\[
 K-K^\star=H+D_0,\qquad D_0=Q_\perp KQ_\perp,
\]
where $Q_\perp$ is the orthogonal projection onto the orthogonal
complement of the column space of $Y$. The first part $H$ satisfies
$Q_\perp HQ_\perp=0$, i.e., it has no component that lies outside the
column space of $Y$ on both sides. We show that such a matrix is
controlled by its centered rows indexed by $\mathcal S$. The second
part $D_0$ is positive semidefinite because $K$ is, so its Frobenius
norm is at most its trace, and its trace is controlled by the trace
certificate. The following lemma makes this argument precise.

\begin{lemma}[From observed logits and trace to the full Gram matrix]
\label{lem:tied-psd-completion}
For every symmetric $H$ satisfying $Q_\perp H Q_\perp=0$,
\begin{equation}
 \|H\|_\fro\le C\|\mathcal{L}(H)_{\mathcal S,:}\|_\fro.
 \label{eq:tied-tangent-observability}
\end{equation}
For every $K\succeq0$ satisfying
$\tr K\le\tr K^\star+\tau$, $\tau\ge0$, one has
\begin{equation*}
 \|K-K^\star\|_\fro
 \le C\{\|\mathcal{L}(K-K^\star)_{\mathcal S,:}\|_\fro+\tau\}.
\end{equation*}
Both statements hold regardless of the rank of $K$.
\end{lemma}

\begin{proof}
We first prove \cref{eq:tied-tangent-observability}. Let $H$ be a
symmetric matrix with $Q_\perp HQ_\perp=0$.

\textbf{Representation of $H$.}
Such a matrix can be written as $H=YV_0^\top+V_0Y^\top$ for some
$V_0\in\R^{V\times r_Y}$. For example, one can take
\[
 V_0=HY(Y^\top Y)^{-1}
       -\tfrac12Y(Y^\top Y)^{-1}(Y^\top HY)(Y^\top Y)^{-1}.
\]
Indeed, substituting this $V_0$ gives
$YV_0^\top+V_0Y^\top=H-Q_\perp HQ_\perp$, which equals $H$ by
assumption.

\textbf{Decomposition of the observed rows.}
Since $Y^\top\1=0$, we have
$\mathcal{L}(H)=Y\bigl((I_V-\1\1^\top/V)V_0\bigr)^\top+V_0Y^\top$.
Thus, after centering the rows indexed by $\mathcal S$, i.e.,
multiplying by $I_N-\1_N\1_N^\top/N$ from the left, the observed rows of
$\mathcal{L}(H)$ become
\[
 (Y_{\mathcal S}-\1_N\bar y_{\mathcal S}^\top)
 \bigl((I_V-\1\1^\top/V)V_0\bigr)^\top
 +(I_N-\1_N\1_N^\top/N)
 \bigl((I_V-\1\1^\top/V)V_0\bigr)_{\mathcal S,:}Y^\top.
\]
We decompose the $N\times r_Y$ matrix in the second term into its
projection onto the column space of the centered observed embeddings
$Y_{\mathcal S}-\1_N\bar y_{\mathcal S}^\top$ and the orthogonal
residual. Specifically, define
\[
 \begin{aligned}
 \Omega={}&(Y_{\mathcal S}^\top Y_{\mathcal S}
                  -N\bar y_{\mathcal S}\bar y_{\mathcal S}^\top)^{-1}
       (Y_{\mathcal S}-\1_N\bar y_{\mathcal S}^\top)^\top\\
 &\quad\cdot\bigl((I_V-\1\1^\top/V)V_0\bigr)_{\mathcal S,:},\\
 W_\perp={}&(I_N-\1_N\1_N^\top/N)
                \bigl((I_V-\1\1^\top/V)V_0\bigr)_{\mathcal S,:}
             -(Y_{\mathcal S}-\1_N\bar y_{\mathcal S}^\top)\Omega,\\
 V_1={}&(I_V-\1\1^\top/V)V_0+Y\Omega^\top.
 \end{aligned}
\]
Here, $\Omega\in\R^{r_Y\times r_Y}$ is the coefficient matrix of the
projection, $W_\perp\in\R^{N\times r_Y}$ is the residual, which
satisfies $(Y_{\mathcal S}-\1_N\bar y_{\mathcal S}^\top)^\top W_\perp=0$,
and $V_1\in\R^{V\times r_Y}$ is obtained from $(I_V-\1\1^\top/V)V_0$ by
adding the component $Y\Omega^\top$, which lies in the column space of
$Y$. With these definitions, we have
\[
 \begin{aligned}
 &(I_N-\1_N\1_N^\top/N)\mathcal{L}(H)_{\mathcal S,:}\\
 &\qquad=(Y_{\mathcal S}-\1_N\bar y_{\mathcal S}^\top)V_1^\top+W_\perp Y^\top
 \end{aligned}
\]
and the two terms on the right-hand side are orthogonal in the Frobenius
inner product, because
$(Y_{\mathcal S}-\1_N\bar y_{\mathcal S}^\top)^\top W_\perp=0$.
Therefore, each of them has Frobenius norm at most
$\|\mathcal{L}(H)_{\mathcal S,:}\|_\fro$. Since the smallest singular
value of $Y_{\mathcal S}-\1_N\bar y_{\mathcal S}^\top$ is at least
$\kappa\sqrt N$ by the coverage condition, and
$\sigma_{r_Y}(Y)\ge\sigma_{r_Y}(Y_{\mathcal S}-\1_N\bar y_{\mathcal S}^\top)$,
we obtain
\[
 \|V_1\|_\fro+\|W_\perp\|_\fro
 \le\frac C{\sqrt N}\|\mathcal{L}(H)_{\mathcal S,:}\|_\fro.
\]

\textbf{Bound on $\Omega+\Omega^\top$.}
By the definitions of $W_\perp$ and $V_1$,
\[
 \begin{aligned}
 (I_N-\1_N\1_N^\top/N)((I_V-\1\1^\top/V) V_0)_{\mathcal S,:}
 &=(Y_{\mathcal S}-\1_N\bar y_{\mathcal S}^\top)\Omega+W_\perp\\
 &=(I_N-\1_N\1_N^\top/N)(V_1)_{\mathcal S,:}\\
 &\quad-(Y_{\mathcal S}-\1_N\bar y_{\mathcal S}^\top)\Omega^\top,
 \end{aligned}
\]
which implies
$(Y_{\mathcal S}-\1_N\bar y_{\mathcal S}^\top)(\Omega+\Omega^\top)
=(I_N-\1_N\1_N^\top/N)(V_1)_{\mathcal S,:}-W_\perp$. Using the coverage
condition and the above bound on $\|V_1\|_\fro+\|W_\perp\|_\fro$, we
obtain
$\|\Omega+\Omega^\top\|_\fro
\le C\|\mathcal{L}(H)_{\mathcal S,:}\|_\fro/N$.

\textbf{Bound on $H$.}
Substituting $(I_V-\1\1^\top/V)V_0=V_1-Y\Omega^\top$ into
$(I_V-\1\1^\top/V)\mathcal{L}(H)
=Y\bigl((I_V-\1\1^\top/V)V_0\bigr)^\top
+\bigl((I_V-\1\1^\top/V)V_0\bigr)Y^\top$, we have
\[
 (I_V-\1\1^\top/V) \mathcal{L}(H)
 =YV_1^\top+V_1Y^\top-Y(\Omega+\Omega^\top)Y^\top,
\]
which has Frobenius norm at most
$C\|\mathcal{L}(H)_{\mathcal S,:}\|_\fro$, since
$\|Y\|_\op\le\sqrt{BV}$ and $N=\Theta(V)$. It remains to control the
part of $H$ that is removed by the centering. Since $Y^\top\1=0$, we
have $\1^\top H\1=0$, and thus
\[
 \begin{gathered}
 H=(I_V-\1\1^\top/V) \mathcal{L}(H)+\frac{\1(H\1)^\top+(H\1)\1^\top}{V},
 \\
 \mathcal{L}(H)_{\mathcal S,:}
 =((I_V-\1\1^\top/V) \mathcal{L}(H))_{\mathcal S,:}+\frac{\1_N(H\1)^\top}{V}.
 \end{gathered}
\]
The second identity gives
\[
 \begin{aligned}
 \frac{\sqrt N}{V}\|H\1\|_2
 &\le\|\mathcal{L}(H)_{\mathcal S,:}\|_\fro\\
 &\quad+\|(I_V-\1\1^\top/V) \mathcal{L}(H)\|_\fro\\
 &\le C\|\mathcal{L}(H)_{\mathcal S,:}\|_\fro.
 \end{aligned}
\]
Combining this bound with the first identity and the bound on
$(I_V-\1\1^\top/V)\mathcal{L}(H)$ proves
\cref{eq:tied-tangent-observability}.

\textbf{Bound on $K-K^\star$.}
Next, we prove the second statement. Let
$D_0=Q_\perp KQ_\perp\succeq0$. Since $K\succeq0$ and
$M_{\rm dual}\succeq Q_\perp$ by
\cref{lem:tied-anchor-trace-certificate}, we have
$\tr D_0=\langle Q_\perp,K\rangle\le\langle M_{\rm dual},K\rangle$.
Thus, the trace certificate in \cref{eq:tied-trace-certificate} gives
\[
 \begin{aligned}
 \tr D_0&\le\langle M_{\rm dual},K\rangle\\
 &=\tr(K-K^\star)\\
 &\quad-\langle Y_{\rm dual},\mathcal{L}(K-K^\star)_{\mathcal S,:}\rangle\\
 &\le\tau+C\|\mathcal{L}(K-K^\star)_{\mathcal S,:}\|_\fro,
 \end{aligned}
\]
where the last inequality follows from $\tr K\le\tr K^\star+\tau$ and
$\|Y_{\rm dual}\|_\fro\le C$. Let $H=(K-K^\star)-D_0$. Since
$Q_\perp K^\star Q_\perp=0$, we have $Q_\perp HQ_\perp=0$, and thus
\cref{eq:tied-tangent-observability} applies to $H$. Moreover, since
selecting rows and applying the centering operator $\mathcal{L}$ do not
increase the Frobenius norm, and $\|D_0\|_\fro\le\tr D_0$ for the
positive semidefinite matrix $D_0$, we have
$\|\mathcal{L}(H)_{\mathcal S,:}\|_\fro
\le\|\mathcal{L}(K-K^\star)_{\mathcal S,:}\|_\fro+\tr D_0$. Therefore,
\[
 \begin{aligned}
 \|K-K^\star\|_\fro
 &\le C\|\mathcal{L}(K-K^\star)_{\mathcal S,:}\|_\fro+(C+1)\tr D_0\\
 &\le C\{\|\mathcal{L}(K-K^\star)_{\mathcal S,:}\|_\fro+\tau\}.
 \end{aligned}
\]
This completes the proof.
\end{proof}

It remains to relate the error of the Gram matrix to the prediction
loss. When the learned embeddings have bounded norms, both the learned
logits and the ground-truth logits are bounded. Thus, on the line
segment between them, every probability is between constant multiples
of $1/V$, and the Hessian of the cross-entropy loss is bounded from
below and from above by constant multiples of $1/V$ on centered
vectors. We use the lower bound to convert the bound on the training
loss in Step 2 into a bound on the error of the logits on the observed
rows, and the upper bound to convert the error of the entire Gram
matrix into a bound on the loss on the unseen rows. The following lemma
summarizes the results of Steps 2 and 3. It is deterministic, i.e., it
holds for every $X$ that satisfies the comparison in Step 1.

\begin{lemma}[From objective comparison to test prediction]
\label{lem:iclr-transfer-mechanism}
Assume \cref{def:appendix-tied-class,def:appendix-transfer-coverage}, with
$N=\Theta(V)$. Fix $C_0>0$ and choose $\lambda>0$ with
$\lambda V$ sufficiently small. Then, every $X\in\R^{V\times d}$ satisfying
\begin{equation*}
 \cF_{\lambda,\mathcal S}(X)
 -\left(\cH_{\mathcal S}+\frac\lambda2\|Y\|_\fro^2\right)
 \le C_0(\lambda V)^2
\end{equation*}
satisfies $\cR_{\mathcal S}(X)-\cH_{\mathcal S}\le C(\lambda V)^2$.
If in addition $\max_j\|x_j\|^2\le R_0$, then
\begin{equation*}
 \|XX^\top-YY^\top\|_\fro\le C(\lambda V) V,
 \qquad \cR_{\mathcal U_0}(X)-\cH_{\mathcal U_0}
 \le\frac{C(\lambda V)^2V}{|\mathcal U_0|}
\end{equation*}
for every nonempty $\mathcal U_0\subseteq\mathcal U$ simultaneously.
Here, the constant $C$ may depend on $C_0$ and, in the last two bounds, on $R_0$.
\end{lemma}

\begin{proof}
\textbf{Training loss.}
We apply \cref{lem:tied-kl-lone} with $\nu=C_0(\lambda V)^2$. When
$\lambda V$ is sufficiently small, it gives
$\cR_{\mathcal S}(X)-\cH_{\mathcal S}\le C(\lambda V)^2$,
$(\tr K-\tr K^\star)_+\le2C_0\lambda V^2$, and
$\|(I_V-\1\1^\top/V) X\|_\fro^2/V\le C$. In particular, the bound on the
training excess loss does not require the bound on the learned
embeddings.

\textbf{Error on the observed rows.}
Suppose in addition that $\max_j\|x_j\|^2\le R_0$. Then, the learned
logits $K_{ij}=x_i^\top x_j$ lie in $[-R_0,R_0]$, and the ground-truth
logits $K^\star_{ij}$ lie in $[-B,B]$. Hence, for every row $i$ and
every logit vector on the line segment between $K^\star_{i,:}$ and
$K_{i,:}$, the corresponding softmax probabilities are between
$c_{R_0}/V$ and $C_{R_0}/V$. For such a probability vector $p$ and a
centered vector $v$, the variance satisfies
\[
 \frac{c_{R_0}}V\|v\|^2
 \le\operatorname{Var}_p(v)
 =\min_a\sum_jp_j(v_j-a)^2
 \le\frac{C_{R_0}}V\|v\|^2.
\]
Note that the gradient of $\CE(q_i,\cdot)$ vanishes at
$K^\star_{i,:}$, and its Hessian is the covariance matrix of the
corresponding softmax distribution. Thus, by Taylor's theorem with the
integral remainder, the lower bound on the variance, and the bound on
the training excess loss, we have
\[
 \begin{aligned}
 \frac{c_{R_0}}{2NV}\|\mathcal{L}(K-K^\star)_{\mathcal S,:}\|_\fro^2
 &\le\cR_{\mathcal S}(X)-\cH_{\mathcal S}\\
 &\le C(\lambda V)^2.
 \end{aligned}
\]
Consequently
\[
 \|\mathcal{L}(K-K^\star)_{\mathcal S,:}\|_\fro
 \le C_{R_0}\lambda V\sqrt{NV}\le C_{R_0}\lambda V^2.
\]

\textbf{Error of the entire Gram matrix.}
By the bound on the trace above, the assumption of
\cref{lem:tied-psd-completion} holds with $\tau=2C_0\lambda V^2$. Thus,
\cref{lem:tied-psd-completion} gives
$\|K-K^\star\|_\fro\le C_{R_0}\lambda V^2$, which is the bound on
$\|XX^\top-YY^\top\|_\fro$ in the lemma.

\textbf{Loss on the unseen rows.}
Finally, for any nonempty $\mathcal U_0\subseteq\mathcal U$, applying
Taylor's theorem again to the rows in $\mathcal U_0$, now with the
upper bound on the variance, we obtain
\[
 \begin{aligned}
 \cR_{\mathcal U_0}(X)-\cH_{\mathcal U_0}
 &\le\frac{C_{R_0}}{|\mathcal U_0|V}\|\mathcal{L}(K-K^\star)_{\mathcal U_0,:}\|_\fro^2\\
 &\le C_{R_0}(\lambda V)^2V/|\mathcal U_0|.
 \end{aligned}
\]
Since the bound on $\|K-K^\star\|_\fro$ does not depend on
$\mathcal U_0$, the result holds for all nonempty
$\mathcal U_0\subseteq\mathcal U$ simultaneously.

\end{proof}

We are now ready to prove \cref{thm:iclr-transfer}, which we restate
below for convenience.

\begin{theorem}[Restatement of \Cref{thm:iclr-transfer}: Conditional tied transfer]
Let $Q\in\mathcal Q^{\rm tied}_{r,B,\Delta_0}$ admit a representative $Y$
satisfying \cref{def:appendix-transfer-coverage}. Suppose
$\log V-\cH_{\mathcal S}\ge\delta_{\rm tr}>0$ and
$\log V-\cH_{\mathcal U}\ge\delta_{\rm te}>0$.
For sufficiently large $V$, let
$\max\{2r,CV/N^{1/3}\}\le d<N$ and
$\lambda=C_\lambda\sqrt{V/(Nd^3)}$.
Then, there exist a step size $h$, a perturbation scale $\sigma$, and a number of iterations $T$, with $h^{-1},\sigma^{-1}$, and $T$ bounded by polynomials in $V$, such that, with probability at
least $1-e^{-cd}$, every integer $t\in[T/d,T]$ satisfies
\begin{equation*}
 g_{\mathcal S}(X_t)\ge1-\frac{CV^3}{Nd^3\delta_{\rm tr}}.
\end{equation*}
Moreover, at every such iterate satisfying
$\max_{j\in[V]}\|x_j(t)\|_2^2\le R_0$ for a fixed $R_0$, the complete
unseen set satisfies
\begin{equation*}
 g_{\mathcal U}(X_t)\ge1-\frac{CV^4}{Nd^3(V-N)\delta_{\rm te}}.
\end{equation*}
The optimization analysis does not establish the learned-row bound.
The constant $C$ in the unseen-set conclusion may depend on $R_0$.
\end{theorem}

\begin{proof}
We choose the constants $C_\lambda$ and $C_{\rm wid}$ in
\cref{eq:tied-polynomial-decay} as in Step 1, so that
$\lambda\ge2C_p/d^2$ and
$\lambda V\le C_\lambda C_{\rm wid}^{-3/2}\le c_{\rm decay}$.
For sufficiently large $V$, the condition
$d\ge C_{\rm wid}V/N^{1/3}$ implies $d\ge C'\log(eV)$, and $d\ge2r$
implies $d\ge2r_Y$. Thus, all assumptions of
\cref{lem:tied-final-window-comparison} hold. By Step 1, there exist
deterministic $h$, $\sigma$, and $T$, which together with $h^{-1}$ and
$\sigma^{-1}$ are bounded by polynomials in $V$ since $N,d<V$ and
$(\lambda V)^{-1}=C_\lambda^{-1}\sqrt{Nd^3/V^3}<V$, and an event of
probability at least $1-e^{-cd}$, on which every integer
$t\in[T/d,T]$ satisfies
\[
 \cF_{\lambda,\mathcal S}(X_t)
 -\left(\cH_{\mathcal S}+\frac\lambda2\|Y\|_\fro^2\right)
 \le C(\lambda V)^2.
\]

Fix any integer $t\in[T/d,T]$ on this event. By
\cref{lem:iclr-transfer-mechanism}, we have
$\cR_{\mathcal S}(X_t)-\cH_{\mathcal S}\le C(\lambda V)^2$.
If in addition $\max_{j\in[V]}\|x_j(t)\|_2^2\le R_0$, then
\cref{lem:iclr-transfer-mechanism} with $\mathcal U_0=\mathcal U$ gives
$\cR_{\mathcal U}(X_t)-\cH_{\mathcal U}\le C(\lambda V)^2V/(V-N)$.
Substituting $(\lambda V)^2=C_\lambda^2V^3/(Nd^3)$, and dividing the two
bounds by $\log V-\cH_{\mathcal S}\ge\delta_{\rm tr}$ and
$\log V-\cH_{\mathcal U}\ge\delta_{\rm te}$, respectively, proves the
two gain bounds. Note that the second bound is a deterministic
implication at each iterate on the event, and the probability that the
bound on the learned embeddings holds is not used.
\end{proof}

\section{Auxiliary Results for the Main Proofs}
\label{app:auxiliary}

In this section, we collect the auxiliary results that are used in the
proofs of the main theorems. Appendix~\ref{app:factor-tools} shows that
a low-rank matrix with bounded entries admits a factorization with
bounded row norms, which is used to construct the comparator in
\cref{lem:shrunken-comparator}. Appendix~\ref{app:factor-derivatives}
provides the gradient and the Hessian of the untied objective, which
are used in Step 2 of Appendix~\ref{sec:positive}.
Appendix~\ref{app:technical} constructs the smooth extension used in
Step 1, and Appendix~\ref{app:optimization-tools} states the results on
perturbed gradient descent used in Step 3.

\subsection{Bounded Factorizations}
\label{app:factor-tools}

The following lemma shows that a matrix with rank at most $r$ and
entries bounded by $B$ admits a factorization whose rows have squared
norms at most $B\sqrt r$. In particular, the total squared norm of the
two factors is linear in $V$. This lemma is applied in
\cref{lem:shrunken-comparator} to construct the comparator. Note that,
in this lemma only, $X$ and $Y$ denote generic factors of the matrix
$M$.

\begin{lemma}[Rank versus max-norm]
\label{lem:maxnorm-factor}
If $M\in\mathbb R^{V\times V}$ has rank $r_0\le r$ and
$\|M\|_\infty\le B$, then there are
$X,Y\in\mathbb R^{V\times r_0}$ such that
\begin{equation}
 M=XY^\top,\qquad
 \max_i\|X_{i,:}\|_2^2\le B\sqrt{r},\qquad
 \max_j\|Y_{j,:}\|_2^2\le B\sqrt{r}.
 \label{eq:maxnorm-factor}
\end{equation}
Moreover, if $M\1=0$, the factors can be chosen with $Y^\top\1=0$.
Consequently,
\begin{equation}
 \|X\|_\fro^2+\|Y\|_\fro^2\le2B\sqrt{r} V.
 \label{eq:maxnorm-energy}
\end{equation}
\end{lemma}

\begin{proof}
If $r_0=0$, the claim holds trivially with empty factors. In the
following, we assume $r_0\ge1$. Let $M=A_0B_0^\top$ be a full-rank
factorization with $A_0,B_0\in\R^{V\times r_0}$, and let $a_i$ and $b_j$
denote the rows of $A_0$ and $B_0$, respectively. Let $\mathcal K$ be
the convex hull of $\{\pm b_j\}_j$, which is a centrally symmetric
convex body with nonempty interior, since $B_0$ has full column rank.
We use the L\"owner--John theorem for symmetric convex bodies
\citep{John2014}: if the minimum-volume ellipsoid containing a
centrally symmetric convex body $K\subset\R^n$ is the Euclidean unit
ball $\mathcal B_2$, then $n^{-1/2}\mathcal B_2\subseteq K\subseteq\mathcal B_2$.
By an invertible linear change of coordinates applied to the rows of
$B_0$, we can assume that the minimum-volume ellipsoid containing
$\mathcal K$ is the unit ball in $\R^{r_0}$, so that
\[
 r_0^{-1/2}\mathcal B_2\subseteq \mathcal K\subseteq\mathcal B_2.
\]
We apply the inverse transpose of this change of coordinates to the rows
of $A_0$, so that all inner products $a_i^\top b_j=M_{ij}$ are
preserved. Then, we have $\|b_j\|_2\le1$ and $|a_i^\top b_j|\le B$ for
all $i,j$. Let
$\mathcal K^\circ=\{z:\sup_{b\in\mathcal K}z^\top b\le1\}$ be the polar
set of $\mathcal K$. The second bound implies
$a_i/B\in \mathcal K^\circ$, and the inclusion
$r_0^{-1/2}\mathcal B_2\subseteq \mathcal K$ implies
$\mathcal K^\circ\subseteq\sqrt{r_0}\mathcal B_2$. Thus,
$\|a_i\|_2\le B\sqrt{r_0}$. Finally, we rescale the two factors by
reciprocal constants such that the maximum row norms of both factors
are at most $\sqrt{B\sqrt{r_0}}\le\sqrt{B\sqrt r}$, and denote the
resulting factors by $X$ and $Y$. This proves
\cref{eq:maxnorm-factor}.

For the remaining claims, note that $X$ has full column rank. Thus, if
$M\1=0$, then $X(Y^\top\1)=0$ implies $Y^\top\1=0$. Summing the bounds
on the squared row norms over the $V$ rows of each factor gives
\cref{eq:maxnorm-energy}.
\end{proof}

\subsection{Untied Factor Derivatives}
\label{app:factor-derivatives}

The following lemma provides the gradient and the Hessian of the untied
objective, as well as some properties of its critical points. The
Hessian identity in \cref{eq:factor-hessian} is the starting point of
Step 2 in Appendix~\ref{sec:untied-step2}. Note that, although the
cross-entropy loss is convex in the logits, the objective is not convex
in $(E,U)$, because the bilinear term in \cref{eq:factor-hessian} can be
negative.

\begin{lemma}[Gradient, balance, and Hessian identities]
\label{lem:factor-calculus}
Fix $\lambda>0$, and let $\cF_\lambda=\cF_{\lambda,[V]}$ be the untied
objective with $\Theta=(E,U)$. Its gradient is given by
\begin{equation}
 \nabla_E\cF_\lambda=\frac1V(P^{E,U}-Q)U+\lambda E,
 \qquad
 \nabla_U\cF_\lambda=\frac1V(P^{E,U}-Q)^\top E+\lambda U.
 \label{eq:factor-gradients}
\end{equation}
For any direction $(\dot E,\dot U)$ with the same dimensions as $(E,U)$,
the Hessian satisfies
\begin{equation}
 \begin{aligned}
 \nabla^2\cF_\lambda(E,U)[(\dot E,\dot U),(\dot E,\dot U)]
 &=\frac1V\sum_{i=1}^V\operatorname{Var}_{j\sim P^{E,U}_{i,:}}
 \bigl[(\dot E U^\top+E\dot U^\top)_{ij}\bigr]\\
 &\quad+\frac2V\langle P^{E,U}-Q,\dot E\dot U^\top\rangle
 +\lambda(\|\dot E\|_\fro^2+\|\dot U\|_\fro^2).
 \end{aligned}
 \label{eq:factor-hessian}
\end{equation}
Here, the variance is over the index $j$ drawn from the $i$-th row of
$P^{E,U}$. We refer to the three terms on the right-hand side as the
softmax curvature term, the bilinear term, and the weight decay term,
respectively. The softmax curvature term is nonnegative. Moreover, at
every critical point,
\begin{align*}
 \1^\top U&=0,\qquad E^\top E=U^\top U,\\
 \|EU^\top\|_*&=\frac12(\|E\|_\fro^2+\|U\|_\fro^2),\\
 \frac1V\langle P^{E,U}-Q,EU^\top\rangle
 &+\lambda\|EU^\top\|_*=0.
\end{align*}
\end{lemma}

\begin{proof}
The gradient of the loss with respect to the logits $EU^\top$ is
$(P^{E,U}-Q)/V$, and thus the chain rule gives
\cref{eq:factor-gradients}. For the Hessian, the first and second
variations of $EU^\top$ along $(\dot E,\dot U)$ are
$\dot E U^\top+E\dot U^\top$ and $2\dot E\dot U^\top$, respectively, and
the Hessian of the cross-entropy loss of row $i$ with respect to its
logits is the covariance matrix
$\operatorname{diag}(P^{E,U}_{i,:})
-(P^{E,U}_{i,:})^\top P^{E,U}_{i,:}$. Combining these facts with the
chain rule gives \cref{eq:factor-hessian}.

Next, consider a critical point. Multiplying the second equation in
\cref{eq:factor-gradients} by $\1^\top$ from the left, and using
$(P^{E,U}-Q)\1=0$ and $\lambda>0$, we obtain $\1^\top U=0$.
Multiplying the first equation in \cref{eq:factor-gradients} by
$E^\top$ from the left and taking the transpose, and comparing the
result with the second equation multiplied by $U^\top$ from the left,
we obtain $E^\top E=U^\top U$. Since the two Gram matrices coincide,
their eigenvalues are the singular values of $EU^\top$, which gives the
identity for the nuclear norm. Finally, taking the inner product of
\cref{eq:factor-gradients} with $(E,U)$ and dividing by two gives the
last identity.
\end{proof}
\subsection{Smooth Localization}
\label{app:technical}

The results on perturbed gradient descent in
Appendix~\ref{app:optimization-tools} require the gradient and the
Hessian of the objective to be globally Lipschitz, while our objectives
are only controlled on a ball. The following lemma constructs an
extension that coincides with the objective in the ball, remains large
outside the ball, and is globally smooth. We note that the extension is
only used in the analysis, and the algorithm is run on the original
objective.

\begin{lemma}[Smooth extension outside a ball]
\label{lem:localization}
Suppose that the ball $\{\Theta:\|\Theta\|_\fro^2\le CV\}$ contains
\(\Theta_0\) and a global minimizer of \(\cF_\lambda\), and that
\[
 \cF_\lambda(\Theta)\ge \cF_\lambda(\Theta_0)+4
 \qquad\text{whenever }\|\Theta\|_\fro^2\ge CV.
\]
Then, for any constant $c>0$ independent of $V$, there exists a smooth
function \(\widetilde{\cF}_\lambda:\R^{D_\Theta}\to\R\) such that
(i) \(\widetilde{\cF}_\lambda=\cF_\lambda\) on the ball;
(ii) \(\widetilde{\cF}_\lambda\ge\cF_\lambda(\Theta_0)+2\) outside the
ball;
(iii) \(\widetilde{\cF}_\lambda\ge \cF_\lambda^*\) everywhere; and
(iv) the gradient and the Hessian of \(\widetilde{\cF}_\lambda\) are
globally Lipschitz.
Moreover, if the derivatives of \(\cF_\lambda\) up to order three on the
ball of radius $(1+c)\sqrt{CV}$ are polynomial in \(V\), then the
Lipschitz constants in (iv) are polynomial in \(V\). Finally, if
\(\cF_\lambda\ge M\) outside the ball for some constant $M$, then the
extension can be chosen such that \(\widetilde{\cF}_\lambda\ge M\)
outside the ball as well.
\end{lemma}

\begin{proof}
Let \(R=\sqrt{CV}\). Let \(s:[0,\infty)\to[0,\infty)\) be a
\(C^\infty\) function such that $s(a)=a$ for $a\in[0,R]$,
$s(a)\in[R,(1+c)R]$ for $a\ge R$, and $s$ is constant for
$a\ge(1+2c)R$. Such a function exists by a standard smoothing argument.
Define
\[
 \widetilde{\cF}_\lambda(\Theta)=
 \begin{cases}
  \cF_\lambda\!\left(s(\|\Theta\|_\fro)\Theta/\|\Theta\|_\fro\right),
       &\Theta\ne0,\\
  \cF_\lambda(0),&\Theta=0.
 \end{cases}
\]
The map \(\Theta\mapsto s(\|\Theta\|_\fro)\Theta/\|\Theta\|_\fro\) is the
identity on the ball $\|\Theta\|_\fro\le R$, and maps every point
outside the ball to the annulus $R\le\|\Theta\|_\fro\le(1+c)R$. It is
smooth at the origin because it is the identity in a neighborhood of
the origin. Thus, \(\widetilde{\cF}_\lambda=\cF_\lambda\) on the ball.
Outside the ball, \(\widetilde{\cF}_\lambda(\Theta)\) is the value of
\(\cF_\lambda\) at a point with squared norm at least $CV$, which is at
least \(\cF_\lambda(\Theta_0)+4\) by assumption, and in particular at
least \(\cF_\lambda(\Theta_0)+2\). For the same reason, any lower bound
\(M\) of \(\cF_\lambda\) outside the ball is preserved. Moreover, since
every value of \(\widetilde{\cF}_\lambda\) is a value of
\(\cF_\lambda\), we have \(\widetilde{\cF}_\lambda\ge\cF_\lambda^*\),
with equality at the global minimizer in the ball.

It remains to bound the Lipschitz constants. Since $c$ is a constant,
the first, second, and third derivatives of the above map have norms
$O(1)$, $O(R^{-1})$, and $O(R^{-2})$, respectively. Its image is
contained in the ball of radius $(1+c)R$, on which the derivatives of
\(\cF_\lambda\) up to order three are bounded. Thus, by the chain rule,
the second and third derivatives of \(\widetilde{\cF}_\lambda\) are
globally bounded, which gives the Lipschitz constants of the gradient
and the Hessian. If the derivatives of \(\cF_\lambda\) are polynomial in
\(V\), so are these constants.
\end{proof}
\subsection{Perturbed-Gradient Visits and Objective Increases}
\label{app:optimization-tools}

In this subsection, we state two results of \citet{jin2019nonconvex} on
perturbed gradient descent, and combine them into the one-block
estimate that is used in Step 3 of our proofs. We state the results for
a generic objective $f:\R^{D_\Theta}\to\R$ with exact gradients, step
size $h$, and perturbation scale $\sigma$. The first result
(\cref{thm:jin-visit}) shows that perturbed gradient descent visits an
approximate SOSP within a bounded number of iterations. The second
result (\cref{lem:jin-descent}) controls how much the objective can
increase over a given number of iterations.

\begin{theorem}[Visiting an approximate SOSP \citep{jin2019nonconvex}]
\label{thm:jin-visit}
Let $f:\R^{D_\Theta}\to\R$ have an $\ell$-Lipschitz gradient and a
$\rho$-Lipschitz Hessian, and let
$f(\Theta_0)-\inf_\Theta f(\Theta)<\infty$.  For every $\epsilon>0$ and failure
level $\zeta_{\rm J}>0$, Algorithm~1 and Theorem~13 of
\citet{jin2019nonconvex}, specialized to exact gradients, show that
\begin{equation*}
 \Theta_{t+1}=\Theta_t-h\{\nabla f(\Theta_t)+\xi_t\},\qquad
 \xi_t\stackrel{\rm iid}{\sim}\mathcal N(0,\sigma^2I_{D_\Theta}/D_\Theta),
\end{equation*}
with $h=\widetilde\Theta(1/\ell)$ and
$\sigma=\widetilde\Theta(\epsilon)$ chosen as in
\citet{jin2019nonconvex}, visits a point satisfying
\begin{equation*}
 \|\nabla f(\Theta_t)\|\le\epsilon,\qquad
 \nabla^2f(\Theta_t)\succeq-\sqrt{\rho\epsilon}\,I
\end{equation*}
within $\widetilde O\bigl(\ell[f(\Theta_0)-\inf f]\epsilon^{-2}\bigr)$ iterations, with
probability at least $1-\zeta_{\rm J}$.  The hidden factors are
polylogarithmic in
$D_\Theta,\ell,\rho,\epsilon^{-1},\zeta_{\rm J}^{-1}$, and $f(\Theta_0)-\inf f$.
\end{theorem}

Note that \cref{thm:jin-visit} only guarantees one iterate that is an
approximate SOSP, and does not control the other iterates. To this end,
we use the following concentration inequality, which bounds the
increase of the objective over any fixed interval of iterations.

\begin{lemma}[Descent concentration \citep{jin2019nonconvex}]
\label{lem:jin-descent}
Under the assumptions of \cref{thm:jin-visit}, fix deterministic integers
$t_0\ge0$ and $s\ge1$.  If $h\le1/\ell$, Lemma~23 of
\citet{jin2019nonconvex}, again specialized to exact gradients, gives
\begin{equation}
 f(\Theta_{t_0+s})-f(\Theta_{t_0})
 \le-\frac h8\sum_{j=0}^{s-1}\|\nabla f(\Theta_{t_0+j})\|^2
 +c_Jh\sigma^2(h\ell s+\iota)
 \label{eq:jin-descent-concentration}
\end{equation}
with probability at least $1-4e^{-\iota}$, where $c_J$ is universal.
\end{lemma}

\begin{proof}
This is Lemma~23 of \citet{jin2019nonconvex}, where the variance of the
stochastic gradient is set to zero. The perturbation there is the
isotropic Gaussian $\mathcal N(0,\sigma^2I_{D_\Theta}/D_\Theta)$ used in
\cref{thm:jin-visit}, i.e., $\sigma^2$ is the total second moment of
the perturbation. With these choices, the bound in that lemma and its
failure probability are exactly \cref{eq:jin-descent-concentration}.
\end{proof}

\paragraph{Specialization of the parameters.}
We now specify the parameters in the above results. Since we use exact
gradients, the variance of the stochastic gradient in
\citet{jin2019nonconvex} is zero. In this case, the quantity denoted by
$N$ in Appendix~B of \citet{jin2019nonconvex}, which should not be
confused with our number of observed states, equals one. Let
$C_J\ge1$ be a universal constant that is sufficiently large to absorb
the numerical constants in the parameter choices of
\citet{jin2019nonconvex}. Then, those choices reduce to
\begin{equation*}
 h=(\iota^9\ell)^{-1},\qquad \sigma=\iota\epsilon.
\end{equation*}
Consequently, when the initial objective gap is at most $A$, the escape
time and the scale of the objective decrease in
\citet{jin2019nonconvex} give the following number of iterations for
visiting an approximate SOSP:
\begin{equation}
 A\max\left\{
 \frac{\iota/(h\sqrt{\rho\epsilon})}
      {\iota^{-5}\sqrt{\epsilon^3/\rho}},
 \frac1{h\epsilon^2}\right\}
 =O(\ell A\epsilon^{-2}\iota^{15}).
 \label{eq:jin-horizon-specialization}
\end{equation}
This is the length of the blocks used below.

We now combine \cref{thm:jin-visit,lem:jin-descent} into an estimate for
one block of iterations. In our proofs, we need to bound the increase
of the objective from the iteration at which an approximate SOSP is
visited, which is random, to the end of the block. Since
\cref{lem:jin-descent} only applies to deterministic intervals, we take
a union bound over all pairs of iterations $u<v$ in the block, so that
the bound holds for all pairs simultaneously.

\begin{lemma}[One-block estimate]
\label{thm:tied-fixed-noise-consequence}
Let $f:\R^{D_\Theta}\to\R$ be bounded from below, with $\ell$-Lipschitz
gradient and $\rho$-Lipschitz Hessian, where $\ell,\rho\ge1$. Suppose
that $f(\Theta_0)-\inf f\le A$ for some $A\ge1$.
For $0<\epsilon,\delta_c\le1$ and $0<\zeta<1/4$, let
\begin{align}
 \iota&\ge C\left[1+
 \log\frac{eD_\Theta(\ell+\rho+1)(A+1)}{\epsilon\zeta}
 +\frac{A+1}{\delta_c}+\delta_c^{-1/6}\right],\nonumber\\
 h&=(\iota^9\ell)^{-1},\qquad \sigma=\iota\epsilon,
 \qquad
 T_{\rm blk}=\left\lceil C_J\ell A\epsilon^{-2}\iota^{15}\right\rceil,
 \label{eq:predeclared-schedule}
\end{align}
where $C$ is a sufficiently large universal constant, and $\iota$ is
further enlarged, if necessary, such that
$4(T_{\rm blk}+1)^2e^{-\iota}\le\zeta$.
Then, with probability at least $1-2\zeta$, the following two events
occur for the iterates $\Theta_0,\ldots,\Theta_{T_{\rm blk}}$ of
perturbed gradient descent: (i) there exists an iterate $\Theta_t$ with
$\|\nabla f(\Theta_t)\|\le\epsilon$ and
$\nabla^2f(\Theta_t)\succeq-\sqrt{\rho\epsilon}I$; (ii)
$f(\Theta_v)-f(\Theta_u)\le\delta_c$ for all
$0\le u<v\le T_{\rm blk}$.
Moreover, the same statement holds for a block that starts from any
iteration $t_0$, conditional on the history up to $t_0$, provided that
$f(\Theta_{t_0})-\inf f\le A$.
\end{lemma}

\begin{proof}
\textbf{Visit.}
We apply \cref{thm:jin-visit} with exact gradients. As discussed above,
the step size and the perturbation scale in \citet{jin2019nonconvex}
reduce to those in \cref{eq:predeclared-schedule}, and by
\cref{eq:jin-horizon-specialization}, an approximate SOSP is visited
within $T_{\rm blk}=\lceil C_J\ell A\epsilon^{-2}\iota^{15}\rceil$
iterations with probability at least $1-\zeta$. Here, the lower bound
on $\iota$ in \cref{eq:predeclared-schedule} absorbs the logarithmic
dependence on the failure probability in \cref{thm:jin-visit}.

\textbf{Objective increase.}
Fix any deterministic pair $0\le u<v\le T_{\rm blk}$, and let $s=v-u$.
By \cref{lem:jin-descent}, with probability at least $1-4e^{-\iota}$,
\[
 f(\Theta_v)-f(\Theta_u)
 \le-\frac h8\sum_{j=0}^{s-1}\|\nabla f(\Theta_{u+j})\|^2
   +c_Jh\sigma^2(h\ell s+\iota).
\]
With the choices in \cref{eq:predeclared-schedule} and
$s\le T_{\rm blk}$, the second term on the right-hand side is at most
$C\{(A+1)/\iota+\epsilon^2/(\ell\iota^6)\}\le\delta_c$, where the last
inequality is due to the lower bound on $\iota$. Taking the union bound
over at most $(T_{\rm blk}+1)^2$ pairs, the failure probability is at
most $4(T_{\rm blk}+1)^2e^{-\iota}\le\zeta$. Since the bound holds for
all pairs simultaneously, it also holds when $u$ is the random
iteration at which the approximate SOSP is visited.

\textbf{Conclusion.}
By the union bound, both events occur with probability at least
$1-2\zeta$. For a block that starts from iteration $t_0$, conditioning
on the history up to $t_0$ does not change the distribution of the
future perturbations, which are independent Gaussian vectors. Thus, the
same argument applies conditionally. Finally, we note that the
additional requirement $4(T_{\rm blk}+1)^2e^{-\iota}\le\zeta$ is
satisfied by some $\iota$ of polynomial size, since $T_{\rm blk}$ is
polynomial in $\iota$, whereas $e^{-\iota}$ decays exponentially.
\end{proof}

\section{Experimental Protocol and Complete Summaries}
\label{app:experiments}

In this section, we provide the details of the population-gradient
experiments in \cref{sec:iclr-experiments}. The experiments use the
same models and the same perturbed gradient descent algorithm as in our
theory, but with practical parameter values instead of the constants
required by the proofs. All seeds and configurations were fixed before
running the reported experiments, and all runs are reported. A
preliminary check of the code used different seeds for the ground
truths and the initializations, and did not change the reported
configurations. No ground truths, runs, or checkpoints were selected
based on the test results.

\subsection{Ground truths, observed states, and initialization}

\paragraph{Ground truths.}
For each $V\in\{64,128,256\}$, we generate three ground truths with
seeds $s\in\{0,1,2\}$, using NumPy's default random generator with seed
$10000+100V+s$. We first draw $V\times4$ matrices with independent
standard Gaussian entries, and normalize each row to unit norm. For the
untied model trained on all states, we draw two such matrices, subtract
the mean row from the second matrix, and divide it by its largest row
norm. The logit matrix is three times the product of the first matrix
and the transpose of the second matrix, and $Q$ is obtained by applying
the row-wise softmax to it. Since the logit matrix is row-centered, it
equals $\mathcal{L}(\log Q)$. Thus, the centered logits have rank four
and entries bounded by three. Note that the rank condition is on the
centered logits, and not on $Q$ itself.

For the tied model, and for the untied model that is trained on the
same observed states as a baseline, we subtract the mean row from the
first normalized matrix, and rescale it such that its largest row norm
is $\sqrt3$. We denote the result by $Y$, and set
$Q=\softmax(YY^\top)$. The set $\mathcal S$ consists of the first
$3V/4$ indices of a random permutation drawn from the same generator,
and $\mathcal U=[V]\setminus\mathcal S$ is fixed before training.
Recall that all $V$ states are possible next states, and ``unseen''
only refers to the role as a current state. For every ground truth, we
check and save the rank and the bound of the centered logits, the
smallest eigenvalue of the covariance matrix in
\cref{def:appendix-transfer-coverage}, and the information gaps. No
ground truth is rejected. Over the nine tied ground truths, the
smallest eigenvalue of the covariance matrix is $0.28957$, and the
smallest information gap over the observed and the unseen states is
$0.47702$.

\paragraph{Initialization.}
For each ground truth, we use five initialization seeds
$a\in\{10,11,12,13,14\}$, with NumPy generator seed $1000000+1000s+a$.
All entries of $E$ and $U$, or of $X$, are drawn independently from
$\mathcal N(0,1/d)$ and stored in single precision (FP32). The
embeddings are not centered, balanced, projected, normalized, frozen,
or aligned with the ground truth at any time. For the untied model, the
gradients with respect to $E$ and $U$ are computed at the same current
parameters before both of them are updated.

\subsection{Training configurations and control runs}

\paragraph{Configurations.}
The main grid consists of $d\in\{V/8,V/4,V/2\}$ for each of the three
values of $V$ and each of the two models, i.e., 18 settings with 15
runs each. All runs use $T=8192$ iterations, step size $h=V/4$, and
perturbation scale $\sigma=0.01/\sqrt V$. The weight decay is
\begin{equation*}
 \lambda=\frac{1}{480V}\quad\text{(untied, all states)},\qquad
 \lambda=0.1\sqrt{\frac{V}{Nd^3}}\quad\text{(tied, and untied on the same split)}.
\end{equation*}
The first value is the weight decay in \cref{thm:iclr-untied} with
$r=4$, $B=3$, and $\Delta_0=0.1$. The second value follows the scaling
in \cref{thm:iclr-transfer} with a practical constant. We note that the
constants required by our proofs for the tied model are not satisfied
in these experiments. Specifically, with $C_p=40B_Yr=480$, the
conditions $\lambda\ge2C_p/d^2$ and $\lambda V\le1$ would require
$d^2\ge960V$, which does not hold for the embedding dimensions that we
test. Thus, the tied experiments illustrate the behavior of the model
beyond the range of constants covered by our theorems.

\paragraph{Training.}
The loss is the population cross-entropy loss, i.e., it sums over all
$V$ next states and averages uniformly over the current states used for
training, and no transitions are sampled. The table $Q$ is generated in
double precision (FP64), and then rounded to FP32 and renormalized in
each row for training. The checks of the ground truths and the final
evaluation of the saved checkpoints use the original FP64 table. The
weight decay penalty is $\lambda/2$ times the squared norm of all
parameters, without any per-parameter normalization. At every update,
we add a fresh Gaussian vector with covariance $\sigma^2 I/D_\Theta$ to
the gradient, and then multiply the sum by $h$, where $D_\Theta=2Vd$
for the untied model and $D_\Theta=Vd$ for the tied model. The
perturbations are generated with seed $271828+100V+d$, independently of
the initialization, and the same sequence of perturbations is shared by
the paired runs with the same parameter dimensions. The perturbations
are added throughout training, and there is no stopping rule based on
the loss or the test data.

\paragraph{Control runs.}
At $V=128$ and $d=32$, we add three control settings for each model:
without perturbation, without weight decay, and without both. These 90
runs use the same step size, number of iterations, ground truths, and
initializations as the main runs. In addition, 15 untied runs use the
ground truths, the split, and the parameters of the tied model, in
order to compare the tied and untied models on the same transfer
problem. In total, there are 25 settings and 375 runs. Note that the
seeds are shared across the settings so that the runs are paired, and
thus the 375 runs are not independent trials. The preliminary check of
the code used $V=64$, $d=16$, 256 updates, ground-truth seed 90, and
initialization seeds 91 and 92, with the same formula for the seeds of
the perturbations.

\subsection{Measurements and numerical checks}

\paragraph{Measurements.}
At every update $t=0,\ldots,T$, we record the training gain, the gain on
the unseen states when applicable, and the largest squared row norm of
the embeddings. The final window is $[T/d,T]$, including both
endpoints. For each run, we compute the minimum gain, the fraction of
iterations with gain above $7/8$, and the longest interval of
consecutive iterations with gain below $7/8$ in this window. The
parameters are saved at $t=0$, $T/d$, and $T$, and the objective, the
squared norm per state, and the largest scaled predicted probability
are recorded every 16 updates. When the model is trained on all states,
there is no test gain to report. If a run diverges, it is still
counted, and its statistics are recorded as missing. No run diverged in
our experiments.

\paragraph{Numerical checks.}
The closed-form gradients agree with automatic differentiation, with a
maximum absolute error of $1.12\times10^{-16}$ in FP64. We also test the
variance of the perturbations, the invariance of the softmax to the
centering of the logits, and the gradient identity for the input
embeddings of the unseen states. After the runs, all reported
statistics in the final window are recomputed from the complete traces,
whose SHA-256 hashes are verified, and the risks with respect to the
original ground truths are evaluated independently in FP64 at all 1125
saved checkpoints. The largest discrepancy in the gain is
$2.99\times10^{-7}$.

\paragraph{{Computing environment.}}
{The experiments were run on} an NVIDIA RTX 3090 {GPU with} PyTorch
2.10.0+cu128 {and} NumPy 2.2.6, {using} FP32 parameters with deterministic
algorithms enabled and TF32 disabled. The total wall-clock time of the
training loops is 40.7 seconds, which includes the transfer of the
checkpoints, and excludes the file compression and the checks after the
runs. This number is only a rough estimate of the running time, since
the timer is not synchronized with the GPU. The code, the versions of
the dependencies, the raw traces, the checkpoints, and the commands for
generating the figures are included in the supplementary material. The
results may differ at the level of floating-point rounding on different
hardware or library versions.

\begin{table}[t]
 \centering
 \caption{Results of all settings, with 15 runs each. ``Min.'' is the
minimum gain over all runs and all iterations in the final window
$[T/d,T]$. ``Kept'' is the number of runs whose gain is above $7/8$
throughout the final window (training/test). ``Max. row'' is the
largest squared row norm of the embeddings over the entire run, which
is below the bound of eight in all runs. In the column ``Arm'',
``Both'' means that both the weight decay and the perturbation are
used, ``Decay'' and ``Noise'' mean that only one of them is used, and
``Neither'' means that none of them is used. Untied-H denotes the
untied model trained on the ground truths and the split of the tied
model. A missing test entry means that all current states are used for
training. The values are rounded for display, and the comparisons with
$7/8$ use the unrounded values.}
 \label{tab:experiment-all-settings}
 \small
 \begin{tabular}{llrrcccr}
\toprule
Model & Arm & $V$ & $d$ & Min. train & Min. test & Kept tr./te. & Max. row\\
\midrule
Untied & Both & 64 & 8 & 0.9999 & -- & 15/-- & 4.493\\
Tied & Both & 64 & 8 & 0.8607 & 0.6150 & 0/0 & 3.246\\
Untied & Both & 64 & 16 & 0.9998 & -- & 15/-- & 4.271\\
Tied & Both & 64 & 16 & 0.9762 & 0.8824 & 15/15 & 2.869\\
Untied & Both & 64 & 32 & 0.9993 & -- & 15/-- & 4.302\\
Tied & Both & 64 & 32 & 0.9964 & 0.9728 & 15/15 & 3.124\\
Untied & Both & 128 & 16 & 0.9998 & -- & 15/-- & 4.313\\
Tied & Both & 128 & 16 & 0.9228 & 0.8024 & 15/0 & 2.522\\
Untied & Both & 128 & 32 & 0.9993 & -- & 15/-- & 4.217\\
Tied & Both & 128 & 32 & 0.9883 & 0.9636 & 15/15 & 3.020\\
Untied & Both & 128 & 64 & 0.9978 & -- & 15/-- & 4.227\\
Tied & Both & 128 & 64 & 0.9982 & 0.9945 & 15/15 & 3.491\\
Untied & Both & 256 & 32 & 0.9993 & -- & 15/-- & 4.254\\
Tied & Both & 256 & 32 & 0.9587 & 0.8962 & 15/15 & 2.694\\
Untied & Both & 256 & 64 & 0.9976 & -- & 15/-- & 4.210\\
Tied & Both & 256 & 64 & 0.9941 & 0.9835 & 15/15 & 3.215\\
Untied & Both & 256 & 128 & 0.9951 & -- & 15/-- & 4.356\\
Tied & Both & 256 & 128 & 0.9974 & 0.9925 & 15/15 & 3.733\\
Untied & Decay & 128 & 32 & 0.9993 & -- & 15/-- & 4.221\\
Tied & Decay & 128 & 32 & 0.9884 & 0.9638 & 15/15 & 3.020\\
Untied & Noise & 128 & 32 & 0.9992 & -- & 15/-- & 4.306\\
Tied & Noise & 128 & 32 & 0.9998 & 0.9754 & 15/15 & 4.041\\
Untied & Neither & 128 & 32 & 0.9992 & -- & 15/-- & 4.271\\
Tied & Neither & 128 & 32 & 0.9998 & 0.9758 & 15/15 & 4.044\\
Untied-H & Both & 128 & 32 & 0.9708 & -0.0013 & 15/0 & 3.123\\
\bottomrule
\end{tabular}

\end{table}

\begin{figure}[t]
 \centering
 \includegraphics[width=\linewidth]{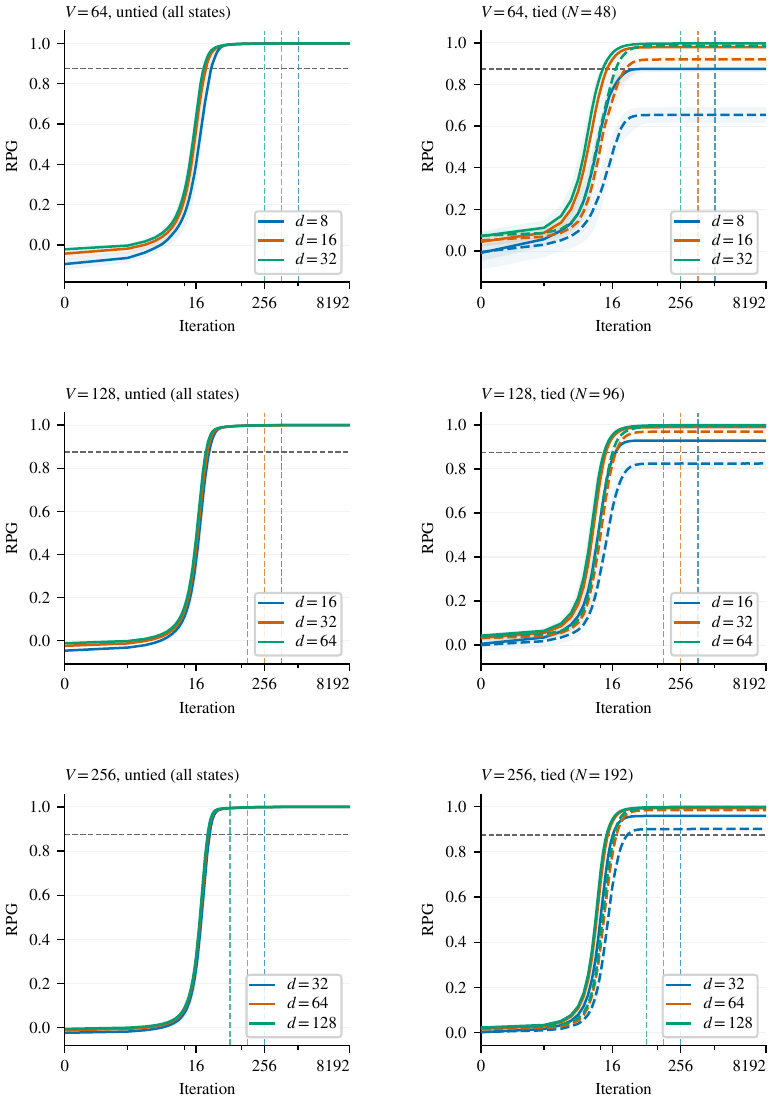}
 \caption{{Trajectories of all settings in the main grid}, shown as medians
 {(curves)} and full ranges {(shaded areas)}. Each panel contains {the} three
 {embedding dimensions}. {The untied model is trained on} all {current}
 states{. For the} tied {model, the} solid {and} dashed curves {show the}
 training gain {and the gain on the unseen states, respectively}. The
 {perturbations are added} throughout {training}. {The curves are subsampled}
 for {plotting} only{, and} all reported {statistics are computed over all
 updates}.}
 \label{fig:experiment-all-traces}
\end{figure}

\paragraph{{Summary of the results.}}
\Cref{tab:experiment-all-settings} and \cref{fig:experiment-all-traces}
show that the untied model trained on all states maintains a high gain
throughout the {final window in all runs. For the tied model, the gain
on the unseen states improves as the embedding dimension increases},
while the {setting with the smallest embedding dimension has a} lower
training gain{. In contrast, the untied model trained on the same
observed states has a gain close to zero on the unseen states. In all
runs, the squared row norms of the embeddings are bounded, which shows
that the condition on the learned embeddings in
\cref{thm:iclr-transfer} holds in these instances. Finally, the control
runs achieve comparable gains without perturbations, and a} lower final
training error without {weight decay. We note that these experiments are
not designed to estimate the optimal dependence on the embedding
dimension or the failure probability}.


\section{Scaling with Sampled Transitions}
\label{app:scaling-sgd}

\begin{figure}[p]
 \centering
 \includegraphics[width=\linewidth,height=0.85\textheight,keepaspectratio]{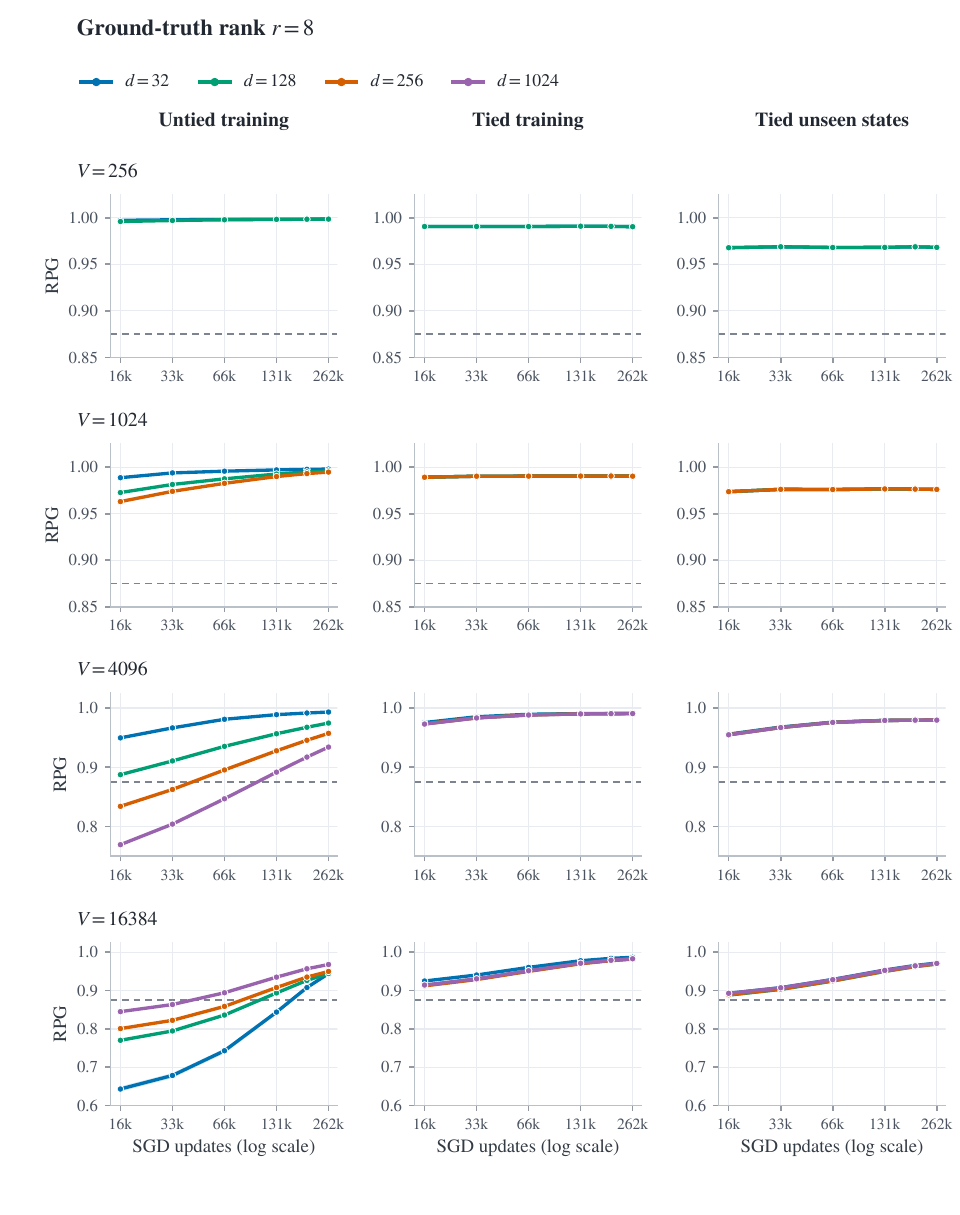}
 \caption{SGD with sampled transitions at rank $r=8$. The rows correspond
 to $V=256,1024,4096,16384$ from top to bottom{, and the} columns show the
 training gain of the untied model, the training gain of the tied
 model, and {the gain of the} tied {model on the} unseen states, from left
 to right. All valid embedding dimensions and all three initialization
 seeds are included. {The curves} show {the} medians {over the seeds,} and
 {the shaded areas show} their full ranges. The dashed line marks $7/8${,
 and the vertical} scales may differ {across rows}.}
 \label{fig:scaling-sgd-r8}
\end{figure}

\begin{figure}[p]
 \centering
 \includegraphics[width=\linewidth,height=0.85\textheight,keepaspectratio]{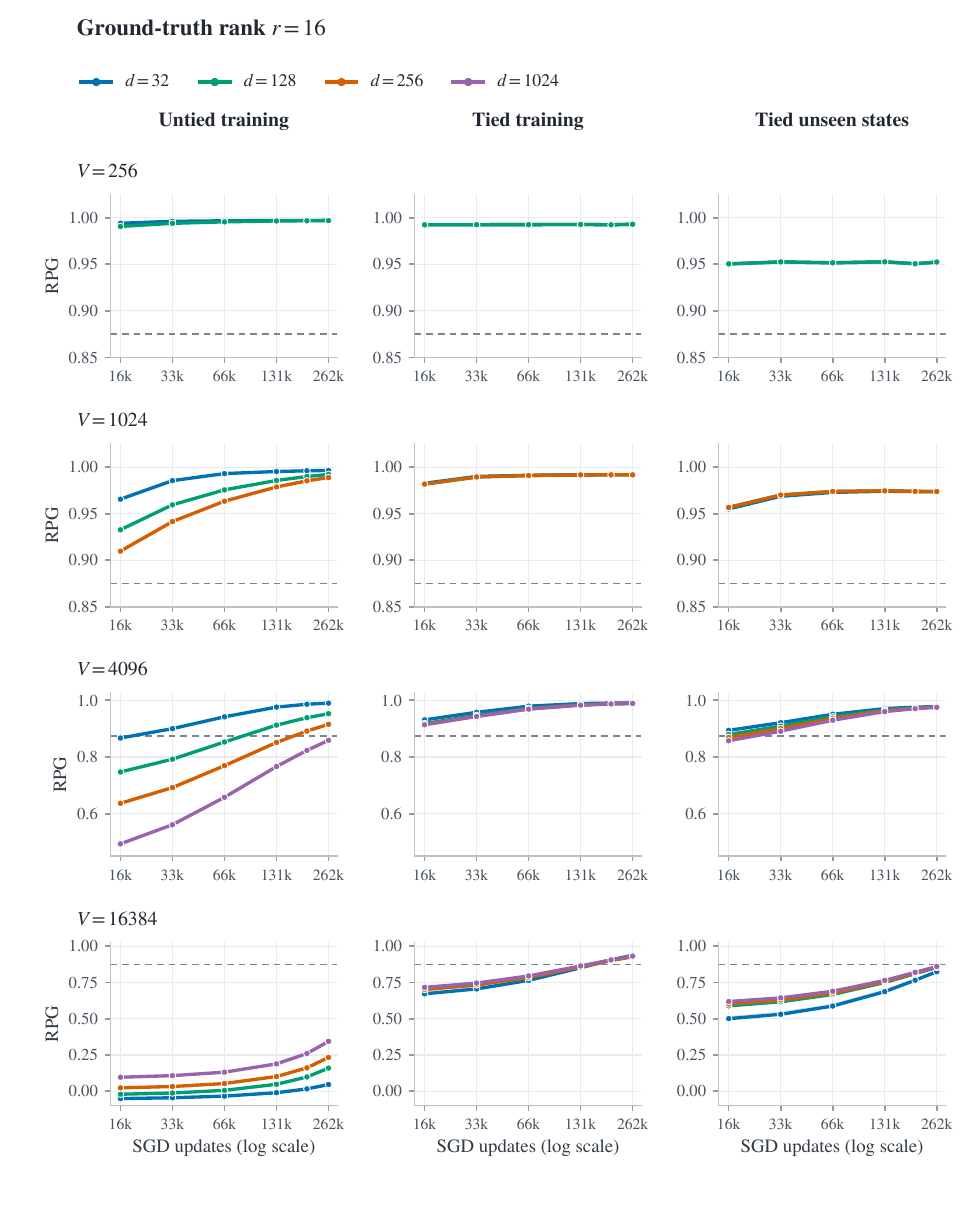}
 \caption{SGD with {sampled transitions at rank} $r=16$, {with}
 $V=256,1024,4096,16384$ from top to bottom. {All} valid {embedding
 dimensions} and all three initialization seeds are included. {The
 columns, curves,} and {shaded areas are the same as in}
 Figure~\ref{fig:scaling-sgd-r8}.}
 \label{fig:scaling-sgd-r16}
\end{figure}

\begin{figure}[p]
 \centering
 \includegraphics[width=\linewidth,height=0.85\textheight,keepaspectratio]{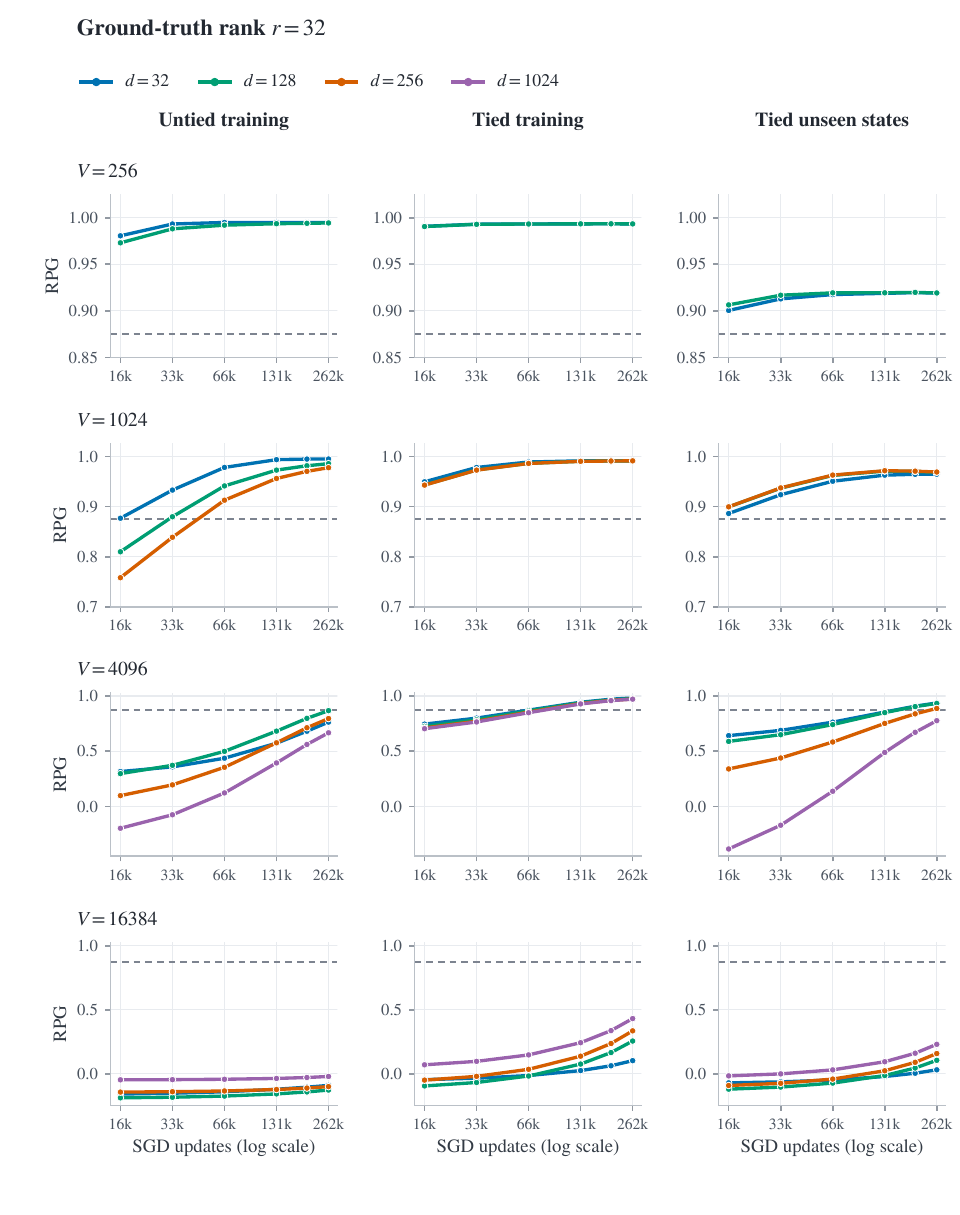}
 \caption{SGD with {sampled transitions at rank} $r=32$, {with}
 $V=256,1024,4096,16384$ from top to bottom. {All} valid {embedding
 dimensions} and all three initialization seeds are included{, and
 negative} gains are {shown as they are}. {The columns, curves,} and {shaded
 areas are the same as in} Figure~\ref{fig:scaling-sgd-r8}.}
 \label{fig:scaling-sgd-r32}
\end{figure}

{In this section, we provide the details of the experiments with sampled
transitions in \cref{sec:iclr-experiments}. We train the models by SGD}
over a grid of {the number of states, the rank of the ground truth, and the
embedding dimension}.
Figures~\ref{fig:scaling-sgd-r8}--\ref{fig:scaling-sgd-r32} show {the
results of} all 12 choices of $(V,r)$, with every valid {embedding
dimension} and all three initialization seeds.

\paragraph{Models and {configurations}.}
{We use the untied model} $P^{E,U}=\softmax(EU^\top)$ with
$E,U\in\mathbb R^{V\times d}$ {and} the tied model
$P^X=\softmax(XX^\top)$ with $X\in\mathbb R^{V\times d}$, as in
\cref{sec:iclr-setup}. We consider $V\in\{256,1024,4096,16384\}$,
ranks $r\in\{8,16,32\}$, and embedding dimensions
$d\in\{32,128,256,1024\}$. The untied model is trained on all $V$
current states. {The tied model is trained on} a fixed random set
$\mathcal S$ with $N=|\mathcal S|=3V/4$, and is evaluated on the unseen
set $\mathcal U=[V]\setminus\mathcal S$ of size $V/4$. Thus, the
numbers of observed and unseen states are $192/64$, $768/256$,
$3072/1024$, and $12288/4096$, respectively. {As before, all} $V$ states
{are possible next states, so the embedding of} an unseen state is still
updated through its {role as a next state. We only keep the embedding
dimensions} with $r\le d<N$, {where} $N=V$ for the untied model. This
gives 78 configurations and 234 runs {with initialization seeds $10$,
$11$, and $12$}.

\paragraph{{Ground truths}.}
For {the untied model, we draw two} independent $V\times r$ Gaussian
matrices {and} normalize each row. We multiply the first matrix by three,
subtract the mean row from the second matrix, and divide the second
matrix by its largest row norm. Then, $Q$ is the row-wise softmax of
the product of the first matrix and the transpose of the second matrix.
For the tied model, we normalize the rows of a $V\times r$ Gaussian
matrix, subtract the mean row, and rescale the resulting matrix $Y$
such that its largest row norm is $\sqrt3$. Then, we set
$Q=\operatorname{softmax}(YY^\top)$. In exact arithmetic, the logits
are row-centered, have rank at most $r$, and have entries bounded {by
three} in absolute value{. We use the ground-truth seed zero in all
configurations, where the generator of the ground truth has} seed
$10000+100V${, and the permutation that determines $\mathcal S$ has} seed
$73000+V$. The ground truth, {the set $\mathcal S$, and the stream of
sampled transitions are shared by} the three initializations of each
configuration. {Note that the ground truths of the tied and untied
models are constructed differently, and thus the panels of the two
models are not directly comparable}.

\paragraph{Training.}
{All entries of the embeddings are initialized independently from}
$\mathcal N(0,1/d)${, and all embeddings are trained jointly. At each
update, we draw} 512 fresh {transitions, where the current state is
sampled} uniformly from the {set of observed states, and the next state
is sampled from the corresponding row of} $Q$. {The next state is sampled
by rejection sampling based on the bound on the logits, which} avoids
storing the full transition table{, while the model still computes the
softmax over all} $V$ {states for each sampled current state}. We minimize
the cross-entropy loss by SGD with the weight decay penalty
$\frac\lambda2\|\Theta\|_\fro^2$, where $\lambda=1/(120rV)$ for the
untied model and $\lambda=0.4/(rV)$ for the tied model. We do not use
Gaussian perturbations, momentum, projection, or gradient clipping. The
learning rate follows a cosine schedule from 64 to 0.25 {in the first
8192 updates, and is fixed at 0.25 afterwards until} 131072 updates{,
with the same weight decay. Since fresh transitions are drawn at every
update, longer training uses more samples, instead of more} passes over
a fixed dataset. {For} $r=8$, the curves in
\cref{fig:scaling-r8-trajectories} {are obtained by continuing the same
runs} to 262144 updates with the same learning rate and {weight decay}.

\paragraph{Measurements.}
{We report the} relative prediction gain {$g_{\mathcal A}(\Theta)$ defined
in \cref{sec:iclr-setup}, where the risk}
$\mathcal R_{\mathcal A}(\Theta)$ is {the} population cross-entropy {loss,
and both the risk and the entropy} $\mathcal H_{\mathcal A}$ {are}
averaged uniformly over $\mathcal A$. {As in the main text, the gain
does not include the weight decay penalty. All reported gains are
evaluated on the entire set of observed states or the entire set of
unseen states, and not on the smaller subsets of states that are used
for monitoring during training.}

\paragraph{{Numerical settings.}}
{After the first 8192 updates, the runs use} FP32 parameters, BF16
{automatic mixed precision for training}, and TF32 where {it is} enabled.
Sampling and risk evaluation use FP32, with TF32 disabled for the
evaluation. The {first 8192 updates} used a different numerical {setting,
which includes} FP64 sampling and risk {evaluation. For this reason, the
figures in this section} start at 16384 updates{, and show the
checkpoints at 16384,} 32768, 65536, and 131072 updates, {which are all
obtained} under the same setting. The {curves in
\cref{fig:scaling-r8-trajectories} additionally include the checkpoints
at} 196608 and 262144 updates, {which are evaluated in the same way. In
the figures of this section, the curves connect the medians over} the
three seeds {at each checkpoint, and the shaded areas show the full
ranges over the seeds, while \cref{fig:scaling-r8-trajectories} only
shows the medians for readability. No smoothing or extrapolation is
applied. Each embedding dimension has a fixed color in all panels},
negative gains are {shown as they are}, and the dashed line marks $7/8${.
The} vertical scales are shared within {each row, and may differ across}
rows.

\paragraph{{Results}.}
{For} each of the three {metrics, i.e., the training gain of the untied
model, the training gain of the tied model, and the gain of the tied
model on the unseen states, the medians of all} 39 {configurations
increase from} 16384 {updates to} 131072 updates. Some configurations
already {achieve a high gain at 16384 updates, while the others are
still} below $7/8$ {and keep improving}. For example, at {$V=4096$, $r=32$,
and $d=256$, the training gain of the tied model increases from $0.711$
to $0.932$, and its gain on the unseen states increases from $0.340$ to
$0.752$} (Figure~\ref{fig:scaling-sgd-r32}). {This shows that the gain on
the unseen states can lag behind the training gain,} even when both {are
improving. We also observe that a larger embedding dimension does not
always give a higher gain under a fixed number of updates. We note that
these results are obtained by SGD with sampled transitions, which is
different from the setting of our theorems. Moreover, a low gain at the
last checkpoint} alone does not {imply that the run} has converged to a
{local minimum with low gain}.

\clearpage
\clearpage
\clearpage
\clearpage

\end{document}